\documentclass[11pt]{article}
\usepackage{amsmath,amsthm,amssymb,amsfonts,fullpage,verbatim,bm,graphicx,enumerate,epstopdf,lscape,subfigure,multirow}
\usepackage{algorithm,algorithmicx}
\usepackage{algpseudocode}
\usepackage{boxedminipage,longtable}
\usepackage[dvipsnames,usenames]{color}
\usepackage[dvipsnames]{xcolor}
\usepackage[pdftex,letterpaper=true,colorlinks=true,pdfpagemode=none,urlcolor=blue,linkcolor=blue,
   citecolor=blue,pdfstartview=FitH,plainpages=true]{hyperref}
\usepackage{tikz,verbatim,bibentry}
\usepackage{bm,mathrsfs,cite,mathabx}
\usepackage[comma,authoryear]{natbib}
\usepackage{thm-restate}

\newtheorem{lemma}{Lemma}

\theoremstyle{definition}
\newtheorem{rmk}{Remark}

\newcommand{\lam}{{\lambda}}

\newcommand{\cA}{{\cal A}}
\newcommand{\cB}{{\cal B}}

\newcommand{\cE}{{\cal E}}

\newcommand{\cX}{{\cal X}}

\newcommand{\bel}{\begin{eqnarray}\label}
\newcommand{\eel}{\end{eqnarray}}
\newcommand{\bes}{\begin{eqnarray*}}
\newcommand{\ees}{\end{eqnarray*}}

\def\mat{\hbox{\rm mat}}
\def\vec{\hbox{\rm vec}}

\def\E{\mathbb{E}}
\def\P{\mathbb{P}}
\def\RR{\mathbb{R}}
\def\R{\mathbb{R}}

\DeclareMathOperator*{\argmin}{\arg\min}

\def\ideal{(\text{\footnotesize ideal})}
\def\vec{\hbox{\rm vec}}
\def\mat{\hbox{\rm mat}}

\def\trace{\hbox{\rm trace}}

\def\textsum{\hbox{$\sum$}}
\def\textmax{\hbox{$\max$}}

\def\textprod{\hbox{$\prod$}}
\def\cpca{{\widehat a}^{\rm\tiny cpca}}
\def\ico{\rm\tiny ico}
\def\cpcatop{{\widehat a}^{\rm\tiny cpca\top}}
\def\cpcalam{{\widehat \lam}^{\rm\tiny cpca}}

\def\ahatm{\widehat a^{(m)}}
\def\bhatm{\widehat b^{(m)}}
\def\ghatm{\widehat g^{(m)}}
\def\ahatmtop{\widehat a^{(m)\top}}
\def\Ahatm{\widehat A^{(m)}}

\def\Bhatm{\widehat B^{(m)}}
\def\eps{\epsilon}

\begin{document}

\begin{center}
{\LARGE \textbf{Tensor Principal Component Analysis in High Dimensional CP Models}}
\end{center}
\medskip \centerline{Yuefeng Han and Cun-Hui Zhang\footnote{
Yuefeng Han is Assistant Professor, Department of Applied and Computational Mathematics and Statistics, University of Notre Dame, Notre Dame, IN 46556. Email: yuefeng.han@nd.edu. Cun-Hui Zhang is Professor, Department of Statistics, Rutgers University, Piscataway, NJ 08854. E-mail: czhang@stat.rutgers.edu. Han was supported in part by National Science Foundation
grant IIS-1741390 and DMS-2052949. Zhang was supported in part by NSF grants IIS-1741390, CCF-1934924, DMS-2052949 and DMS-2210850.
}}
\medskip \centerline{University of Notre Dame and Rutgers University}
\date{}

\begin{abstract}
The CP decomposition for high dimensional non-orthogonal spiked tensors is an important problem with broad applications across many disciplines. However, previous works with theoretical guarantee typically assume restrictive incoherence conditions on the basis vectors for the CP components. In this paper, we propose new computationally efficient composite PCA and concurrent orthogonalization algorithms for tensor CP decomposition with theoretical guarantees under mild incoherence conditions. The composite PCA applies the principal component or singular value decompositions twice, first to a matrix unfolding of the tensor data to obtain singular vectors and then to the matrix folding of the singular vectors obtained in the first step. It can be used as an initialization for any iterative optimization schemes for the tensor CP decomposition. The concurrent orthogonalization algorithm iteratively estimates the basis vector in each mode of the tensor by simultaneously applying projections to the orthogonal complements of the spaces generated by other CP components in other modes. It is designed to improve the alternating least squares estimator and other forms of the 
high order orthogonal iteration for tensors with low or moderately high CP ranks, and it is guaranteed to converge rapidly when the error of any given initial estimator is bounded by a small constant. Our theoretical investigation provides estimation accuracy and convergence rates for the two proposed algorithms. Both proposed algorithms are applicable to deterministic tensor, its noisy version, and the order-$2K$ covariance tensor of order-$K$ tensor data in a factor model with uncorrelated factors. Our implementations on synthetic data demonstrate significant practical superiority of our approach over existing methods.

\ \\
\noindent%
\end{abstract}

\section{Introduction} \label{section:introduction}

Motivated by modern scientific research, analysis of tensors, or high-order arrays, has emerged as one of the most important and active areas in machine learning, electrical engineering, and statistics. Tensors arise in numerous applications involving genomics \citep{alter2005, omberg2007}, multi-relational learning \citep{nickel2011three}, neuroimaging analysis \citep{zhou2013tensor, sun2017store}, recommender systems \citep{bi2018}, computer vision \citep{liu2012}, longitudinal data analysis \citep{hoff2015multilinear}, economic indicators \citep{chen2021statistical, chen2022factor}, finance data \citep{han2022rank} and more. In addition, tensor based methods have been applied to many statistics and machine learning problems where the observations are not necessarily tensors, such as community detection \citep{anandkumar2014community}, topic and
latent variable models \citep{anandkumar2014tensor}, graphical models \citep{chaganty2014estimating}, and high-order interaction pursuit \citep{hao2020sparse}. In many of these settings, the tensor of interest is high-dimensional, e.g. the ambient dimension is substantially larger than the sample size in the factor model in \eqref{model1} below. However, in practice, the tensor parameter often has intrinsic dimension-reduced structure, such as low-rankness and sparsity \citep{kolda2009tensor,udell2019big}, which 
motivates research in tensor estimation and in the 
recovery of the underlying structure. 

Low rank tensor decomposition is one of the most important tools for recovering and estimating the intrinsic tensor structure based on noisy data. It plays a similar role to matrix singular value decomposition (SVD) and eigendecomposition which are of fundamental importance throughout a wide range of fields including computer science, applied mathematics, machine leaning, statistics, signal processing, etc. Despite the well-established theory for low-rank decomposition of matrices, tensors present unique challenges. There are several notions of low-rankness in tensors, including the most popular CANDECOMP/PARAFAC (CP) low-rankness and multilinear/Tucker low-rankness. While CP models are more parsimonious and easier to interpret in many applications, compared with Tucker models, the computation of the best low-rank CP approximation of a given tensor is NP hard in general \citep{haastad1990tensor,kolda2009tensor,hillar2013most}.

In this paper, we develop a new framework of tensor principal component analysis (tensor PCA) applicable to deterministic tensors, their noisy version, and factor models with uncorrelated factors. To be specific, let us first consider the factor model. Suppose we have i.i.d. matrix or tensor valued observations (such as 2-D or 3-D images) $\cX_i$, $1\le i\le n$, of the following form
\begin{equation}\label{model1}
\cX_i=\textsum_{j=1}^r w_j f_{ij} a_{j1}\otimes a_{j2}\otimes\cdots\otimes a_{jK}+\cE_i,
\end{equation}
where $\otimes$ denotes tensor product, $f_{ij}$ are i.i.d $N(0,1)$, $w_j>0$ represent certain weights, $ a_{jk}\in \R^{d_k}$ are basis vectors with $\| a_{jk}\|_2=1$ for all $1\le j\le r$, $1\le k\le K$, $\cE_i$ are i.i.d. noise tensors each with i.i.d $N(0,\sigma^2)$ entries. Tensor factor models like \eqref{model1}, where the data is written as the sum of a low-rank factor and noise, have been studied extensively in the literature. While both the Tucker and CP decompositions can be used to model such data, the Tucker model has been the focus of the literature in the consistent estimation of tensor structure in the presence noise, largely due to the direct expression of the Tucker decomposition with matrix SVD. However, in many applications, CP decomposition is a more attractive modeling option. To estimate the structural parameters in model \eqref{model1}, 
we construct the covariance tensor of the data $\cX_i$, $T = n^{-1}\sum_{i=1}^n\cX_i\otimes \cX_i$, 
which can be written as 
\begin{align}\label{model}
T=\textsum_{j=1}^r \lambda_j \otimes_{k=1}^{2K} a_{jk}+\Psi,    
\end{align}
where $\lambda_j=w_j^2$, $a_{j,K+k}=a_{jk}$ for $1\le j\le K$, and $\Psi$ is a noise tensor. We treat \eqref{model} as a general CP model in which $\Psi$ is allowed to have an identity component $\E[\Psi]\,\propto\, \text{Id}$, e.g. $\E[\Psi] = \sigma^2 \text{Id}$ under \eqref{model1}, as the estimated basis vectors do not depend on the identity component in our approach. Here $\text{Id}$ is the identity tensor given by $\max_{[K]}(\text{Id})=I_{d\times d}$. Our main goal is to estimate the basis vectors $a_{jk}$, which can be also called loading vectors, from the noisy tensor $T$. We call \eqref{model} spiked covariance tensor model as it is analogous to the so called ``spiked covariance model'' in the study of matrix PCA in high dimensions \citep{johnstone2009consistency}. Here, $a_{jk}$, $1\le j\le r$, are not necessarily orthogonal to each other in each mode $k$.

When $K=2$, our model \eqref{model1} is closely related to $(2D)^2$-PCA in the community of image signal processing, which has been extensively studied \citep{yang2004two, zhang20052d, kong2005generalized, pang2008binary, kwak2008principal, li2010l1, meng2012improve, wang2015robust}. 
This literature has been mainly focused on the algorithmic properties. However, statistical guarantees such as consistency of estimators and risk analysis, in high demand in many applications, are much less understood in the CP model. Among notable exceptions is \cite{anandkumar2014guaranteed}. 
In Tucker factor models, statistical analysis has been carried out by  \cite{hoff2011separable,fosdick2014separable,chen2020semiparametric, chen2021statistical, yu2021projected, lam2021rank} 
under a very different setting from \eqref{model1}, and by \cite{chen2020constrained,chen2022factor,han2020iterative} with matrix and tensor time series.

In view of \eqref{model}, a natural approach to the estimation of $a_{jk}$ is minimizing the empirical loss
\begin{align} \label{problem:ls}
( a_{jk}, 1\le k\le K, 1\le j\le r) 
= \argmin_{\substack{ \| a_{jk}\|_2 =1, k\in [K], j\in [r]} } \min_{\lam_j, j\in [r]}
\Big\| T -\textsum_{j=1}^r \lambda_j (\otimes_{k=1}^K a_{jk} )^{\otimes2} \Big\|_{\rm HS}^2,
\end{align}
where $\|\cA\|_{{\rm HS}}$, defined as $\| \vec(\cA)\|_2$, is the Hilbert Schmidt norm of tensor $\cA$. 
However, due to the non-convexity of \eqref{problem:ls}, a straightforward implementation of local search algorithms, such as gradient descent and alternating minimization, may get trapped into local optimums and result in sub-optimal statistical performance. 
Still, if one starts from an initialization not too far from the true basis vectors, local search is likely to perform well. 

In addition to the order $2K$ tensor $T$ in \eqref{model} with paired CP basis vectors $a_{jk}$, we study in this paper the following more general low-rank CP model, 
\begin{align}\label{model2}
T=\textsum_{j=1}^r \lambda_j \otimes_{k=1}^{N} a_{jk}+\Psi,    
\end{align}
where $\lambda_j>0$ and $\Psi$ is a noise tensor, including the noiseless version with $\Psi=0$. While $a_{jk}$ can be all different in \eqref{model2}, $a_{j1}=\cdots =a_{jN}$ for the empirical $N$-th moment tensor in certain latent-variable models \citep{anandkumar2014tensor}.

\subsection{Our contributions}

We propose a new method for the estimation of the basis/loading vectors $a_{jk}\in\R^{d_k}$ in the spiked covariance tensor model \eqref{model} or the low-rank CP model \eqref{model2}, both can be viewed as spiked CP models. 
The new method is composed of two steps: (i) a composite PCA (CPCA) as a warm-start initialization; (ii) an iterative concurrent orthogonalization (ICO) scheme to refine the estimator. The intuition is that the CP components in higher order tensors are closer to orthogonal and tend to have higher order coherence in a multiplicative form, and the proposed method is designed to take advantage of this feature of the CP model to achieve higher statistical and computational efficiency.
To the best of our knowledge, this proposal is the first to explicitly aim to benefit from this multiplicative higher order coherence in CP decomposition. Existing initialization procedures require random projections and may need to generate many copies to yield a reasonably good choice, while the CPCA produces definitive initial estimates of the CP basis vectors via tensor unfolding/refolding and spectral decomposition. 
The ICO scheme aims to achieve higher order of numerical convergence than the alternating least squares and other forms of the high order orthogonal iteration (HOOI) \citep{de2000, liu2014, zhang2018tensor, han2020iterative} after the warm-start, again by taking benefits of the multiplicative coherence.

The CPCA and ICO algorithms are developed in Section \ref{section:estimation} along with some sharp tensor perturbation bounds to motivate them. These tensor perturbation bounds, new or not readily available and potentially useful elsewhere, heuristically justify our ideas and the individual elements of the proposed algorithms. Statistical guarantees for the CPCA and ICO estimators are provided 
in Section \ref{section:theories}. 
Our perturbation and risk bounds explicitly exhibit the benefits of the multiplicative nature of the coherence of the tensor bases and the rapid growth of such benefits as the order of the tensor increases.

Our analyses of the proposed methods focus on the cases where the tensor dimensions $d_k$ are typically much larger than the CP rank $r$ but $r$ can be also large. Theoretical studies of existing proposals of tensor de-noising in CP models typically imposes very restrictive incoherence conditions on the CP components; For example, the incoherence condition $\vartheta_{\max} = \max_{k, j_1\neq j_2}|a_{j_1k}^\top a_{j_2k}| \lesssim \text{ploylog} (d_k)/\sqrt{d_k}$ in \cite{anandkumar2014guaranteed}. In contrast, we prove that the CPCA yields useful estimates when $r^{2}\vartheta_{\max}^K$ is small and the ICO provides fast convergence rates when $r^{5/2}\vartheta_{\max}^K$ is small, demonstrating the advantage of our approach in terms of model assumption. Computationally, the errors in the ICO propagate in the quadratic or higher order. Similar to Nesterov's acceleration in gradient descent, the high-order of error propagation guarantees 
$\epsilon$ numerical precision within $\log\log(1/\epsilon)$ iterations. To the best of our knowledge, this is the first provable $\log\log(1/\epsilon)$ iteration guarantee in non-orthogonal CP models. Numerical comparisons with existing methods demonstrate advantages of the proposed approach.


\subsection{Related work}
There is a large literature on tensor decomposition. As it is beyond the scope of this paper to give a comprehensive survey, we only review the most related papers.

The most commonly used algorithm for CP decomposition is alternating least squares \citep{comon2009tensor}, which has no general convergence guarantee. 
Theoretical studies of alternating least squares have focused on the estimation of tensors with orthogonal CP decomposition from randomized initialization. Noticeably, \cite{anandkumar2014tensor} developed a robust tensor power method for orthogonal CP decomposition. \cite{montanari2014statistical} proposed a tensor unfolding approach for rank-one tensors and compared it with tensor power iteration with random initialization. \cite{wang2017tensor} improved the initialization procedure for orthogonal CP decomposition by projecting the observed tensor down to a matrix and then applying the matrix power method.  \cite{wangmy2017tensor} developed a two-mode higher-order SVD algorithm for higher order tensors. For the non-orthogonal tensors, one may first convert the tensor into an orthogonal form known as {\it whitening}, but the procedure is ill-conditioned in high dimensions \citep{le2011ica, souloumiac2009joint} and computationally expensive \citep{huang2013fast}.

Recently, another line of research has been developed on non-orthogonal tensor CP decomposition, still focused on randomized initialization. \cite{anandkumar2014guaranteed} studied non-orthogonal CP decomposition and established convergence guarantees for a modification of the alternating least squares. In addition to their incoherence conditions on deterministic CP bases as discussed in the previous subsection, they considered independent random basis vectors uniformly distributed in the unit sphere, essentially imposing 
a {\it soft orthogonality} constraint. 
\cite{sun2017provable} further extended their work to the case where the CP basis vectors are sparse. In the noiseless case ($\Psi=0$), \cite{sharan2017orthogonalized} introduced orthogonalized alternating least squares algorithm and studied its performance under  the soft orthogonality constraint or small $r^2\vartheta_{\max}$. \cite{kuleshov2015tensor} developed a minimum distance algorithm for non-orthogonal CP decomposition which uses random projections to reduce the problem to simultaneous matrix diagonalization, but the applicability of their theoretical results to diverging $d_k$ is unclear. \cite{colombo2016tensor} developed an iterative Gauss-Newton algorithm for joint matrix diagonalization. However, \cite{sharan2017orthogonalized} claimed that the simultaneous diagonalization algorithm is not computationally efficient. 


\subsection{Notation and tensor preliminaries} \label{section:notation}

Let $[n]$ denote the set $\{1, 2, \ldots, n\}$. For a vector with entries $\pi_j$ or a set of real numbers $\{\pi_j\}$, we denote by $\pi_{j,\pm}=\min_{i\neq j}|\pi_i-\pi_j|\wedge|\pi_j|$ the gap from $\pi_j$ to $\{0, \pi_i, i\neq j\}$ and set  $\pi_{\min}=\min_j|\pi_j|$ and $\pi_{\max}=\max_j|\pi_j|$. 
For convenience, we call $\lam_{j,\pm}$ the $j$-th eigengap in models \eqref{model} and \eqref{model2}. 
For a matrix $B = (b_{ij})\in \RR^{p\times n}$, we denote its singular values by $\sigma_1(B)\ge\sigma_2(B)\ge \cdots\ge \sigma_{\min\{p,n\}}(B)\ge 0$, 
its Frobenius norm by $\|B\|_{\rm F} = (\sum_{ij} b_{ij}^2)^{1/2}=(\sum_{j=1}^{\min\{p,n\}}\sigma_i^2(B))^{1/2}$, 
and its spectral norm by $\|B\|_{\rm S} =\sigma_1(B)$. 

For any two vectors $u$ and $\widehat u$ of unit length, we measure the distance between the spaces they generate by the absolute sine of the angle $\theta(\widehat u,u)$ between the two vectors,
\begin{equation}\label{loss}
\big|\sin \theta(\widehat u,u)\big| 
= \|\widehat u\widehat u^\top - uu^\top\|_{\rm S} 
=(1-(u^\top\widehat u)^2)^{1/2} =  \|\widehat u\widehat u^\top - uu^\top\|_{\rm F}/\sqrt{2}. 
\end{equation}
We note that $\min_{\pm} \|\widehat u \pm u\|_2 = (1-|u^\top\widehat u|)^{1/2}
= \big|\sin \theta(\widehat u,u)\big|/(1+|u^\top\widehat u|)^{1/2}$.


For any two tensors $\cA\in\RR^{m_1\times m_2\times \cdots \times m_K}, \cB\in \RR^{r_1\times r_2\times \cdots \times r_N}$, denote the tensor product $\otimes$ as $\cA\otimes \cB\in \RR^{m_1\times \cdots \times m_K \times r_1\times \cdots \times r_N}$, such that
$(\cA\otimes\cB)_{i_1,...,i_K,j_1,...,j_N}=(\cA)_{i_1,...,i_K}(\cB)_{j_1,...,j_N}$. 
For two vectors $a$ and $b$, $a \otimes b$ is equivalent to the outer product $ab^\top$. 
Given $\cA\in \R^{m_1\times\cdots\times m_K}$ and $m=\prod_{j=1}^K m_j$, let ${\rm{vec}}(\cA)\in \R^m$ be vectorization of the matrix/tensor $\cA$, 
$\mat_k(\cA)\in \R^{m_k\times(m/m_k)}$ the mode-$k$ matrix unfolding of $\cA$, and 
$\mat_k(\vec(\cA))=\mat_k(\cA)$. 
For example, for $K=3$ 
$$({\rm{mat}}_1(\cA))_{i,(j+m_2(k-1))}= ({\rm{mat}}_2(\cA))_{j,(k+m_3(i-1))}= ({\rm{mat}}_3(\cA))_{k,(i+m_1(j-1))} =\cA_{ijk}.$$ 
Similarly, for nonempty $J\subseteq [K]$, $\text{mat}_J(\cA)$ is the mode $J$ matrix unfolding which maps $\cA$ to $m_J\times m_{-J}$ matrix with $m_J=\prod_{j\in J}m_j$ and $m_{-J}=m/m_J$, e.g. $\text{mat}_{\{1,2\}}(\cA)=\text{mat}_3^\top(\cA)$ for $K=3$. The mode-$k$ product of $\cA\in\RR^{m_1\times m_2\times \cdots \times m_K}$ with a matrix $U\in\RR^{m_k\times r_k}$ is an order $K$-tensor of size $m_1\times \cdots m_{k-1} \times r_k\times m_{k+1}\times m_K$ and will be denoted as $\cA\times_k U$, so that
$$ (\cA\times_k U)_{i_1,...,i_{k-1},j,i_{k+1},...,i_K}=\textsum_{i_k=1}^{m_k} \cA_{i_1,i_2,...,i_K} U_{i_k,j}.  $$
The Hilbert Schmidt norm for a tensor $\cA\in\RR^{m_1\times m_2\times \cdots \times m_K}$ is defined as 
$\|\cA\|_{{\rm HS}}=\|\vec(\cA)\|_2$. 
An order $K$ tensor $T \in \RR^{m_1\times m_2\times \cdots \times m_K}$ is said to have rank one if it can be written as
$$T =w\cdot a_1\otimes\cdots \otimes a_K,$$
where $w\in\R$ and $a_k\in\R^{m_k}$ are unit vectors for identifiability. A tensor $T\in \RR^{m_1\times m_2\times \cdots \times m_K}$ is said to have a CP rank $r\ge 1$ if it can be written as a sum of $r$ rank-1 tensors, 
$$T=\textsum_{j=1}^r w_j\cdot a_{i1}\otimes\cdots \otimes a_{iK}.$$

\section{Estimation procedures} \label{section:estimation}

\subsection{Spiked covariance tensor model}\label{section:spike}
In this section, we focus on the spiked covariance tensor model \eqref{model}. 
We introduce the composite PCA (CPCA) as Algorithm \ref{algorithm:initial}, and the iterative concurrent orthogonalization (ICO) as Algorithm~\ref{algorithm:projection}. 

As mentioned in the introduction, our main idea is to take advantage of the multiplicative 
higher-order coherence of the CP components for faster convergence. We begin with an explicit description of this phenomenon. 
Let $\Sigma_k = (\sigma_{ij,k})_{r\times r}=A_k^\top A_k$ with the mode-$k$ basis matrix $A_k = (a_{1k},\ldots,a_{rk})\in \R^{d_k\times r}$ in \eqref{model}. 
As $\sigma_{jj,k}=\|a_{jk}\|_2^2=1$, the correlation among columns of $A_k$ can be measured by 
\bel{corr-k}
\vartheta_k = \textmax_{1\le i < j\le r}|\sigma_{ij,k}|,\ \ 
\delta_k = \|\Sigma_k - I_{r}\|_{\rm S},\ \ 
\eta_{jk} = (\textsum_{i\in [r]\setminus \{j\}}\sigma_{ij,k}^2)^{1/2}.
\eel
However, the CP components are much less correlated. 
By \eqref{model}, the matrix unfolding of $T$, 
\bel{mat_{[K]}}
\mat_{[K]}(T) = n^{-1}\textsum_{i=1}^n \vec(\cX_i)\vec(\cX_i)^\top 
= \textsum_{j=1}^r \lambda_j a_ja_j^\top+\mat_{[K]}(\Psi)\in \R^{d\times d}, 
\eel
has basis matrix $A = (a_1,\ldots,a_r)\in \R^{d\times r}$ with $a_j = \vec(\otimes_{k=1}^Ka_{jk})$ and correlation measures 
\bel{corr-all}
\vartheta = \textmax_{1\le i < j\le r}|a_{i}^\top a_{j}|,\quad 
\delta = \|A^\top A - I_{r}\|_{\rm S}, 
\eel
where $d=\prod_{j=1}^Kd_j$. 
As $a_{i}^\top a_{j} = \prod_{k=1}^K a_{ik}^\top a_{jk} =\prod_{k=1}^K \sigma_{ij,k}$, 
the coherence is bounded by $\vartheta\le \prod_{k=1}^K\vartheta_k\le \vartheta_{\max}^K$. 
The spectrum norm $\delta$ is also bounded by 
the products of quantities in \eqref{corr-k}. 
We summarize these elementary relationships in the following proposition.

\begin{restatable}{proposition}{PropositionDelta}\label{prop:delta} 
For any set $S$ of tensor modes, define $a_{jS}=\vec(\otimes_{k\in S}a_{jk})$, $A_{S} = (a_{1S},\ldots,a_{rS})$, 
$\vartheta_S = \max_{1\le i < j\le r}|a_{iS}^\top a_{jS}|$ and 
$\delta_S = \|A_S^\top A_S - I_{r}\|_{\rm S}$. Define 
\bes
\mu_S = \max_j \min_{k_1,k_2\in S} 
\max_{i\neq j} \prod_{k\neq k_1,k\neq k_2,k\in S}
\sqrt{r}|\sigma_{ij,k}|/\eta_{jk}. 
\ees 
as the (leave-two-out) mutual coherence of $\{A_j,j\in S\}$. 
Then, $\mu_S\in~[1, r^{|S|/2-1}]$, 
\begin{align}
&\delta_S\le\min_{k \in S}\delta_k, \quad 
\delta_S \le (r-1)\vartheta_S \le (r-1)\prod_{k\in S} \vartheta_k,\\
&\delta_S \le \mu_S r^{1-|S|/2}\textmax_{j\le r}\prod_{k\in S}\eta_{jk}
\le \mu_S r^{1-|S|/2}\prod_{k\in S}\delta_k.  \label{eq:prop}
\end{align} 
When $S=[K]$, the above inequalities hold with $\{\delta_S, \vartheta_S\}$ replaced by the $\{\delta,\vartheta\}$ in \eqref{corr-all}.
\end{restatable}


We note that \eqref{eq:prop} implies $\delta\le \max_{j\le r} \prod_{k=1}^K\eta_{jk} \le \prod_{k=1}^K\delta_k$ 
due to $\mu_*r^{1-K/2}\le 1$. 
When (most of) the quantities in \eqref{corr-k} are small, the products in \eqref{eq:prop} would be much smaller, 
so that $a_j$ are nearly orthogonal in \eqref{mat_{[K]}}.  
This motivates the use of the PCA of  
\eqref{mat_{[K]}} to estimate $\lam_j$ and $a_j$, 
\bel{uhat_j}
\mat_{[K]}(T) = \textsum_{j} \cpcalam_j \widehat u_j \widehat u_j^\top. 
\eel
The following proposition gives explicit justifications of \eqref{uhat_j} with sharp perturbation bounds.

\begin{restatable}{proposition}{PropositionTrans}\label{lemma-transform}
Let $d \ge r$ and $A \in \R^{d\times r}$ with $\|A^\top A - I_r\|_{\rm S}\le\delta$. 
Let $A=\widetilde U_1 \widetilde D_1 \widetilde U_2^\top$ be the SVD of $A$, and 
$U = \widetilde U_1\widetilde U_2^\top$. Then, 
$\|A \Lambda A^\top - U \Lambda U^{\top}\|_{\rm S}\le \delta \|\Lambda\|_{\rm S}$  
for all nonnegative-definite matrices 
$\Lambda$ in $\R^{r\times r}$. 
\end{restatable}

By Proposition \ref{lemma-transform}, $\widehat u_j$ in \eqref{uhat_j} can be viewed as an estimate of $u_j$, $1\le j\le r$, satisfying 
\bel{error-u_j}
\|a_ja_j^\top - u_ju_j^\top\|_{\rm S} 
= \|A (e_je_j^\top) A^\top - U (e_je_j^\top) U^{\top}\|_{\rm S}  
\le \delta.
\eel
Because $\mat_k(a_j) 
=a_{jk}\vec(\otimes_{\ell\in [K]\setminus\{k\}} a_{j\ell})^\top$, the natural estimate of $a_{jk}$ based on 
the $\widehat u_j$ in \eqref{uhat_j} is 
\bel{ahat_jk}
\cpca_{jk} = \hbox{the top left singular vector of $\mat_k(\widehat u_j)$}.  
\eel
The following proposition explicitly justifies \eqref{ahat_jk} 
with a sharp perturbation bound.

\begin{restatable}{proposition}{PropositionRankOneApprox}
\label{prop-rank-1-approx} 
Let $M\in \R^{d_1\times d_2}$ be a matrix with $\|M\|_{\rm F}=1$ and 
${a}$ and ${b}$ be unit vectors respectively in $\R^{d_1}$ and $\R^{d_2}$. 
Let $\widehat a$ be the top left singular vector of $M$. 
Then, 
\begin{equation}\label{prop-rank-1-approx-1}  
\big(\|{\widehat a} {\widehat a}^\top - {a} {a}^\top\|_{\rm S}^2\big) \wedge (1/2)
\le \|\vec(M)\vec(M)^\top - \vec({a} {b}^\top)\vec({a} {b}^\top)^\top\|_{\rm S}^2. 
\end{equation}
\end{restatable}

Proposition \ref{prop-rank-1-approx} is sharp in the sense that equality is attainable in \eqref{prop-rank-1-approx-1} when the right-hand side is less than $1/2$, 
and that for any $c\in [1/2,1]$ the maximal distance $\|\widehat a \widehat a^\top -aa^\top\|_{\rm S}=1$ is attainable for some $\{M,a,b\}$ with the right-hand side of  \eqref{prop-rank-1-approx-1} being exactly $c$. 

The CPCA is the two-step procedure given by the PCA in \eqref{uhat_j} and SVD in \eqref{ahat_jk}. 
With $\vec(M)=\widehat u_j$, $a=a_{jk}$ and $b=\otimes_{\ell\neq k}a_{j\ell}$, Proposition \ref{prop-rank-1-approx} asserts that the second step of the CPCA is a contraction once the first step yields an estimate of $a_j=\otimes_k a_{jk}$ within 45 degrees. 
As the correlations among $a_j$ is much smaller than those among $a_{jk}$ in each mode, 
this condition is much more explicit and of a much weaker form than those in the literature for the estimation of $a_{jk}$ after random projection \citep{anandkumar2014guaranteed, sharan2017orthogonalized}. 
By the perturbation bounds in Propositions \ref{lemma-transform} and \ref{prop-rank-1-approx} 
and Wedin's perturbation theorem, in the noiseless case ($\sigma=0$ and $n\to\infty$) with $\Psi=0$ in \eqref{model}  
\bel{noiseless-bd}
|\cpcalam_j -\lam_j| \le \delta\lam_1,\quad 
\big(\|\cpca_{jk}\cpcatop_{jk} - {a}_{jk} {a}_{jk}^\top\|_{\rm S}^2\big) \wedge (1/2)
\le (1 + 2\lam_1/\lam_{j,\pm})^2\delta^2, 
\eel
so that the CPCA takes advantage of the multiplicative coherence of 
$\otimes_{k=1}^K a_{jk}$ in view of the product bounds for 
$\delta$ in Proposition \ref{prop:delta}. We state the CPCA as Algorithm 1 as follows.

\begin{algorithm}[htbp]
\caption{Composite PCA (CPCA) for pairwise symmetric tensors}\label{algorithm:initial}
\begin{algorithmic}[1]
\Require noisy tensor $T = n^{-1}\sum_{i=1}^n\cX_i\otimes \cX_i$,  
CP rank $r$
\State Formulate $T$ to be a $d\times d$ matrix $\mat_{[K]}(T)$ as in \eqref{mat_{[K]}} 
with $d=\prod_{k=1}^K d_k$
\State Compute the $r$ top $\cpcalam_j$ and $\widehat u_j$ 
in the eigenvalue decomposition of $\mat_{[K]}(T)$ as in \eqref{uhat_j}
\State Compute $\cpca_{jk}$ as the top left singular vector of 
$\mat_k(\widehat u_j)\in\R^{d_k\times (d/d_k)}$ as in \eqref{ahat_jk} 
\Ensure 
$\cpca_{jk}, \ \cpcalam_j=\kappa_j, \ j=1,...,r, \ k=1,...,K$
\end{algorithmic}
\end{algorithm}

After obtaining a warm start through the CPCA (Algorithm \ref{algorithm:initial}), we propose to use the 
ICO (Algorithm \ref{algorithm:projection} below) to refine the solution. 
The ICO can be viewed as an extension of HOOI \citep{de2000,zhang2018tensor} and the iterative projection algorithm in \cite{han2020iterative} to undercomplete ($r < d_{\min}$) and non-orthogonal CP decompositions. 
However, ICO differentiates from these methods and the alternating least squares in the following important way: In updating the model-$k$ basis vector $a_{jk}$, the ICO projects the observed tensor $T$ to the orthogonal complements of the span of $\{a_{i\ell}, i \neq j, i\le r\}$ in $\R^{d_\ell}$ for all $\ell\neq k$ simultaneously from $2(K-1)$ sides. 
Here the word ``concurrent" in ICO refers to the feature that the projections take place in all modes $\ell\neq k$ at the same time point/step in the computational iterations. In contexts where ``time" has special meaning such as time series, iterative simultaneous orthogonalization can be used instead of ICO. 
Given estimates $\widetilde A_\ell=(\widetilde a_{1\ell},\ldots, \widetilde a_{r\ell})$ for the mode-$\ell$ basis matrix $A_\ell=(a_{1\ell},\ldots, a_{r\ell})\in\R^{d_\ell\times r}$, 
this is done by projecting $T$ to the nonnegative-definite
\begin{align} \label{eq:cp-ideal}
\widetilde  T_{jk} = T \times_{l\in [2K]\backslash\{k,K+k\} } \widetilde b_{jl}^\top 
\approx \lambda_j a_{jk} a_{jk}^\top + \widetilde \Psi_{jk} 
\end{align}
with $\widetilde B_\ell = (\widetilde b_{1\ell},..., \widetilde b_{r\ell}) = \widetilde A_\ell(\widetilde A_\ell^{\top} \widetilde A_\ell)^{-1}$ and $\widetilde b_{j,K+\ell} = \widetilde b_{jl}$, as $A_\ell^\top \widetilde B_\ell\approx I_r$ when $\widetilde A_\ell\approx A_\ell$. Thus, it is natural to update $a_{jk}$ using the top eigenvector of $\widetilde  T_{jk}$. This is the ICO in Algorithm \ref{algorithm:projection} below.


\begin{algorithm}[ht]
\caption{Iterative Concurrent Orthogonalization (ICO) for pairwise symmetric tensors}\label{algorithm:projection}
\begin{algorithmic}[1]
\Require noisy tensor $T = n^{-1}\sum_{i=1}^n\cX_i\otimes \cX_i$,  
CP rank $r$, warm-start $\widehat a_{jk}^{(0)}\in\R^{d_k}, j\in [r], k\in [K]$, 
tolerance parameter $\epsilon>0$, maximum number of iterations $M$
\State  Compute $(\widehat b_{1k}^{(1)},\ldots, \widehat b_{rk}^{(1)})\in \R^{d_k\times r}$ as the right inverse of 
$(\widehat a_{1k}^{(0)}, \ldots, \widehat a_{rk}^{(0)})^\top, k\in [K]$; Set $m=0$
\Repeat 
\State Set $m=m+1$ 
\For $k=1$ to $K$
\For $j = 1$ to $r$
\State Compute 
$T_{jk}^{(m)} =T \times_{l\in [2K]\setminus \{k,K+k\}}(\widehat  b_{jl}^{(m)})^\top \in \R^{d_k \times d_k}$ as in \eqref{eq:cp-ideal}, 
$b_{j,K+l}^{(m)} = b_{jl}^{(m)}$
\State Compute $\widehat a_{jk}^{(m)}$ as the top eigenvector
of $T_{jk}^{(m)}$
\EndFor
\State Compute $(\widehat b_{1k}^{(m)},\ldots, \widehat b_{rk}^{(m)})$ as the right inverse of 
$(\widehat a_{1k}^{(m)}, \ldots, \widehat a_{rk}^{(m)})^\top$ 
\State Set $(\widehat b_{1k}^{(m+1)},\ldots, \widehat b_{rk}^{(m+1)})=(\widehat b_{1k}^{(m)},\ldots, \widehat b_{rk}^{(m)})$
\EndFor
\Until $m=M$ or
$\max_{j,k}\| \widehat a_{jk}^{(m)}  \widehat a_{jk}^{(m)\top} - \widehat a_{jk}^{(m-1)} \widehat a_{jk}^{(m-1)\top}\|_{\rm S}\le \epsilon$

\Ensure 
$\widehat  a_{jk}^{\ico}=\widehat  a_{jk}^{(m)}$,  
$\widehat \lambda_{j}^{\ico} = T\times_{k=1}^{2K} (\widehat  b_{jk}^{(m)})^\top , \ j=1,...,r,\ k=1,...,K$
\end{algorithmic}
\end{algorithm}

\begin{restatable}{proposition}{PropositionICO}\label{prop-ICO} 
Let $T^*=\E[T]$ with the tensor $T$ in \eqref{model}. 
Given $\widetilde A_\ell=(\widetilde a_{1\ell},\ldots, \widetilde a_{r\ell}),\ell\in[K]\setminus\{k\}$, 
let $\widetilde  a_{jk}^*$ be the top eigenvector of 
$\widetilde  T_{jk}^* = T^*\times_{l\in [2K]\backslash\{k,K+k\} } \widetilde b_{jl}^\top\in\R^{d_k\times d_k}$ 
with the $\widetilde b_{jl}$ in \eqref{eq:cp-ideal}. Then,  
\bes
\big\|a_{jk}a_{jk}^\top - \widetilde a^*_{jk}\widetilde a^{*\top}_{jk}\big\|_{\rm S} 
\le 2\big(1 + \delta_k\big)(\lam_1/\lam_j)\prod_{\ell\in [K]\setminus \{k\}}(\widetilde\phi_\ell/(1-\widetilde\phi_\ell)_+)^2, 
\ees
where $\widetilde\phi_\ell = \widetilde\psi_\ell/\big(\sqrt{(1-\delta_\ell)(1-1/(4r))} -\sqrt{r}\widetilde\psi_\ell\big)_+$ 
with $\widetilde\psi_\ell = \max_{j\le r}\big\|\widetilde a_{j\ell}\widetilde a_{j\ell}^\top - a_{j\ell}a_{j\ell}^\top \big\|_{\rm S}$.
\end{restatable}

The perturbation bound in Proposition \ref{prop-ICO} explicitly proves the power of concurrent orthogonalization: 
In terms of the angle between the one-dimensional spaces generated by ${\widehat a}_{jk}^{\ico}$ and $a_{jk}$ and up to some scaling constants, the error in the estimation of $a_{jk}$ in each step is bounded by a product of $2(K-1)$ carryover errors in all other modes in the noiseless case $\Psi=0$ in model \eqref{model}, i.e. $\sigma=0$ and $n\to\infty$ in model \eqref{model1}. In this sense, the ICO error propagates in the order of $2(K-1)>1$, which implies high order contraction.
See Subsection \ref{subsec:errors} for a more detailed discussion in a comparison between the ICO and the alternating least squares in closely related model \eqref{model2}. 
As in the analysis of accelerated gradient descent in which the error propagates in the second order, the ICO is expected to achieve $\epsilon$ accuracy within $\log\log(1/\epsilon)$ iterations 
in the noiseless case in model \eqref{model}. 
This property of the ICO is confirmed in Theorem~\ref{thm:noiseless} and extended to the noisy case in Theorem \ref{thm:projection2} below in Section~\ref{section:theories}.


\subsection{General high order tensors}\label{section:low_rank}

In Section \ref{section:spike}, we focus on $2K$-th order tensors which can be unfolded as a symmetric matrix. 
In this section, we extend the CPCA and ICO algorithms to general $N$-th order tensors. 

In model \eqref{model2}, we present the following proposition as an extension of Proposition \ref{lemma-transform}. It also covers the study of the CPCA of spiked covariance tensors developed in Section \ref{section:spike}. Similar to Section \ref{section:spike}, Proposition \ref{prop:delta}, Proposition \ref{lemma-transform-ext} and Proposition~\ref{prop-rank-1-approx} together provide heuristic justifications for the CPCA in Algorithm \ref{algorithm:initial2} below and a road map to study it in model \eqref{model2}. 

\begin{restatable}{proposition}{PropositionTransExt}\label{lemma-transform-ext} 
Let  $A \in \R^{d_1\times r}$ and $B \in \R^{d_2\times r}$ with 
$\|A^\top A - I_r\|_{\rm S}\vee \|B^\top B - I_r\|_{\rm S} \le\delta$ and $d_1\wedge d_2\ge r$. 
Let $A=\widetilde U_1 \widetilde D_1 \widetilde U_2^\top$ be the SVD of $A$, 
$U = \widetilde U_1\widetilde U_2^\top$, $B=\widetilde V_1 \widetilde D_2 \widetilde V_2^\top$ 
the SVD of $B$, and $V = \widetilde V_1\widetilde V_2^\top$. 
Then, $\|A \Lambda A^\top - U \Lambda U^{\top}\|_{\rm S}\le \delta \|\Lambda\|_{\rm S}$  
for all nonnegative-definite matrices $\Lambda$ in $\R^{r\times r}$, and 
$\|A Q B^\top - U Q V^{\top}\|_{\rm S}\le \sqrt{2}\delta \|Q\|_{\rm S}$  
for all $r\times r$ matrices $Q$. 
\end{restatable}

We note that in Proposition \ref{lemma-transform-ext}, $U$ is a function of $A$ and $V$ is the same function of $B$. Specifically, $U$ does not depend on $\Lambda$, and $V$ does not depend on $\{A,U,Q\}$.


\begin{algorithm}[htbp]
\caption{Composite PCA (CPCA) for general $N$-th order tensors}\label{algorithm:initial2}
\begin{algorithmic}[1]
\Require noisy tensor $T=\sum_{j=1}^r \lambda_j \otimes_{k=1}^N a_{jk} + \Psi \in \RR^{d_1\times\cdots\times d_N}$, CP rank $r$, $S \subset [N]$
\State If $S=\emptyset$, pick $S$ to maximize $\min(d_S,d/d_S)$ 
with $d_S = \prod_{k\in S}d_k$ and $d = \prod_{k=1}^N d_k$
\State Unfold $T$ to be a $d_S\times (d/d_S)$ matrix $\mat_S(T)$
\State Compute $\cpcalam_j, \widehat u_j, \widehat v_j$ 
as the top components in the SVD $\mat_S(T) = \sum_{j}\cpcalam_j \widehat u_j \widehat v_j^\top$
\State Compute $\cpca_{jk}$ as the top left singular vector of 
$\mat_k(\widehat u_j)$, $k\in S$, or $\mat_k(\widehat v_j)$, $k\in S^c$
\Ensure
$\cpca_{jk}, \ \cpcalam_j, \ j=1,...,r, \ k=1,...,N$
\end{algorithmic}
\end{algorithm}

In practice, a sensible way to unfold $T$ is to form a matrix as square as possible with input $S=\emptyset$  in Algorithm \ref{algorithm:initial2}. As in Algorithm \ref{algorithm:projection} we propose to use $\widehat a_{jk}^{(0)}=\cpca_{jk}$ as warm-start of the ICO in Algorithm~\ref{algorithm:projection2} below.  

\begin{rmk}
For order 3 tensors, either $|S|=1$ or $|S^c|=1$ in Step 1 of Algorithm \ref{algorithm:initial2}. Assume $d_1\ge d_2\vee d_3$ for definiteness so that we choose $S=\{1\}$. The CPCA exhibits advantage in terms of coherence by Proposition \ref{prop:delta} if and only if $\delta_1< \delta_2\vee\delta_3$, e.g when $\delta_k\downarrow d_k$.  
\end{rmk}

\begin{algorithm}[ht]
\caption{Iterative Concurrent Orthogonalization (ICO) for general $N$-th order tensors}\label{algorithm:projection2}
\begin{algorithmic}[1]
\Require noisy tensor $T=\sum_{j=1}^r \lambda_j \otimes_{k=1}^N a_{jk} + \Psi \in \RR^{d_1\times\cdots\times d_N}$, CP rank $r$, warm-start $\widehat a_{jk}^{(0)}, j\in [r], k\in [N]$, tolerance parameter $\epsilon>0$, maximum number of iterations $M$

\State  Compute $(\widehat b_{1k}^{(1)},\ldots, \widehat b_{rk}^{(1)})$ as the right inverse of 
$(\widehat a_{1k}^{(0)}, \ldots, \widehat a_{rk}^{(0)})^\top, k\in [N]$; Set $m=0$
\Repeat 
\State Set $m=m+1$
\For $k=1$ to $N$ 
\For $j=1$ to $r$ 
\State Compute 
$T_{jk}^{(m)} = T \times_{l\in [N]\setminus\{k\}}  (\widehat  b_{jl}^{(m)})^\top \in \R^{d_k }$ 
\State Compute $\widehat a_{jk}^{(m)} = T_{jk}^{(m)}/\| T_{jk}^{(m)} \|_2$
\EndFor
\State Compute $(\widehat b_{1k}^{(m)},\ldots, \widehat b_{rk}^{(m)})$ as the right inverse of 
$(\widehat a_{1k}^{(m)}, \ldots, \widehat a_{rk}^{(m)})^\top$
\State Set $(\widehat b_{1k}^{(m+1)},..., \widehat b_{rk}^{(m+1)})=(\widehat b_{1k}^{(m)},..., \widehat b_{rk}^{(m)})$ 
\EndFor
\State Compute $\widehat \lambda_{j}^{(m)} = \big|T\times_{k=1}^N (\widehat  b_{jk}^{(m)})^\top\big|, j\in [r]$
\Until $m=M$ or
$\max_{j,k}\| \widehat a_{jk}^{(m)}  \widehat a_{jk}^{(m)\top} - \widehat a_{jk}^{(m-1)} \widehat a_{jk}^{(m-1)\top}\|_{\rm S}\le \epsilon$
\Ensure 
$\widehat  a_{jk}^{\ico}=\widehat  a_{jk}^{(m)}$, 
$\widehat \lambda_{j}^{\ico} = \big|T\times_{k=1}^N (\widehat  b_{jk}^{(m)})^\top\big|, 
j\in [r], k\in [N]$ 
\end{algorithmic}
\end{algorithm}


Similar to Proposition \ref{prop-ICO}, we present a fresh Proposition \ref{prop-ICO-gen} to describe the high order of error propagation in the ICO iterations in model \eqref{model2}.

\begin{restatable}{proposition}{PropositionICOgen}\label{prop-ICO-gen} 
Let $T^*=\E[T]$ with the tensor $T$ in \eqref{model2}. 
Given $\widetilde A_\ell=(\widetilde a_{1\ell},\ldots, \widetilde a_{r\ell}), \ell\in[N]\setminus\{k\}$, 
let $(\widetilde b_{1\ell},\ldots,\widetilde b_{r\ell})=\widetilde A_\ell(\widetilde A_\ell^\top\widetilde A_\ell)^{-1}$,    
$\widetilde  T_{jk}^* = T^*\times_{l\in [N]\backslash\{k\} } \widetilde b_{jl}^\top\in\R^{d_{k}}$, $\widetilde  a_{jk}^* = \widetilde  T_{jk}^*/\|\widetilde  T_{jk}^*\|_2$ 
and $\widetilde\lambda_j^* = T^*\times_{l\in [N]} \widetilde b_{jl}^\top$. Then,  
\begin{align*}
2-2\big|a_{jk}^\top \widetilde a^*_{jk}\big|
&\le 2(r-1) \big(1 + \delta_k\big)(\lam_1/\lam_j)^2 \prod_{\ell\in [N]\setminus \{k\}}(\widetilde\phi_\ell/(1-\widetilde\phi_\ell))^2, \\
\big| \widetilde\lambda_j^*/\lambda_j -1 \big| & 
\le\sum_{\ell=1}^N \widetilde\phi_{\ell} + (r-1)(\lambda_1/\lambda_j) \prod_{\ell=1}^N \widetilde \phi_{\ell},
\end{align*}
where $\widetilde\phi_\ell = \widetilde\psi_\ell/\big(\sqrt{1-\delta_\ell}-\sqrt{r}\widetilde\psi_\ell \big)_+$ 
with $\widetilde\psi_\ell = \max_{j\le r}\big(2 - 2\big|\widetilde a_{j\ell}^\top a_{j\ell}\big|\big)^{1/2}$.
\end{restatable} 

We note that $2-2|a^\top b|=\min_{\pm}\|a\pm b\|_2^2$ and that each $a_{jk}$ is identifiable only up to a $\pm$ sign. Again, by Proposition \ref{prop-ICO-gen}, the ICO is expected to have a super-linear computational convergence under the loss \eqref{loss} in the noiseless case $\Psi=0$ in model \eqref{model2}. This is confirmed in Theorem \ref{thm:noiseless-gen} and extended to the noisy case in Theorem \ref{thm:projection}
below in Section \ref{section:theories}.

\def\tLS{{\hbox{\rm\footnotesize \,LS}}}
\subsection{Error propagation in ICO and alternating LSE}\label{subsec:errors}

The merit of the ICO can be more directly seen from a comparison with alternating least squares 
in model \eqref{model2}, $T=T^*+\Psi$ with target tensor $T^* = \sum_{j=1}^r \lam_j\otimes_{k=1}^Na_{jk}$ and noise $\Psi$. Given estimates $\widehat a_{jk}^{(m)}, k>1$,  
the LSE of $A_1\Lambda =(\lam_1a_{11},\ldots,\lam_ra_{r1})$ is 
\bes
\mat_1(T)\widehat B^\tLS_{-1}
&=& \hbox{$\argmin$}_{M\in \R^{d_1\times r}}\Big\| \mat_1(T) - M\big(\widehat A^{(m)}_{-1}\big)^\top\Big\|_{\rm HS}^2
\cr &=& A_1\Lambda + A_1\Lambda \big(A_{-1} -\widehat A^{(m)}_{-1}\big)
\widehat B^\tLS_{-1} + \mat_1(\Psi)\widehat B^\tLS_{-1}, 
\ees
where $\widehat B^\tLS_{-1}$ is the right inverse of 
$\widehat A^{(m)}_{-1}=(\widehat a^{(m)}_{1,-1},\ldots,\widehat a^{(m)}_{r,-1}\big)\in R^{d_{-1}\times r}$ 
with $\widehat a^{(m)}_{j,-1} = \vec\big(\otimes_{k=2}^N\widehat a^{(m)}_{jk}\big)$ and $d_{-1}=d/d_1$. 
Because $\widehat A^{(m)}_{-1} - A_{-1}$ is an $(N-1)$-degree polynomial of the carryover errors 
$\widehat a^{(m)}_{jk} - a_{jk}$ and the polynomial has a nonvanishing linear term, 
the leading term of the bias $\mat_1(T^*)\widehat B^\tLS_{-1} - A_1\Lambda$ of the LSE is linear 
in the carryover error. 
In comparison, in the ICO, the right inverse is 
taken in Algorithm \ref{algorithm:projection2} in individual modes before tensor multiplication, 
\bes
\widehat A_1^{(m+1)} \widehat \Lambda^{(m+1,1)} 
= \mat_1(T) \widehat B^{(m)}_{-1} 
= \mat_1(T^*) \widehat B^{(m)}_{-1} + \mat_1(\Psi)\widehat B^{(m)}_{-1}, 
\ees
where $\widehat B^{(m)}_{-1}=(\widehat b^{(m)}_{1,-1},\ldots,\widehat b^{(m)}_{r,-1}\big)\in R^{d_{-1}\times r}$ 
with $\widehat b^{(m)}_{j,-1} = \vec\big(\otimes_{k=2}^N\widehat b^{(m)}_{jk}\big)$, 
$\widehat \Lambda^{(m+1,1)}$ is a diagonal matrix to normalize the estimated basis vectors to $\big\|\widehat a_{j1}^{(m+1)}\big\|_2=1$.
The noise terms $\mat_1(\Psi)\widehat B^\tLS_{-1}$ and $\mat_1(\Psi)\widehat B^{(m)}_{-1}$ 
are comparable between the two methods. 
However, as 
\bes
\mat_1(T^*) \widehat B^{(m)}_{-1} 
&=& \bigg(\bigg(\lam_1\prod_{k=2}^N a_{1k}^\top \widehat b^{(m)}_{1k}\bigg)a_{11}\,,\ldots,\, 
\bigg(\lam_r \prod_{k=2}^N  a_{rk}^\top \widehat b^{(m)}_{rk}\bigg)a_{r1}\bigg)
\cr
&& + \bigg( 
\sum_{j=2}^r \bigg(\lam_j \prod_{k=2}^N a_{jk}^\top \widehat b^{(m)}_{1k}\bigg)a_{j1}\,,\ldots,\,
\sum_{j=1}^{r-1} \bigg(\lam_j \prod_{k=2}^N a_{jk}^\top \widehat b^{(m)}_{rk}\bigg)a_{j1}\bigg)
\ees
with $a_{j_1k}^\top b^{(m)}_{j_2k} = I\{j_1=j_2\}+(a_{j_1k}-a^{(m)}_{j_1k})^\top b^{(m)}_{j_2k}$,  
the leading term in the bias of the ICO, as the second term above, is a homogeneous polynomial of degree $N-1$ in terms 
of the carryover error. 
We note that the errors in the diagonal $\prod_{k=2}^N a_{jk}^\top \widehat b^{(m)}_{jk}$ 
is linear in terms of the carryover error but they are absorbed into $\widehat\Lambda^{(m+1,1)}$. 

In summary, the alternating least squares operator $\widehat B^\tLS_{-1}$ is the inverse of tensor product, while 
the ICO operator $\widehat B^{(m)}_{-1}$ is the tensor product of inverses in $N-1$ individual modes. Consequently, the bias of an alternating least squares step is proportional to the norm of the carryover error 
and the bias of an ICO step is proportional to the $(N-1)$-th power of the norm of the carryover error. 
Meanwhile, the noise terms of the two methods are comparable.

\subsection{Algorithm complexity}
Assume the input tensor is $T$.  Algorithm \ref{algorithm:initial} (CPCA) costs $O(d^2r)$ floating-point operations (flops) for $r$-truncated eigen decomposition of ${\rm mat}_{[K]}(T)$ and $O(d)$ flops for 1-truncated SVD of ${\rm mat}_k(\widehat u_i)$, so that the total cost of CPCA is $O(d^2r)$. In each iteration of Algorithm \ref{algorithm:projection} (ICO), the calculation of $\widehat B_k$ costs $O( d_kr^2)$ flops, the matrix manipulation in step 6 costs $O(d^2)$ flops, and the 1-truncated eigen decomposition of $T_{jk}^{(m)}$ in step 7 costs $O(d_k^2)$ flops. Hence, the total cost per iteration in Algorithm \ref{algorithm:projection} is also $O(d^2r)$. 
Similarly, in Section \ref{section:low_rank}, the total cost of Algorithm \ref{algorithm:initial2} is $O(r\prod_{k=1}^N d_k)$, and the cost of each iteration in Algorithm \ref{algorithm:projection2} is also $O(r\prod_{k=1}^N d_k)$.
In summary, the cost of CPCA and each iteration of ICO is of the order of the product of the CP rank and the number of entries in tensor $T$. 

In a spiked covariance tensor model \eqref{model1}, the top $r$ eigenvalue decomposition of the unfolded covariance tensor $\mat_{[K]}(T)$ is equivalent to the top $r$ SVD of the $n^{1/2}$-normalized unfolded data matrix $(\vec(\cX_1),...,\vec(\cX_n))/\sqrt{n}\in\R^{d\times n}$. In this sense, Algorithms \ref{algorithm:initial} and \ref{algorithm:projection} can be modified accordingly to adopt matrix SVD. The total cost of the first SVD in Algorithm \ref{algorithm:initial} becomes $O(dnr)$, so that the total cost of Algorithm \ref{algorithm:initial} is $O(dnr)$. Similarly, the total cost per iteration in Algorithm \ref{algorithm:projection} is $O(dnr)$. As the cost to construct covariance tensor $T$ is $O(d^2n)$, it can be computationally more efficient to perform the SVD directly.

While the topic is beyond the scope of this paper, we note that random projection and other remedies can be used to reduce the cost of computing low-rank PCA and SVD when the signal to noise ratio is high.

\subsection{Identification and estimation of CP component groups}

In principle, the top $r$ singular space $(\widehat u_1,...,\widehat u_r)$ in CPCA (Algorithms \ref{algorithm:initial} and \ref{algorithm:initial2}) might not be uniquely determined; for example, this occurs in the presence of ties in $\lam_j$. 
In such cases, CPCA and ICO may still be used to identify and estimate CP component groups with tied singular values. To avoid redundancy, we describe the procedures below only for the symmetric tensors in \eqref{model}. 

Suppose there are $g$ groups of singular values with distinct representative values $\lambda_{(1)} > \cdots > \lambda_{(g)}>0$ and respective group sizes $r_1,...,r_g$, $r_1+\cdots+r_g=r$. Suppose \eqref{model} can be written as 
\bes
T= \textsum_{i=1}^g T_{(i)} + \Psi,\quad 
T_{(i)} = \textsum_{j\in G_i}\lam_j \otimes_{k=1}^{2K} a_{jk} \approx  \lambda_{(i)}\textsum_{j\in G_i}  \otimes_{k=1}^{2K} u_{jk}, 
\ees
where $\{G_1,\ldots,G_g\}$ is a partition of $[r]$ with $|G_i|=r_i$. 
By Proposition \ref{lemma-transform}, it is reasonable to consider the case where $\{u_{jk}, j\in G_i, i=1,\ldots,g\}$ are orthonormal for each $k\in [K]$ and 
\bes 
\max_{j\le r}\|a_{jk}a_{jk}^\top - u_{jk}u_{jk}^\top\|_{\rm S} \le \delta^*,\quad \|A\Lambda A^\top - \textsum_{i=1}^g \lambda_{(i)}\textsum_{j\in G_i} u_{j}u_j^\top \big\|_{\rm S}\le \lam_{(1)}\delta^*
\ees
with $u_j=\vec\big(\otimes_{k=1}^K u_{jk}\big)$ and a certain $\delta^*\approx\delta$. Suppose further that $2\delta^*\lam_{(1)}<\min_{i\le g}(\lam_{(i)}-\lam_{(i+1)})$ with $\lam_{(g+1)}=0$. 
It would then be reasonable to consider clustering of the outputs of CPCA and ICO to identify the groups $G_i$, 
e.g. using $\cpcalam_j$ by Proposition \ref{lemma-transform}. 
Given $G_j$, a group ICO could be used to estimate the individual $T_{(i)}$ and then a Tucker decomposition would give the column space of $A_{G_ik}=(a_{jk}, j\in G_i)$ for each $(i,k)$. 

Once a good estimate of the group tensor $T_{(i)}$ becomes available, the identification of individual components $\lam_j\otimes _{k=1}^{2K}a_{jk}, j\in G_i$, in the group would be feasible if a rank-one component can be identified in the linear span of the group components. This feasibility can be seen from Proposition \ref{prop-rotation} below. 
Since the identifiability issue does not require paired CP bases as in \eqref{model}, Proposition \ref{prop-rotation} is stated under model \eqref{model2}. Kruskal's Theorem \citep{kruskal1977three} also provides the uniqueness of tensor CP decomposition.


\begin{restatable}{proposition}{PropositionRotate}\label{prop-rotation}
Let $\text{SP}:= \text{span} \{a_1,...,a_r\}$, where $a_j=\vec(\otimes_{k=1}^N a_{jk})\in\R^d$. The elements of SP can be viewed as either length $d$ vectors or $d_1\times\cdots \times d_N$ tensors. Suppose $N>2$ and $\delta_k<1$ for every $k=1,\ldots,N$ in \eqref{corr-k}, then every rank-1 tensor in SP is one of $a_j$'s up to a scalar.
\end{restatable}

The above discussions, written in response to an interesting question raised by referees, seem to deserve further investigation. However, a more comprehensive discussion or further development in this direction is beyond the scope of this paper.

\section{Theoretical properties} \label{section:theories}

\subsection{Spiked covariance tensor models}
In this section, we investigate theoretical guarantees of the proposed algorithms for the estimation of the CP basis vectors $a_{jk}$ for the spiked covariance tensor \eqref{model} with data in \eqref{model1}. 
As in \eqref{loss} we use 
$\|\widehat a_{jk}\widehat a_{jk}^\top  - a_{jk} a_{jk}^\top \|_{\rm S}
=(1-(\widehat a_{jk}^\top  a_{jk})^2)^{1/2} = \sup_{z\perp a_{jk},\|z\|_2=1}|z^\top \widehat a_{jk}|$  
to measure the distance between $\widehat a_{jk}$ and $ a_{jk}$.

We do not impose the orthogonality condition on the mode-$k$ CP basis vectors 
$\{a_{jk}, j\le r\}$ or even global incoherence condition on 
$\vartheta_{\max}:=\max_k\max_{1\le i< j\le r} | a_{ik}^\top a_{jk}|$ as in the literature 
\citep{anandkumar2014guaranteed, anandkumar2014tensor, sun2017provable, hao2020sparse, sharan2017orthogonalized}. 
However, we require the vectorized basis tensors $a_j=\vec( \otimes_{k=1}^K a_{jk})$ to satisfy the isometry condition 
$\delta =\|A^\top A-I_r\|_{\rm S}<1$, $A=(a_1,\ldots,a_r)$, or more conveniently the incoherence condition 
$\vartheta = \max_{i\neq j}|a_i^\top a_j|<1/r$. We recall that by Proposition \ref{prop:delta},  $\delta$ and $\vartheta$ 
are bounded by the respective products of their mode-$k$ counterparts defined in \eqref{corr-k}, so that we impose much weaker conditions compared with the existing ones on $\vartheta_{\max}$. In fact, the higher the tensor order $K$, the faster the convergence rate we offer given $\{r,\delta,\vartheta\}$, and the smaller $\delta$ and $\theta$ given $r$ and $\vartheta_{\max}$. 
Our analysis is based on the perturbation bounds in Propositions \ref{lemma-transform}, \ref{prop-rank-1-approx} 
and \ref{prop-ICO} in Section \ref{section:estimation} and proper concentration inequalities. 
For simplicity, we assume $\lambda_1>\lambda_2 > \cdots > \lambda_r$ with $\lambda_j=w_j^2$ in \eqref{model1} and \eqref{model}.

\begin{restatable}{theorem}{TheoremNoiseless}
\label{thm:noiseless} 
Suppose Algorithm \ref{algorithm:initial} (CPCA) is applied to the noiseless
$T^* = \sum_{j=1}^r \lam_j \otimes_{k=1}^{2K}a_{jk}$ with $a_{j,K+k}=a_{jk}$. Then, \eqref{noiseless-bd} holds 
for the resulting $\cpcalam_j$ and $\cpca_{jk}$. Let $\lam_{\min,\pm}= \min_{1\le j\le r}\lambda_{j,\pm}$ be the minimum eigengap.
Suppose further that 
\bel{thm:noiseless-1} 
2\max\big\{\delta_{\max},(\sqrt{r}+1)\psi_0\big\} \le 1,\ \  3(\lam_1/\lam_r)\psi_0^{2K-3} 
\le \rho <1, 
\eel
where $\delta_{\max}=\max_{k\le K}\delta_k$ with the $\delta_k$ in \eqref{corr-k} and $\psi_0 = (1 + 2\lam_1/\lam_{\min,\pm})\delta$ 
with the $\delta$ in \eqref{corr-all}. 
Let $\gamma_K\in (3 -3/K,3)$ be the solution of $\gamma_K^K - 3\gamma_K^{K-1}+2=0$,
e.g. $\gamma_3=2.732$, $\gamma_4=2.919$. 
If the resulting $\cpca_{jk}$ are used as the initialization of Algorithm \ref{algorithm:projection} (ICO) with the same data $T^*$, then 
\bes
\max_{j\le r}\big\|\widehat a_{jk}^{(m)}\widehat a_{jk}^{(m)\top} - a_{jk}a_{jk}^\top \big\|_{\rm S} 
\le \psi_{m,k} = \psi_0 \rho^{\gamma_K^{(m-1)K+ k-1}}
\ees
and $\max_{1\le k\le K}\psi_{m,k}\le \epsilon$ within 
$m=\lceil K^{-1}\{1 + (\log \gamma_K)^{-1}\log(\log(\psi_0/\epsilon)/\log(1/\rho)) \}\rceil$ iterations.
\end{restatable}

\begin{rmk}[Condition on the initial estimator]\label{rmk:initial}
The constant factors 2 and 3 in \eqref{thm:noiseless-1} are not sharp. In fact, condition \eqref{thm:noiseless-1} is simplified from the following, 
\bel{thm:noiseless-2} 
\frac{ 2(1+\delta_{\max})(\lam_1/\lam_r)\psi_0^{2K-2}} {\big(\sqrt{(1 - \delta_{\max})(1-1/(4r))}-(\sqrt{r}+1)\psi_0\big)_+^{2K-2}} 
\le \rho \psi_0 <\psi_0, 
\eel
which is slightly sharper and actually used in the proof. Here $\psi_0$ is an error bound for the initial estimator. The essence of our analysis of the ICO is that under \eqref{thm:noiseless-2}, $\psi_m \le C_0\psi_{m-1}^{2K-2}$ for the error bound $\psi_m=\max_{k\le K}\psi_{m,k}$ in the $m$-th iteration. 
\end{rmk}

\begin{rmk}[Incoherence condition]
When the minimum eigenvalue gap satisfies $\lam_{\min,\pm}\gtrsim \lam_1/r$, condition \eqref{thm:noiseless-1} asserts that the CPCA needs no stronger incoherence condition than $\vartheta_{\max}=O(r^{-5/(2K)})$, in view of Proposition \ref{prop:delta}. In comparison, conditions of stronger form are imposed in the literature; For example the initial estimator in \cite{anandkumar2014guaranteed} requires the incoherence condition $\vartheta_{\max} \le {\rm polylog}(d_{\min})/\sqrt{d_{\min}}$ for 3-way tensors. Compared with the previous work, \eqref{thm:noiseless-1} implies a weaker incoherence condition when $r\lesssim d_{\min}^{(K/5)\wedge 1}$. 
\end{rmk}

Theorem \ref{thm:noiseless} explicitly guarantees the high-order convergence of the ICO algorithm with the CPCA initialization in the noiseless case. To the best of our knowledge, the proposed ICO is the first algorithm known to achieve $\epsilon$-accuracy guarantee within $\log\log(1/\epsilon)$ number of iteration passes in non-orthogonal CP models. 

We proceed to present the statistical properties of the proposed estimator in the presence of noise, with input data $T$ in \eqref{model}. 
Define  
\bes
\hbox{\rm SNR} 
= \frac{\E\big\|\textsum_{j=1}^r w_j f_{ij}\otimes_{k=1}^K a_{jk}\big\|_{\rm HS}^2}{\E\|\cE_i\|_{\rm HS}^2}
\ees 
as the signal-to-noise ratio (SNR) in the   covariance tensor CP model \eqref{model1}. As $\lam_j=w_j^2$ and $\E[f_{ij}^2]=1$, 
\bel{SNR}
\hbox{\rm SNR} 
= \frac{\trace\big(\mat_{k}(T^*)\big)}{\sigma^2 d}
= \frac{\textsum_{j=1}^r\lam_j}{\sigma^2 d} 
= \frac{r_{\rm \tiny eff}\lam_1}{\sigma^2d}  
\eel
with the signal tensor $T^* = \sum_{j=1}^r\lam_j \otimes_{k=1}^{2K}a_{jk}$, where $r_{\rm \tiny eff} = \sum_{j=1}^r\lam_j/\lam_1$, no greater than the CP rank $r$, can be viewed as the effective rank of $T^*$. 

\begin{restatable}{theorem}{theoremfactorcpca}\label{thm:initial2}
Consider spiked covariance tensor model \eqref{model} with data in \eqref{model1}, $\lam_j=\omega_j^2$ and $\delta = \|A^\top A-I_r\|_{\rm S}$ as in \eqref{corr-all}. In an event with probability at least $1-e^{-t}$, 
Algorithm \ref{algorithm:initial} (CPCA)
gives the following error bound for the estimation of the CP basis vectors $a_{jk}$, 
\begin{equation}\label{thm:initial2:eq1}
\|\cpca_{jk}\cpcatop_{jk} - a_{jk} a_{jk}^\top \|_{\rm S} \le (1
+  2\lambda_1/\lam_{j,\pm})\delta  + C(\lam_1/\lam_{j,\pm})
\big(R^{(0)}+\sqrt{t/n}\big)
\end{equation}
for all $1\le j\le r$, $1\le k\le K$ 
and $0\le t\le d$, where $C$ is a numeric constant, $\lam_{j,\pm}$ is the $j$-th eigengap with $\{\lam_j\}$, 
and 
$R^{(0)} = \sqrt{(r_{\rm \tiny eff}/n)(1+1/\hbox{\rm SNR})(1+(r_{\rm \tiny eff}/d)/\hbox{\rm SNR})}
\le \sqrt{(r+\sigma^2d/\lam_1)(1+\sigma^2/\lam_1)/n}$. 
\end{restatable} 

The CPCA error bound \eqref{thm:initial2:eq1} consists of two parts. The first part involving $\delta$ is induced by the non-orthogonality of the vectors $a_{jk}$, which can be viewed as bias; The second part comes from a concentration bound for the centered random noise tensor $\Psi - \E[\Psi]$, which can be viewed as stochastic error. 
When the minimum eigengap satisfies $\lam_{\min,\pm} \gtrsim \lam_1/r$, Theorem \ref{thm:initial2} asserts that the CPCA needs no stronger incoherence condition than $\vartheta_{\max}=O(r^{-2/K})$, in view of Proposition \ref{prop:delta}. As long as $r\lesssim d_{\min}^{K/4}$, this incoherence condition is weaker than those in the existing literature for tensor denoising in CP models  \citep{anandkumar2014guaranteed}. 
The error bound \eqref{thm:initial2:eq1} is dominated by the bias when $\delta\gtrsim R^{(0)}$, and by the stochastic error when $R^{(0)}\gtrsim \delta$. The stochastic error $R^{(0)}$ can be further divided into two components: the impact of the fluctuation of the signal factor $f_{ij}$ represented by the parametric rate $\sqrt{r_{\rm \tiny eff}/n}$, and the impact of the noise $\cE_i$ in \eqref{model1} represented by $\sqrt{(r_{\rm \tiny eff}/n)/\hbox{\rm SNR}}$.  
The noise component dominates the stochastic error iff SNR$\,>1$. Still, the consistency of the CPCA in Theorem \ref{thm:initial2} requires a SNR condition $\text{SNR}\gtrsim r^3/n$, 
parallel to the condition $\sqrt{\lambda_r/\sigma^2}\ge Cr\sqrt{d/n}$ in the scenario considered in \cite{zhang2018tensor}.  

Next, we consider the theoretical properties of the ICO. We assume below for simplicity that $d_1\le \cdots\le d_K$. Let 
\bel{R_jk}
R^{\ideal}_{jk} &=& \big(\sigma^2/\lam_j
 + \sigma/\lam_j^{1/2}\big)\sqrt{d_k/n}. 
\eel 
and for $\phi\ge 0$ define 
\bel{R_jk-ideal}
R^{\ideal}_{jk,\phi} = R^{\ideal}_{jk} 
+ (\phi\wedge 1)\textsum_{\ell\in [K]\setminus \{k\}}R^{\ideal}_{j\ell}. 
\eel 
For constants $\psi_0\in (0,1)$ and $C_0 \ge 1$, define 
\bel{alpha}
\alpha &=& \sqrt{1-\delta_{\max}}-(r^{1/2}+1)\psi_0/\sqrt{1-1/(4r)},
\cr \rho &=& C_{0,\alpha}(\lam_1/\lam_r)\psi_0^{2K-3},
\cr \rho_1 &=& C_{0,\alpha}\sqrt{(\lam_1/\lam_r)r/n}\psi_0^{K-2},
\\ \nonumber \phi_0 &=& C_{0,\alpha}\sqrt{2r/(1-1/(4r))}R^{\ideal}_{rK,1},
\eel
with $\delta_{\max}=\max_{k\in [K]}\delta_k$ and $C_{0,\alpha}=C_0\alpha^{2-2K}$. 
Let $\mathscr P_\pm$ be the class of all $r\times r$ diagonal matrices $\Pi_r$ with $\Pi_r^2=I_r$. 

\begin{restatable}{theorem}{theoremfactorico}\label{thm:projection2}
Suppose that with a proper numeric constant $C_0$ and 
the quantities defined in \eqref{R_jk}, \eqref{R_jk-ideal} and \eqref{alpha},
\bel{C_0}\label{thm-projection2:eq1}
\alpha > 0,\ \rho_1\le \rho<1,\ 
C_{0,\alpha}R^{\ideal}_{rK,1}\le \psi_0 < 1. 
\eel 
Let $\Omega_0=\{\max_{j,k}\|\widehat a^{(0)}_{jk}\widehat a^{(0)^\top}_{jk} - a_{jk}a_{jk}^\top\|_{\rm S}\le\psi_0\}$ 
for any initial estimates $\widehat a^{(0)}_{jk}$. 
Then, Algorithm \ref{algorithm:projection} (ICO) provides 
\bel{th-3-bd}\label{thm-projection2:eq2}
&& \P\bigg\{\max_{j,k} \min_{\Pi_r\in\mathscr P_\pm}
\frac{\big\|\widehat A_k^{\ico}\Pi_r  - A_k\big\|_{\rm F}}
{(4r/3)^{1/2}(\epsilon_{rk}\vee \eps)} \le 1\bigg\}
\cr &\ge & \P\bigg\{\max_{j,k} 
\frac{\big\|\widehat a_{jk}^{\ico}\widehat a_{jk}^{\ico\top}  - a_{jk}a_{jk}^\top\big\|_{\rm S}}
{\epsilon_{jk}\vee \eps} \le 1\bigg\}
\cr & \ge& \P\big\{\Omega_0\big\} -  mrKe^{-2(d_1\wedge \sqrt{n})} 
\eel
within $m\ge m_\eps+3$ iterations, where $\epsilon_{jk} = C_{0,\alpha}R^{\ideal}_{jk,\phi_0}$,  
$m_\eps=\lceil \log(\log(\eps/\psi_0)/\log\rho)/\log 2\rceil$ for 
$(\eps_{r2}\vee \eps_0)\wedge \eps_{r3}\le\eps<\psi_0$ 
and $m_\eps=\lceil \log(\eps/\psi_0)/\log\rho\rceil$ for $\eps_{r2}\le \eps< \eps_0 \wedge \eps_{r3}$, 
with $\eps_0=C_{0,\alpha}r/n$. 
Moreover, \eqref{th-3-bd} holds within $m_{\eps_{r2}}+4$ iterations for $\eps=\eps_*\vee\sqrt{\eps_*\eps_0}$ where 
$\eps_*=C_{0,\alpha}(\lam_1/\lam_r)\prod_{k=2}^K\eps^2_{rk}$. 
In particular, if Algorithm \ref{algorithm:initial} (CPCA) is used to initialize Algorithm \ref{algorithm:projection} and 
$\psi_0$ is taken as the maximum of the right-hand side of \eqref{thm:initial2:eq1}, 
then \eqref{th-3-bd} holds with $\P\{\Omega_0\}\ge 1-e^{-t}$. 
\end{restatable}

In Theorem \ref{thm:projection2}, $\eps_{jk}$ can be viewed as statistical error and $\eps$ as computational error. It asserts that by iteratively 
projecting data (and thus the noise) to the direction $b_{j\ell}$ in mode-$\ell$ for all $\ell\neq k$, $(b_{1\ell},\ldots,b_{r\ell})=A_\ell(A_\ell^\top A_\ell)^{-1}$, Algorithm \ref{algorithm:projection} (ICO) effectively strengthens SNR from \eqref{SNR} to $r\lam_1/(\sigma^2d_k)$ in the estimation of $a_{jk}$ while quickly reduces the bias to below the level of stochastic error. 
As expected from the $\log\log(1/\epsilon)$ convergence in Theorems \ref{thm:noiseless} and \ref{thm:projection2}, the algorithm typically converges within very few steps in our practical implementations. 

Theorem \ref{thm:projection2} indicates that Algorithm \ref{algorithm:projection} converges linearly in its last phase with $\eps_{r2}\le \eps< \eps_0 \wedge \eps_{r3}$. However, if we treat the covariance tensor $T$ in model \eqref{model} as a general order $2K$ tensor and apply Algorithm \ref{algorithm:projection2}, high-order convergence can be also achieved in this last phase. The constant $\log 2$ in the definition of $m_\eps$ is conservative. In fact, by the proof of Theorem \ref{thm:projection2}, Algorithm \ref{algorithm:projection} converges in multiple phases beginning from order $2K-2$ convergence in its first phase. 

The right-hand side of \eqref{algorithm:projection} can be improved to $\P\{\Omega_0\} - rK e^{-2(d_1\wedge \sqrt{n})}$ if the constants in \eqref{alpha} are raised by a factor of at most order $K$ if we apply the probability calculation in the proof of Theorem \ref{thm:projection}. 
The Gaussian assumption can be replaced by sub-Gaussian in our analysis.

In Theorem \ref{thm:projection2}, $\psi_0$ is the required accuracy of the initial estimator. Given $\{C_0, r, \delta_{\max}, \lam_1/\lam_r\}$, 
the first two conditions in \eqref{C_0} hold when $\psi_0$ is sufficiently small, so that the third condition in \eqref{C_0} is a signal strength condition in terms of $R^{\ideal}_{rK,1}=\max_{j,\phi}R^{\ideal}_{jk,\phi}$. In view of the definition of $\alpha$ in \eqref{alpha}, 
condition \eqref{C_0} requires $r^{1/2}\psi_0$ be small, with an extra factor $r^{1/2}$ on the initial error in the estimation of individual basis vectors. This is a technical issue due to the need to invert the estimated $\Sigma_\ell=A_\ell^\top A_\ell$ in our analysis to construct the mode-$\ell$ projection in the ICO. In practice, if this issue is of concern, one may consider regularized inverse such as by adding a small constant to $\widehat \Sigma_\ell$ before computing the inverse or shrinking the singular values of $\widehat \Sigma_\ell$ as \cite{anandkumar2014guaranteed} suggested. If the right-hand side of \eqref{thm:initial2:eq1} is taken as $\psi_0$ for the CPCA initialization, condition \eqref{thm-projection2:eq1} can be reduced to an incoherence condition $r^{3/2}\delta\lesssim 1$ when $\lambda_1\asymp \lambda_r \asymp r\lam_{j,\pm}$ and $\sigma^2$ and $1/n$ are sufficiently small. 


When $\sqrt{r}(R^{\ideal}_{rK})^2 \lesssim R^{\ideal}_{jk}$, the statistical error $\epsilon_{jk}\lesssim R^{\ideal}_{jk}$. In the literature of tensor factor models with a Tucker structure \citep{chen2021statistical,han2020iterative}, the estimation of $a_{jk}$ may achieve faster convergence rate than $O_{\P}(n^{-1/2})$ when $\lam_j=w_j^2$ is sufficiently large. Similarly, 
\eqref{thm-projection2:eq2} may also converge faster than $O_{\P}(n^{-1/2})$. 

\begin{rmk}[Statistical Optimality]
The performance bound in \eqref{thm-projection2:eq2} is free of rank $r$. The rate $R^{\ideal}_{jk}$ matches the statistical lower bound of \cite{birnbaum2013minimax} and  \cite{han2020iterative} under specific rank one spiked covariance models respectively for matrix and tensor data. Therefore, under proper conditions, the proposed method (Algorithm \ref{algorithm:projection}) achieves the minimax optimal rate of convergence in the estimation of $a_{jk}$.
\end{rmk}

\subsection{General high order tensors} \label{section:extension}

In the noiseless case with $\Psi=0$ in \eqref{model2}, the extension of Theorem \ref{thm:noiseless} to Algorithms \ref{algorithm:initial2} and \ref{algorithm:projection2} is straightforward, which explicitly guarantees the high-order convergence of ICO with CPCA initialization. 
As in Proposition \ref{prop:delta} let $a_{jS}=\vec(\otimes_{k\in S}a_{jk})$, 
$A_{S} = (a_{1S},\ldots,a_{rS})$, $\Sigma_S=A_S^\top A_S$ and 
$\delta_S = \| \Sigma_S-I_r\|_{\rm S}$ for any nonempty subset $S$ of 
$[N]=\{1,\ldots,N\}$. 

\begin{restatable}{theorem}{TheoremNoiselessGen}
\label{thm:noiseless-gen} 
Suppose Algorithm  \ref{algorithm:initial2} (CPCA) is applied to the noiseless data $T^* = \sum_{j=1}^r \lam_j \otimes_{k=1}^{N}a_{jk}$ through the SVD of 
$\mat_S(T^*)$ for some nontrivial subset $S\subset [N]$. 
Let $\psi_0 = (\sqrt{2} + 4\lam_1/\lam_{\min,\pm})\delta$ with $\delta = \delta_S \vee \delta_{S^c}$, where $S^c=[N]\setminus S$.
Then, 
\bel{thm:noiseless-gen-0}
|\cpcalam_j -\lam_j| \le \sqrt{2} \delta\lam_1,\quad \big(\|\cpca_{jk}\cpcatop_{jk} - {a}_{jk} {a}_{jk}^\top\|_{\rm S}^2\big)\wedge (1/2)
\le \psi_0^2/2, 
\eel
for the resulting $\cpcalam_j$ and $\cpca_{jk}$. 
Suppose further that for $\delta_{\max}=\max_{k\le N}\delta_k$,
\bel{thm:noiseless-gen-1} 
3\max\big\{\delta_{\max},(\sqrt{r}+1)\psi_0\big\} \le 1,\ \  4\sqrt{r-1}(\lam_1/\lam_r)\psi_0^{N-2} 
\le \rho <1. 
\eel
Let $\gamma_N\in (2 -2/N,2)$ be the solution of $\gamma_N^N - 2\gamma_N^{N-1}+1=0$, e.g. $\gamma_3=1.618$, $\gamma_4=1.839$. 
If the resulting $\cpca_{jk}$ is used as the initialization of Algorithm \ref{algorithm:projection2} (ICO), then 
\begin{align*}
\max_{j\le r}\big(2 - 2\big|a_{jk}^\top \widehat a_{jk}^{(m)}\big| \big)^{1/2}
&\le \psi_{m,k} = \psi_0 \rho^{\gamma_N^{(m-1)N+k-1}}, \\
\max_{j\le r}\big| \widehat\lambda_j^{(m)}/\lambda_j -1 \big| & 
\le \textsum_{k=1}^N \psi_{m,k}+\rho\psi_{m,N},
\end{align*}
and $\max_{1\le k\le N} \psi_{m,k}\le \epsilon$ within 
$m = \lceil N^{-1}\{1+(\log \gamma_N)^{-1}\log(\log(\psi_0/\epsilon)/\log(1/\rho))\} \rceil$ iterations. 
\end{restatable}

\begin{rmk}
Condition \eqref{thm:noiseless-gen-1} specifies the required incoherence condition via $\delta$. 
Again, the constant factors 3 and 4 in the condition is not sharp, as \eqref{thm:noiseless-gen-1} is simplified from the following condition actually used in the proof,
\bel{thm:noiseless-gen-2} 
\frac{\sqrt{2(r-1)(1+\delta_{\max})}(\lam_1/\lam_r)\psi_0^{N-1} }
{\big((1 - \delta_{\max})^{1/2}-(\sqrt{r}+1)\psi_0\big)_+^{N-1} } \le \rho \psi_0 <\psi_0. 
\eel
As we have discussed in Remark \ref{rmk:initial}, such conditions guarantee the high-order contraction of the ICO and the resulting $\log\log(1/\epsilon)$ rate. 
\end{rmk}

Now consider statistical properties of Algorithms~\ref{algorithm:initial2} and \ref{algorithm:projection2} for general (asymmetric) tensors $T=\sum_{j=1}^r \lambda_j \otimes_{k=1}^N a_{jk} + \Psi$ in model  \eqref{model2}, where $a_{jk}\in \R^{d_k}$ are basis vectors with $\| a_{jk} \|_2=1$, and $\Psi$ is the noise tensor. 
Similar to the analysis of the spiked covariance tensor model given by \eqref{model1} and \eqref{model}, we assume 
for notational simplicity $\lambda_1>\lambda_2>\cdots >\lambda_r>0$.


\begin{restatable}{theorem}{theoremdenoisecpca}\label{thm:initial}
Let $T= \sum_{j=1}^r \lambda_j \otimes_{k=1}^N a_{jk}+\Psi$ as in \eqref{model2}. 
Suppose $\Psi\in\R^{d_1\times\cdots\times d_N}$ has i.i.d $N(0,\sigma^2)$ entries. Then, in an event with probability at least $1-e^{-2d_S-2(d/d_S)}$, Algorithm \ref{algorithm:initial2} (CPCA) gives the following bound in the estimation of the CP basis vectors $ a_{jk}$ , $1\le j\le r$, $1\le k\le N$,
\begin{equation}\label{thm:initial:eq1}
\| \cpca_{jk} \cpcatop_{jk}  - a_{jk} a_{jk}^\top \|_{\rm S} \le (1
+  2\sqrt{2}\lambda_1/\lam_{j,\pm})\delta +  6\sigma(\sqrt{d_S}+\sqrt{d/d_S})/\lam_{j,\pm} 
\end{equation}
where $\delta = \|A_S^\top A_S - I\|_{\rm S}\vee\|A_{S^c}^\top A_{S^c}-I_r\|_{\rm S}$ as in Theorem \ref{thm:noiseless-gen} 
and $\lam_{j,\pm}=\min(\lam_{j-1}-\lam_j,\lam_j-\lam_{j+1})$ are the eigengaps with $\lam_0=2\lam_1$ and $\lam_{r+1}=0$.    
\end{restatable}

The second term in \eqref{thm:initial:eq1}, representing the stochastic error, describes the required SNR for the CPCA. It is comparable to the SNR for tensor unfolding method in rank one symmetric case \citep{montanari2014statistical}, which is proved in \cite{brennan2020reducibility} to match an optimal computational lower bound under certain conditions. 
Moreover, the SNR condition here is weaker than the perturbation condition of the initialization in \cite{anandkumar2014guaranteed} when $\lambda_r/\lambda_{\min,\pm}=o(\sqrt{d_{\max}/\log(r)})$, which is typically satisfied for large $d_k$.

For simplicity, we assume below $d_1\le \cdots\le d_N$. Let 
\bel{R_jk*}
R^{*\ideal}_{jk} &=& \sigma \sqrt{d_k}/\lambda_j. 
\eel 
and for $\phi\ge 0$ define 
\bel{R_jk-ideal*}
R^{*\ideal}_{jk,\phi} = R^{*\ideal}_{jk} 
+ (\phi\wedge 1)\textsum_{\ell=1}^N R^{*\ideal}_{j\ell}. 
\eel 
For constants $\psi_0\in (0,1)$, define 
\bel{alpha2}
\alpha_* &=& \sqrt{1-\delta_{\max}}-(r^{1/2}+1)\psi_0,
\cr \rho^* &=& 6\alpha_*^{1-N}\sqrt{r-1}(\lam_1/\lam_r)\psi_0^{N-2},
\\ \nonumber \phi_0^* &=& (N-1)\alpha_*^{-1}\sqrt{2r}R^{*\ideal}_{rN,1}. 
\eel

\begin{restatable}{theorem}{theoremdenoiseico}
\label{thm:projection} 
Let data $T$ be as in Theorem \ref{thm:initial} and $\Omega_0=\{\max_{j,k}
(2-2|a_{jk}^\top\widehat a^{(0)}_{jk}|)^{1/2}\le\psi_0\}$ for any initial estimates $\widehat a^{(0)}_{jk}$. 
Let $\mathscr P_\pm$ be as in \eqref{thm-projection2:eq2}. 
Suppose 
\bel{thm-projection:eq1}
\alpha_* > 0,\ \rho^*<1,\ 
6\alpha_*^{1-N}R^{*\ideal}_{rK,1}\le \psi_0<1, 
\eel 
with the quantities defined in \eqref{R_jk*}, \eqref{R_jk-ideal*} and \eqref{alpha2}.
Then, in an event with probability at least $\P\{\Omega_0\} - e^{-d_N} - \sum_{k=1}^N e^{-d_k} $, Algorithm \ref{algorithm:projection2} (ICO) provides 
\begin{align}
|\widehat \lambda_j^{\ico}/\lambda_j - 1| & \le  \epsilon_{jN}^*\vee \eps,   \label{thm-projection:eq3} \\
\|\widehat a_{jk}^{\ico}\widehat a_{jk}^{\ico\top}  - a_{jk} a_{jk}^\top \|_{\rm S} &\le \epsilon_{jk}^*\vee \eps, \label{thm-projection:eq2} \\
\min_{\Pi_r\in\mathscr P_\pm} \|\widehat A_k^{\ico} \Pi_r-A_k\|_{\rm F} & \le {r^{1/2}(\epsilon_{rk}^* \vee \eps)}, 
\label{thm-projection:eq2b} 
\end{align}
simultaneously for all $1\le j\le r$ and $1\le k\le N$, within $m\ge m_\eps+3$ iterations, 
where $\epsilon_{jk}^* = 6\alpha_*^{N-1}R^{*\ideal}_{jk,\phi_0^*}$  and $m_\eps=\lceil \log(\log(\eps/\psi_0)/\log\rho^*)/\log 2\rceil$ for 
$\eps_{r2}^*\le\eps<\psi_0$. Moreover, \eqref{thm-projection:eq3}, \eqref{thm-projection:eq2} and \eqref{thm-projection:eq2b} 
hold in the same event within $m_{\eps_{r2}}+4$ iterations for $\eps=6\alpha_*^{1-N}\sqrt{r-1}(\lam_1/\lam_r)\prod_{k=2}^N\eps^*_{rk}$. 
If Algorithm~\ref{algorithm:initial2} (CPCA) is used as initialization, then
$\P\{\Omega_0\}\ge 1-\sum_{k=1}^N e^{-2d_k}$ for $\psi_0=6[\lambda_1 \delta+ \sigma(\sqrt{d_S}+\sqrt{d/d_S}) ]/\lambda_{\min,\pm}$.
\end{restatable}

We briefly discuss the conditions and conclusions of Theorem \ref{thm:projection} as the details are parallel to the discussions below Theorem \ref{thm:projection2}. In Theorem \ref{thm:projection}, $\eps_{jk}^*$ can be viewed as statistical error and $\eps$ as computational error. When $\sqrt{r}(R^{*\ideal}_{rN})^2 \lesssim R^{*\ideal}_{jk}$, the statistical error $\epsilon_{jk}^*\lesssim R^{*\ideal}_{jk}$ is rate minimax.
Condition \eqref{thm-projection:eq1} specifies the required strength of the signal and accuracy of the initialization. 
It guarantees that the ICO has a high-order error contraction effect in the iteration. Ignoring the perturbation error and assuming $\lambda_1\asymp \lambda_r$, it can be reduced to an incoherence condition $r^{3/2}\delta\lesssim 1$ when CPCA is used as initialization. In addition, the performance bound in \eqref{thm-projection:eq2} is free of CP rank $r$ and matches the statistical lower bound of \cite{zhang2018tensor} for rank one noisy tensor model. It shows the optimality of the convergence rate of the proposed ICO (Algorithm \ref{algorithm:projection2}).

\subsection{Comparison with existing theoretical results}

In this subsection, we compare the proposed Algorithms \ref{algorithm:initial2} and \ref{algorithm:projection2} with existing theories of tensor decomposition methods. Several important implications are provided, and comparisons in incoherence condition, iteration complexity, and statistical error bounds are summarized in Table \ref{tab:compare}. For simplicity, the following discussion assumes model \eqref{model2} with $\lambda_1\asymp \lambda_r$, $\lambda_{\min,\pm}\asymp \lambda_r/r$ and Gaussian noise $\Psi$.  

{\it Super-linear convergence}. In the absence of noise, the proposed algorithm attains $\epsilon$ accuracy within $O(\log\log(1/\epsilon))$ iterations. In the noisy setting, the algorithm reaches an ideal statistical accuracy within an iterated logarithmic number of iterations. 
The perturbation bounds in Propositions \ref{prop-ICO} and \ref{prop-ICO-gen} explicitly give the order of convergence for ICO: Up to some scaling constants, the error in the estimation of $a_{jk}$ in each step is bounded by the product of the up-to-date errors in all other modes. As in the analysis of Nesterov's acceleration of gradient descent, this multiplicative nature of error propagation leads to a $\log\log(1/\epsilon)$ convergence rate. In alternating least squares \citep{anandkumar2014guaranteed} and HOOI, the error propagation is linear due to tensor unfolding so that the convergence rate is of the order $\log(1/\epsilon)$. Still, in certain problems where computationally feasible initialization leads to very high signal-to-noise ratio, one-step least squares or HOOI update would reduce the error to the level of statistical efficiency \citep{zhang2018tensor, han2020iterative, luo2021low}. 


{\it Statistical accuracy}. 
While our theoretical analysis is focused on the estimation of individual basis vectors $a_{jk}$, our results have direct implications on the estimation under different loss functions or of related functions beyond the explicite statements of Theorems \ref{thm:projection}. 
For example, for the estimation of the entire tensor $T^*=\E[T]$ in model \eqref{model2}, Theorems \ref{thm:projection} directly yields the 
Frobenius error bound
\bes
\big\|\widehat T - T^*\big\|_{\rm F}
\lesssim K\lam_1r^{1/2}(\eps^*_{rK}\vee\eps). 
\ees 
Compared with \cite{anandkumar2014guaranteed}, Theorems \ref{thm:projection} provide comparable or sharper error bounds under their conditions. The error bound of the CP decomposition algorithms in \cite{anandkumar2014guaranteed} is $\|\widehat{A}_{k}\Pi_r - A_{k}\|_F\le C\sqrt{r}\|\Psi\|_{*}/\lambda_r$, where $\|\Psi\|_{*}$ is the tensor spectrum norm, with $\|\Psi\|_{*}\asymp \sigma\sqrt{d_1+\cdots+d_N}$ in the Gaussian case. In comparison, Theorem \ref{thm:projection} provides $\|\widehat A_k^{\ico} \Pi_r-A_k\|_{\rm F} \le C\sigma\sqrt{d_kr}/\lambda_r$ for ICO with CPCA initialization, matching the statistical lower bound of \cite{zhang2018tensor}.

{\it Incoherence condition for initialization}. Existing initialization approaches \citep{anandkumar2014guaranteed, cai2021nonconvex, cai2021subspace} focus on randomized projection in each tensor mode simultaneously to reduce the original data tensor to matrices of effective rank near 1, followed by matrix SVD to obtain rough estimates of CP basis $a_{jk}$, one from each ``good" projection selected by clustering or some other methods. When the basis vectors $a_{jk},j\le r$, are nearly orthogonal to each other, the leading singular vector of the selected projected matrix is expected to be reasonably close to one of the CP components, approximating $a_{jk}$ for the same $j$ in all mode $k$. As the possible directions of randomized projection increase rapidly with dimension $d_k$, the incoherence condition must decrease with $d_k$ to allow a moderate restart number (i.e. required number of randomized projections) to capture a single CP component. Therefore, the existing incoherence condition in individual tensor modes is hard to avoid in such approaches. Our approach is fundamentally different. As discussed in Section \ref{section:estimation}, the CPCA is designed to take advantage of the multiplicative nature of the higher order coherence.


{\it Tucker models}. There exists a large body of work that handles low-rank tensor Tucker decomposition, including \cite{liu2012, zhang2018tensor, xia2019polynomial,zhang2019optimal, tong2022scaling, han2022optimal}.
For example, \cite{zhang2018tensor} studied HOOI and provides rate optimal statistical bound under Gaussian noise tensor. 
In the rank-1 case where the CP and Tucker representations are identical, our performance bound in Theorem \ref{thm:projection} is equivalent to theirs. Our results and theirs are also in agreement for the estimation of the projection to 
the column space of CP basis $A_k=(a_{jk},j\le r)$. The theoretical tool for the analyses of HOOI and our ICO share a similar spirit as both involve projections in the iteration. However, there are several major differences between the statistical analyses in the Tucker and CP models.  Moreover, the projection in ICO is very different from previous proposals as discussed in Subsection \ref{subsec:errors}, thus requiring much more sophisticated analysis. In addition, we develop sharp and useful tensor perturbation bounds in our analysis.

\begin{table}[ht]
\centering
\begin{tabular}{c|c|c|cc}\hline
Algorithms & Incoherence  &  Iteration complexity & Error (Noisy) \\  \hline
robust tensor power method & \multirow{2}{*}{$0$}  & \multirow{2}{*}{$\log(r)+\log\log(1/\epsilon)$} & \multirow{2}{*}{$\|\Psi\|_{*}/\lambda_r$} \\
\cite{anandkumar2014tensor} & & &  \\ \hline
Two-mode HOSVD & \multirow{2}{*}{$0$} & \multirow{2}{*}{n/a} & \multirow{2}{*}{$\|\Psi\|_{*}/\lambda_r$} \\
\cite{wang2017tensor} & & &  \\ \hline
randomized projection + power update & \multirow{2}{*}{$\vartheta_{\max}\lesssim 1/\sqrt{d_1}$} & \multirow{2}{*}{$\log(1/\epsilon)$} & \multirow{2}{*}{$\|\Psi\|_{*}/\lambda_r$} \\
+ CD \cite{anandkumar2014guaranteed} & & &  \\ \hline
spectral method + (vanilla) GD & \multirow{2}{*}{$\vartheta_{\max}\lesssim 1/\sqrt{d_1}$} & \multirow{2}{*}{$\log(1/\epsilon)$} & \multirow{2}{*}{$\sigma\sqrt{d_1}/\lambda_r$} \\
\cite{cai2021nonconvex} & & &  \\ \hline
CPCA + ICO & \multirow{2}{*}{$\delta\lesssim 1/r^{3/2}$} & \multirow{2}{*}{$\log\log(1/\epsilon)$} & \multirow{2}{*}{$\sigma\sqrt{d_1}/\lambda_r$} \\
(this paper) & & &  \\ \hline
\end{tabular}
\caption{Comparison with previous theories for existing CP decomposition methods when $d_1\asymp...\asymp d_N\asymp d^{1/N},\lambda_1\asymp \lambda_r$ (neglecting logarithmic factors). Here CD and GD are coordinate descent and gradient descent, respectively.}
\label{tab:compare}
\end{table}

\section{Numerical experiments} \label{section:simulation}

In this section, we provide some synthetic experiments to compare the performance of the proposed methods, CPCA initialization followed by ICO iterations as in Algorithms \ref{algorithm:initial}-\ref{algorithm:projection2} (Alg1+Alg2 for covariance tensor, Alg3+Alg4 otherwise), with the modified rank one alternating least squares (ALS) \citep{anandkumar2014guaranteed}, orthogonalized alternating least squares (OALS) \citep{sharan2017orthogonalized}, and higher order SVD (HOSVD). In our simulations, both ALS and OALS use the initialization method proposed in \cite{anandkumar2014guaranteed} and used in  \cite{sun2017provable} and \cite{hao2020sparse}, which applies power and clustering methods to random basis vectors  
and uses the resulting centroids as initialization. HOSVD, widely used in CP decomposition and tensor completion \citep{han2021guaranteed,cai2021nonconvex,cai2021subspace}, can be viewed as a baseline initialization method. 
To better understand CPCA, we also present the results of the method (Alg1 or Alg3) without further improvements 
and its performance as the initialization of ALS and OALS updates (Alg1-ALS, Alg3-ALS, Alg1-OALS, Alg3-OALS). The estimation error is given by $\max_{j,k}\|\widehat a_{jk}\widehat a_{jk}^\top  - a_{jk} a_{jk}^\top \|_{\rm S}$. The CP basis vectors $a_{jk}$ are first generated independently and uniformly at random from the $d_k$ dimensional unit spherical shell, and then linearly adjusted to satisfy $\max_{i\neq j}|a_{ik}^\top a_{jk}|=10^{-1/2}$ for order 4 tensors in models \eqref{model} and \eqref{model2}.

\begin{figure}[htbp]
\centering
\includegraphics[width=5in]{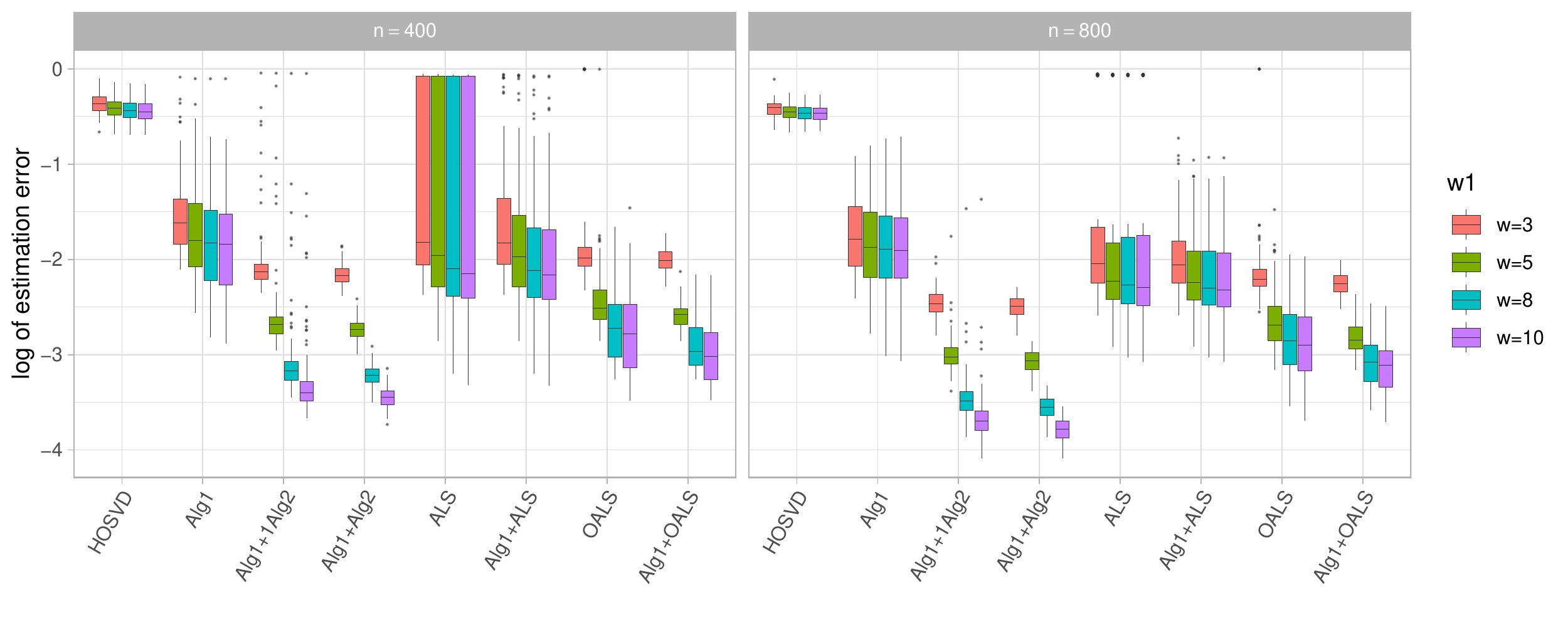}
\caption{Boxplots of the logarithm of the estimation error over 100 replications under the spiked covariance tensor setting with $K=2$ and $\lambda_1=w_1^2$. The two panels correspond to sample sizes $n=400,800$ respectively. The proposed algorithms are labeled as Alg1 (CPCA) and Alg2 (ICO).} \label{figure:factor_matrix}
\end{figure}

We first study the finite sample performance with spiked covariance tensors  \eqref{model1}. We set $w_{\max}/w_{\min}=1.25, d_1=d_2=20, r=3, n=400,800$, $K=2$, $w_{\max}=3,5,8,10$, so that the covariance tensor is of the order $4=2K$. Figure \ref{figure:factor_matrix} depicts the boxplots of the logarithm of the estimation errors over 100 replicates. In the plot, Alg1+1Alg2 is the one-step ICO estimator after the CPCA initialization. Overall, our method Alg1+Alg2 outperforms all the other methods in all cases. 
The ICO (Alg2) converges in very few steps, although the number of steps is not reported here. Besides, the one step estimator Alg1+1Alg2 significantly improves over the CPCA initialization (Alg1), and is very close to the final estimator Alg1+Alg2. HOSVD performs much worse than the CPCA initialization (Alg1), probably due to the benefit of multiplicative higher order coherence of the CPCA. 
The comparisons of ALS against the hybrid Alg1+ALS and OALS against the hybrid Ag1+OALS demonstrate the CPCA as a better method than clustering or other randomized screening methods for initialization, although the CPCA initialization (Alg1) standing alone  
may perform worse than iterative methods (slightly so compared with ALS and more clearly so with OALS). In fact the 
hybrid methods with the CPCA initialization improve the original randomized initialized ALS and OALS significantly, especially when the signal strength $w_{\max}$ is large.

\begin{figure}[htbp]
\centering
\includegraphics[width=5in]{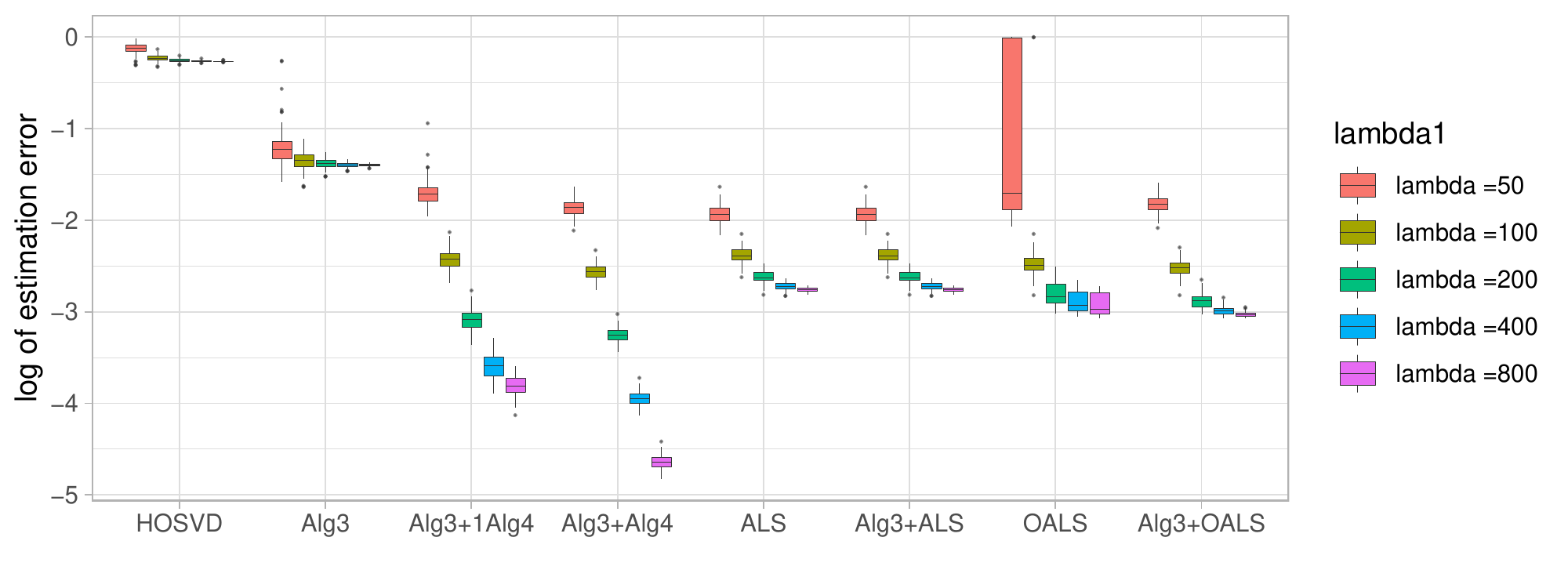}
\caption{Boxplots of the logarithm of the estimation error over 100 replications under the low-rank tensor de-noising setting with $N=4$. The proposed algorithms are labeled as Alg3 (CPCA) and Alg4 (ICO).} \label{figure:tensor4}
\end{figure}

We also explore our methods under the low-rank tensor de-noising setting \eqref{model2}. We consider a 4-way tensor with $d_1=d_2=d_3=d_4=20$, $\lambda_{\max}/\lambda_{\min}=1.25, r=3$, and $\lambda_{\max}=50,100,200,400,800$. Figure \ref{figure:tensor4} quantifies the performance of different algorithms in terms of the logarithm of the estimation errors. Except for $\lambda_{\max}=50$, Alg3+Alg4 is superior to all the other algorithms. When $\lambda_{\max}=50$, ALS and Alg3+ALS are slightly better than Alg3+Alg4 and Alg3+OALS. Again, HOSVD underperforms the CPCA initialization (Alg3). Figure \ref{figure:tensor4} also shows the benefits of one step estimator Alg3+1Alg4. Although Alg3+ALS has similar behavior as ALS in this setting, we do not need to generate a large number of random initialization in the hybrid method Alg3+ALS.


\begin{figure}[htbp]
\centering
\includegraphics[width=5in]{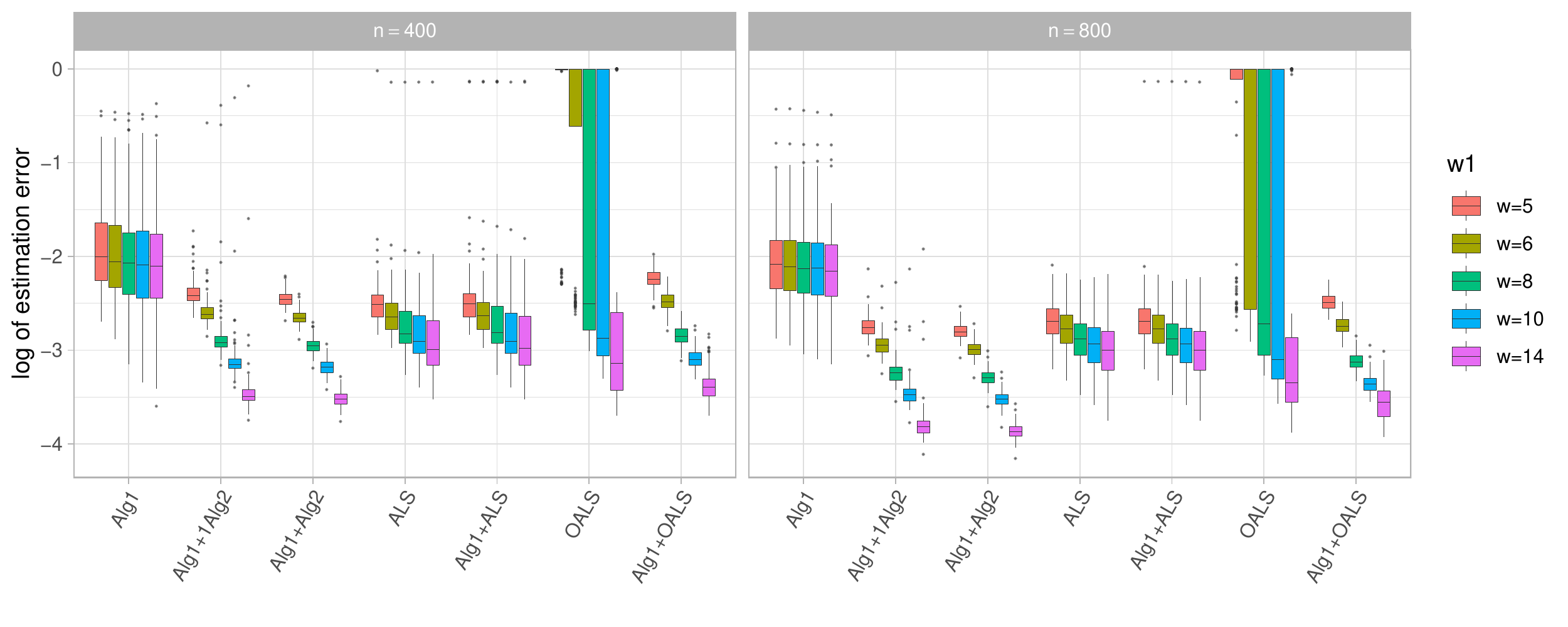}
\caption{Boxplots of the logarithm of the estimation error over 100 replications under the spiked covariance tensor setting with $K=3$ and $\lambda_1=w_1^2$. Two panels correspond to two sample sizes $n=400,800$. The proposed algorithms are labeled as Alg1 (CPCA) and Alg2 (ICO).} \label{figure:factor_tensor3}
\end{figure}

\begin{figure}[htbp]
\centering
\includegraphics[width=5in]{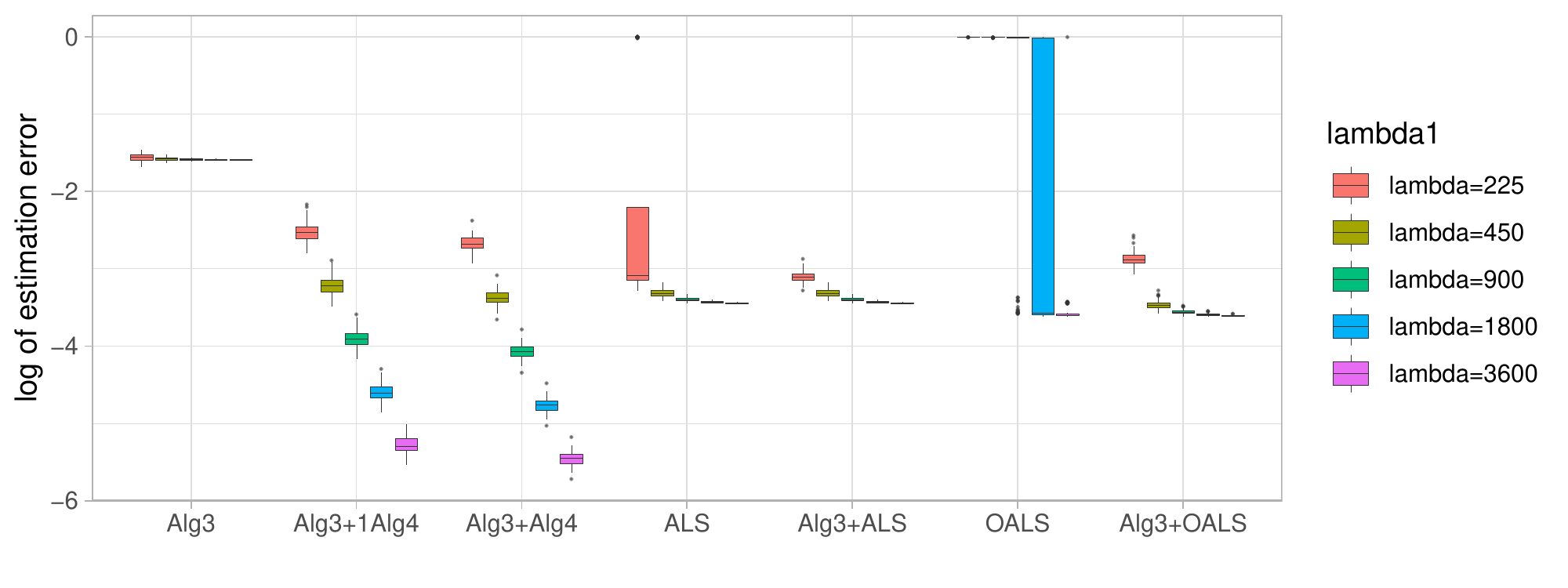}
\caption{Boxplots of the logarithm of the estimation error over 100 replications under the low-rank tensor denoising setting with $N=6$. The proposed algorithms are labeled as Alg3 (CPCA) and Alg4 (ICO).} \label{figure:tensor6}
\end{figure}

Next, we consider two additional cases of order 6 tensors in models \eqref{model} and \eqref{model2} 
with basis vectors satisfying $\max_{i\neq j}|a_{ik}^\top a_{jk}|^3=0.1$. 
In a spiked covariance tensor setting \eqref{model1}, we set $w_{\max}/w_{\min}=1.25, r=3, d_1=d_2=d_3=20, n=400,800, K=3$, and $w_{\max}=5,6,8,10,14$. In the low-rank tensor denoising setting \eqref{model2}, we set $d_k=20, 1\le k\le 6, \lambda_{\max}/\lambda_{\min}=1.25, r=3$, and $\lambda_{\max}=225,450,900,1800,3600$. We omit HOSVD as it is always much worse than the CPCA initialization.
The results are similar to order 4 tensors. From Figure \ref{figure:factor_tensor3}, Alg1+Alg2 are the best one in all cases. The advantages are more obvious when $w_{\max}$ is large. OALS with randomized initialization has a great deal of variabilities, which can be significantly improved by the CPCA initialization (Alg1+OALS). Though ALS and Alg1+ALS have almost the same performance, Alg1+ALS does not require a large number of random initialization. The results in the tensor denoise setting, reported in Figure \ref{figure:tensor6}, are similar to those in the spiked covariance tensor model setting in Figure \ref{figure:factor_tensor3}, except the case $\lambda_{\max}=225,450$. Alg3+ALS fares better than the other approaches for $\lambda_{\max}=225$, while Alg3+OALS is the best for $\lambda_{\max}=450$. Although the proposed algorithms do not always outperform ALS and OALS, they underperform only slightly and in very few simulation configurations and they are faster and easier to implement. Moreover, the simulation results demonstrate that the CPCA initialization is superior to the randomized initialization with ALS and OALS.

To evaluate the 
computational cost of  
different initialization methods, we also report the run time of the CPCA initialization, HOSVD, and the randomized initialization in \cite{anandkumar2014guaranteed} (ALS-init) 
under a spiked covariance tensor setting \eqref{model1}. We set $w_{\max}/w_{\min}=1.25, r=3, n=800$, $K=2$, $w_{\max}=10$, and vary $d_1=d_2=20,30,40,50,60$. From Table \ref{tab:init}, it can be seen that ALS-init requires much longer run time for each simulation than the other methods. The reason may be that ALS-init needs a large number of restarts to recover all the CP basis. 
Meanwhile, HOSVD has significantly shorter 
run time than the CPCA, and the ratio of the costs seems stable  as the dimension increases. 
Thus, the far superior performance of the CPCA justifies its (still manageable) computational costs compared with HOSVD.

\begin{table}[http]
\centering
\begin{tabular}{c|c|c|c|c|c}\hline
Algorithms & $d_1=d_2=20$  &  $d_1=d_2=30$ & $d_1=d_2=40$ & $d_1=d_2=50$ & $d_1=d_2=60$ \\  \hline
HOSVD & $0.12_{(0.02)}$ & $1.18_{(0.28)}$ & $4.62_{(0.89)}$ & $12.58_{(2.27)}$ & $27.98_{(5.37)}$\\  \hline
ALS-init & $6.97_{(1.93)}$ & $38.95_{(8.21)}$ & $133.20_{(25.78)}$ & $332.68_{(81.74)}$ & $726.75_{(177.53)}$\\  \hline
Alg1 (CPCA) & $0.19_{(0.02)}$ & $2.07_{(0.28)}$ & $10.10_{(1.13)}$ & $33.71_{(2.96)}$ & $94.55_{(9.41)}$\\ \hline
\end{tabular}
\caption{Run time for different initialization methods over 100 replications under the spiked covariance tensor setting with $K=2$. Here run time is the mean and standard deviations of the run time in seconds. ALS-init uses 30 restart numbers.}
\label{tab:init}
\end{table}

In summary, the proposed Algorithms \ref{algorithm:initial}-\ref{algorithm:projection2} are more accurate than existing methods in the simulation experiments in general. Algorithms \ref{algorithm:initial} and \ref{algorithm:initial2} can also be a superior initialization to plug in existing algorithms, and is faster and much simpler to implement than randomized initializations. 
It is worth noting that in the case of order 6 
tensors where the incoherence $\max_{i\neq j}|a_{ik}^\top a_{jk}| = 10^{-1/3}$ is larger, both ALS and OALS perform poorly, while the proposed methods still work well.

\section{Final remark}
In this paper, we propose new initialization (CPCA) and refinement (ICO) algorithms for tensor CP decomposition of high dimensional non-orthogonal spike tensors. Our methods tolerate a higher level of coherence among the basis vectors ($a_{jk}$), and achieve faster computational convergence rate and sharper statistical error bounds, compared with existing methods.
The proposed methods are applicable to a broad class of structured tensors, including the spiked covariance tensors \eqref{model} and general noisy high order tensors \eqref{model2}. In particular, our proposed algorithms show
stable convergence and exhibit pronounced advantage especially as the order of the tensor increases. Numerical studies display empirically favorable performance of the proposed methods.


\section{Analysis in the noiseless case: matrix and tensor perturbation bounds}\label{AppendixB}

This section provides the analysis of the CPCA and ICO algorithms in the noiseless case with $\Psi=0$ in models \eqref{model} and \eqref{model2}. The results in this section are dimension free in the sense that their conditions and conclusions depend only on the angles among the basis vectors and their estimates and the principle angles among spaces, not on $d_k$. 
We first present the proofs of Propositions \ref{prop:delta}, 
\ref{lemma-transform}, 
\ref{prop-rank-1-approx} and 
\ref{prop-ICO}. 
These propositions provide a road map of the proof of Theorem~\ref{thm:noiseless}, which is to follow, and some general techniques to study model \eqref{model}. Then, we present the proofs of Propositions \ref{lemma-transform-ext} and \ref{prop-ICO-gen}. Propositions \ref{prop:delta}, \ref{lemma-transform-ext}, 
\ref{prop-rank-1-approx} and  \ref{prop-ICO-gen} provide a road map of the proof of Theorem~\ref{thm:noiseless-gen} at the end of this section and some general techniques to study model \eqref{model2}. For readers' convenience, we restate the propositions and theorems before their proofs. 

\PropositionDelta*

\begin{proof}[\bf Proof of Proposition \ref{prop:delta}]
For notational simplicity, we only prove the case $S=[K]$, as the extension to general $S$ is straightforward. Recall that $\delta = \|A^\top A-I_r\|_{\rm S}$ and $\delta_k = \|A_k^\top A_k-I_r\|_{\rm S}$. 
Because $A^\top A= (A_1^\top A_1)\circ \cdots \circ(A_K^\top A_K)$ is 
the Hadamard product of correlation matrices, the spectrum of $A^\top A$ is 
contained inside the spectrum limits of $A_k^\top A_k$ for each $k$, so that 
\begin{align*}
\delta\le\min_{1\le k\le K}\delta_k. 
\end{align*}
Because $A^\top A -I_r$ is symmetric, its spectrum norm is bounded by its $\ell_1$ norm, 
\begin{align*}
\delta \le \max_{j\le r} \textsum_{i\neq j}|a_i^\top  a_j| \le (r-1)\vartheta \le (r-1) \prod_{k=1}^K \vartheta_k 
\end{align*}
due to $|a_i^\top  a_j| =\prod_{k=1}^K|a_{ik}^\top a_{jk}|=\prod_{k=1}^K|\sigma_{ij,k}|$. 
Moreover, for any $j\le r$ and $1\le k_1<k_2\le K$, 
\bes
\sum_{i\neq j}\prod_{k=1}^K |\sigma_{ij,k}| 
&\le& \sum_{i\neq j} |\sigma_{ij,k_1}\sigma_{ij,k_2}| 
\,\max_{i\neq j}\prod_{k\neq k_1,k\neq k_2}|\sigma_{ij,k}|
\cr &\le& \Big(\textprod_{k=1}^K \eta_{jk}\Big)r^{-(K-2)/2}
\,\max_{i\neq j}\prod_{k\neq k_1,k\neq k_2}\sqrt{r}|\sigma_{ij,k}|/\eta_{jk} ,
\ees
as $\eta_{jk} = (\textsum_{i\neq j}\sigma_{ij,k}^2)^{1/2}$.  
The proof is complete as $k_1$ and $k_2$ are arbitrary.
\end{proof}

\PropositionTrans*

\begin{proof}[\bf Proof of Proposition \ref{lemma-transform}]
An extension of Proposition \ref{lemma-transform}, Proposition \ref{lemma-transform-ext}, 
is proved later.
\end{proof}

\PropositionRankOneApprox*
\begin{proof}[\bf Proof of Proposition \ref{prop-rank-1-approx}.] 
Let $\sum_{j=1}^r \sigma_j u_j v_j^\top$ be the SVD of $M$ with singular values 
$\sigma_1\ge \ldots \ge \sigma_r$ where $r$ is the rank of $M$. 
Because $\vec(u_jv_j^\top)$ are orthonormal in $\R^{d_1d_2}$, 
\bes
\vec(M)^\top \vec({a} {b}^\top) 
= {a}^\top M {b} = \sum_{j=1}^{r} \sigma_j (u_j^\top {a}) (v_j^\top {b})
\ees
with $\sum_{j=1}^{r} \sigma_j^2=\|M\|_{\rm F}^2=1$, 
$\sum_{j=1}^{r} (u_j^\top {a})^2 \le \|{a}\|_2^2 = 1$ and 
$\sum_{j=1}^{r}(v_j^\top {b})^2 \le \|{b}\|_2^2 = 1$. 
Because $\sigma_1\ge \cdots\ge \sigma_r$,
\bes
\big| {a}^\top M {b} \big|\le \sigma_1
\bigg(\sum_{j=1}^{r} (u_j^\top {a})^2\bigg)^{1/2}
\bigg(\sum_{j=1}^{r}(v_j^\top {b})^2\bigg)^{1/2} = \sigma_1
\ees
Similarly, by Cauchy-Schwarz, 
\bel{pf-prop-rank-1-approx-1}
\big| {a}^\top M {b}\big|^2
\le \sum_{j=1}^{r} \sigma_j^2(u_j^\top {a})^2
\le \sigma_1^2(u_1^\top {a})^2
+ \big(1 -\sigma_1^2\big)\big(1 - (u_1^\top {a})^2\big). 
\eel
When $(u_1^\top {a})^2\ge 1/2$, 
the maximum on the right-hand side above is achieved at $\sigma_1^2=1$, 
so that $\big| {a}^\top M {b}\big|^2\le(u_1^\top {a})^2$; 
Otherwise, the right-hand side of \eqref{pf-prop-rank-1-approx-1} 
is maximized at $\sigma_1^2=\big| {a}^\top M {b}\big|^2$, 
so that $\big| {a}^\top M {b} \big|^2
\le 1 - \big| {a}^\top M {b} \big|^2$. Thus, 
$\big|{a}^\top M {b}\big|^2 > 1/2$ implies 
$\big|{a}^\top M {b} \big|^2\le(u_1^\top {a})^2$. 
By \eqref{loss}, this is equivalent to \eqref{prop-rank-1-approx-1}. 
\end{proof}

\PropositionICO*

\begin{proof}[\bf Proof of Proposition \ref{prop-ICO}] 
For any diagonal matrix $D$ with $D^2=I_r$, $\widetilde A_\ell D((\widetilde A_\ell D)^\top \widetilde A_\ell D)^{-1} = \widetilde A_\ell(\widetilde A_\ell^\top \widetilde A_\ell)^{-1}D$. 
Thus, because $\widetilde T^*_{jk}$ does not depend on the signs of $\widetilde b_{j\ell} =\widetilde b_{j,\ell+K}$, we assume without loss of generality that $\widetilde a_{i\ell}^\top a_{i\ell}\ge 0$ for all $i$ and $\ell$.
Let $\widetilde \Sigma_\ell = \widetilde A_\ell^\top \widetilde A_\ell$. 
Assume without loss of generality that $r\widetilde\psi_\ell^2/(1-1/(4r))\le 1$, so that 
$2(1-(1-\widetilde\psi_\ell^2)^{1/2}) \le \widetilde\psi_\ell^2/(1-1/(4r))$. Consequently, 
\bes
\|\widetilde A_\ell -A_\ell\|_{\rm S}^2 
\le r \max_{j\le r}\big\|\widetilde a_{j\ell} - a_{j\ell}\big\|_2^2
= 2r(1-(1-\widetilde\psi_\ell^2)^{1/2}) 
\le  r\widetilde\psi_\ell^2/(1-1/(4r)) 
\ees
As $\widetilde b_{j\ell} = \widetilde A_\ell(\widetilde A_\ell^\top\widetilde A_\ell)^{-1}e_j$, 
$\|\widetilde b_{j\ell}\|_2=\{e_j^\top(\widetilde A_\ell^\top\widetilde A_\ell)^{-1}e_j\}^{1/2} 
\le \max_{\|u\|_2=1}\|\widetilde A_\ell u\|_2^{-1}$, so that 
\bes
\max_{i\le r}\big\|\widetilde a_{i\ell} - a_{i\ell}\big\|_2\|\widetilde b_{j\ell}\|_2 
\le \big(\widetilde\psi_\ell/\sqrt{1-1/(4r)}\big)
\big/\big(\sqrt{1-\delta_\ell}-r^{1/2}\widetilde \psi_\ell /\sqrt{1-1/(4r)}\big)_+ = \widetilde \phi_\ell. 
\ees 
Let
$w_{ij,\ell} = a_{i\ell}^\top \widetilde b_{j\ell}/a_{j\ell}^\top \widetilde b_{j\ell}$, 
$v_{i,jk} = (\lam_i/\lam_j)\prod_{\ell\in[K]\setminus \{k\}} w_{ij,\ell}^2$, 
$v_{jk}\in\R^r$ be the vector with elements $v_{i,jk}$, and 
$\widetilde \lam_j = \lam_j\prod_{\ell \in [K]\setminus \{k\}}(a_{j\ell}^\top\widetilde b_{j\ell})^2$. 
As $\widetilde a_{i\ell}^\top \widetilde b_{j\ell} = I_{\{i=j\}}$, 
for $i\neq j$, 
\bes
|w_{ij,\ell}| = \frac{|( a_{i\ell} - \widetilde a_{i\ell})^\top \widetilde b_{j\ell}| }
{|1 + (a_{j\ell} - \widetilde a_{j\ell})^\top \widetilde b_{j\ell}|}
\le \frac{\widetilde\phi_\ell}{1-\widetilde\phi_\ell},\quad  
\big|v_{i,jk}\big| \le (\lam_1/\lam_j)\bigg(\prod_{\ell\in [K]\setminus \{k\}} \frac{\widetilde \phi_\ell}{1-\widetilde\phi_\ell} \bigg)^2. 
\ees
As $\widetilde  T_{jk}^*/\widetilde\lam_j = \sum_{i=1}^r a_{ik}a_{ik}^\top v_{i,jk}$ and eigenvectors do not depend in scaling, 
\begin{align*}
\|a_{jk} a_{jk}^\top -\widetilde a_{jk}^* \widetilde a_{jk}^{*\top} \|_{\rm S} \le 
2 \| \textsum_{i\ne j}^r a_{ik} a_{ik} v_{i,jk} \|_{\rm S}
\le 2 \|A_k\|_{\rm S}^2 \max_{i\neq j} v_{i,jk}
\end{align*}
by Wedin's theorem \cite{wedin1972}. The conclusion follows.
\end{proof}

\TheoremNoiseless*

\begin{proof}[\bf Proof of Theorem \ref{thm:noiseless}] 
Let $U=(u_1,\ldots,u_r)$ be the orthonormal matrix corresponding to $A=(a_1,\ldots,a_r)$ as in Proposition~\ref{lemma-transform} where $a_j=\vec(\otimes_{k=1}^Ka_{jk})$. Let $\text{mat}_{[K]}(T^*) = \textsum_{j=1}^r \cpcalam_j \widehat u_j \widehat u_j^\top$ be the eigenvalue decomposition as in \eqref{uhat_j}. By Proposition~\ref{lemma-transform} and Wedin's perturbation theorem, 
\bes
\|\widehat u_j \widehat u_j^\top - a_ja_j^\top\|_{\rm S} \le \|u_ju_j^\top - a_ja_j^\top\|_{\rm S} + \|\widehat u_j \widehat u_j^\top - u_ju_j^\top\|_{\rm S} \le \delta+2(\lam_1\delta)/\lam_{j,\pm}\le\psi_0 
\ees
and $|\cpcalam_j -\lam_j| \le \delta\lam_1$. 
Thus, \eqref{noiseless-bd} follows from Proposition~\ref{prop-rank-1-approx}. Moreover, under \eqref{thm:noiseless-2} we have $\psi_0<1/(\sqrt{r}+1)\le 1/2$, so that \eqref{noiseless-bd} yields $\max_{j\le r}\big\|\cpca_{j\ell}\cpcatop_{j\ell} - a_{j\ell}a_{j\ell}^\top \big\|_{\rm S}\le \psi_0$.
Now define 
$\psi_{m,k}'=\max_{j\le r}\big\|\widehat a_{jk}^{(m)}\widehat a_{jk}^{(m)\top} - a_{jk}a_{jk}^\top \big\|_{\rm S}$ with $\widehat a_{jk}^{(0)}=\cpca_{jk}$. By Proposition \ref{prop-ICO} and \eqref{thm:noiseless-2}, $\psi_{1,1}'\le \rho \psi_0$ and this would contribute the extra factor $\rho$ twice in the application of Proposition~\ref{prop-ICO} to $\psi_{1,2}'$, resulting in $\psi_{1,2}'\le \rho^3 \psi_0$, so on and so forth. In general, $\psi_{m,k}' \le \rho^{n_{(m-1)K+k}}\psi_0$ with $n_1 =1$, $n_2=3, \ldots, n_{K}=3^{K-1}$, and $n_{k+1} = 1+2\sum_{\ell=1}^{K-1}n_{k+1-\ell}$ for $k>K$. As $2(1-\gamma_K^{-K+1})=\gamma_K-1$, by induction for $k \ge K$, 
\bes
n_{k+1} \ge 2\big(\gamma_K^{k-1}+\cdots+\gamma_K^{k-K+1}\big) = \gamma_K^k \frac{2(1-\gamma_K^{-K+1})}{\gamma_K-1} = \gamma_K^k. 
\ees
The function $f(\gamma) = \gamma^K - 3\gamma^{K-1}+2$ is decreasing in $(1,3-3/K)$ and increasing $(3-3/K,\infty)$. Because $f(1)=0$ and $f(3) = 2>0$, we have $3-3/K < \gamma_K <3$. 
\end{proof}

\PropositionTransExt*
\begin{proof}[\bf Proof of Proposition \ref{lemma-transform-ext}] 
Let $A=\widetilde U_1 \widetilde D_1 \widetilde U_2^\top$ and $B=\widetilde V_1 \widetilde D_2 \widetilde V_2^\top$ be respectively the SVD of $A$ and $B$ with $\widetilde D_1 = \text{diag}(\widetilde\sigma_{11},...,\widetilde\sigma_{1r})$ and $\widetilde D_2 = \text{diag}(\widetilde\sigma_{21},...,\widetilde\sigma_{2r})$. 
Let $U = \widetilde U_1 \widetilde U_2^\top$ 
and $V = \widetilde V_1 \widetilde V_2^\top$.  
We have $\|\widetilde D_1^2 -I_r\|_{\rm S}=\|A^\top A -I_r\|_{\rm S}\le \delta$ 
and $\|\widetilde D_2^2 -I_r\|_{\rm S}=\|B^\top B -I_r\|_{\rm S}\le \delta$. 
Moreover, 
\bes
\|A Q B^\top - U Q V^\top\|_{\rm S}^2 
&=& \textmax_{\|u_1\|_2=\|u_2\|_2=1} 
\big|{u}_1^\top\big(\widetilde D_1 \widetilde U_2^\top Q \widetilde V_2 \widetilde D_2 
- \widetilde U_2^\top Q \widetilde V_2 \big) {u}_2 \big|^2
\cr &\le & 2\|Q\|_{\rm S}^2
\,\textmax_{\|u_1\|_2=\|u_2\|_2=1} 
\big\|\widetilde D_2 {u}_2 {u}_1^\top \widetilde D_1  - {u}_2 {u}_1^\top\big\|_{\rm F}^2
\cr &=& 2\|Q\|_{\rm S}^2
\max_{\|u_1\|_2=\|u_2\|_2=1}\sum_{i=1}^r\sum_{j=1}^r {u}_{1i}^2 {u}_{2j}^2\big(\widetilde\sigma_{1i}
\widetilde\sigma_{2j}-1\big)^2
\ees
with ${u}_{\ell}=({u}_{\ell 1},...,{u}_{\ell r})^\top$,  $\sqrt{(1-\delta)_+}\le \widetilde\sigma_{\ell j}
\le\sqrt{1+\delta}$, $\ell=1,2$. The maximum  
on the right-hand side above is attained at $\widetilde\sigma_{\ell j} =\sqrt{(1-\delta)_+}$ or $\sqrt{1+\delta}$ by convexity.  
As $(\sqrt{(1-\delta)_+}\sqrt{1+\delta}-1)^2
\le \delta^4\wedge 1$, we have $\|A Q B^\top - U Q V^\top\|_{\rm S}^2 
\le 2\|Q\|_{\rm S}^2\delta^2$. 
For nonnegative-definite $\Lambda$ and $B=A$, 
$\|A \Lambda A^\top - U \Lambda U^\top\|_{\rm S} 
= \|\widetilde D_1 \widetilde U_2^\top \Lambda \widetilde U_2 \widetilde D_1 
- \widetilde U_2^\top \Lambda \widetilde U_2\|_{\rm S}$ and 
\begin{align*}
\big|{u}^\top\big(\widetilde D_1 \widetilde U_2^\top \Lambda \widetilde U_2 \widetilde D_1 
- \widetilde U_2^\top \Lambda \widetilde U_2 \big) {u}\big|
= \bigg|\sum_{j=1}^2 \tau_j {v}_{j}^\top \widetilde U_2^\top \Lambda \widetilde U_2 {v}_{j} \bigg|
\le \begin{cases} \|\Lambda\|_{\rm S}(|\tau_1|\vee|\tau_2|), & \tau_1\tau_2<0, \cr 
\|\Lambda\|_{\rm S}(|\tau_1+\tau_2|), & \tau_1\tau_2\ge 0, \end{cases} 
\end{align*}
where $\sum_{j=1}^2 \tau_j {v}_{j} {v}_{j}^\top$ is the eigenvalue decomposition of 
$\widetilde D_1 {u} {u}^\top \widetilde D_1  - {u} {u}^\top$. 
Similar to the general case, 
$(|\tau_1|\vee|\tau_2|)^2 \le \tau_1^2+\tau_2^2 
=\big\|\widetilde D_1 {u} {u}^\top \widetilde D_1  - {u} {u}^\top\big\|_{\rm F}^2 
\le \delta^2$ and $|\tau_1+\tau_2| = |\text{tr}(\widetilde D_1 {u} {u}^\top \widetilde D_1  - {u} {u}^\top)|
\le \|\widetilde D_1 \widetilde D_1-I_r \|_{\rm S}\le\delta$. 
Hence, $\|A \Lambda A^\top - U \Lambda U^\top\|_{\rm S} \le \|\Lambda\|_{\rm S}\delta$. 
\end{proof}

\PropositionICOgen*

\begin{proof}[\bf Proof of Proposition \ref{prop-ICO-gen}] 
By the argument in the beginning of the proof of Proposition \ref{prop-ICO}, the conclusion of Proposition \ref{prop-ICO-gen} does not depend on the signs of $\widetilde a_{j\ell}$. Thus, we assume without loss of generality that $\widetilde a_{i\ell}^\top a_{i\ell}\ge 0$ for all $i$ and $\ell$.
Instead of $\max_{j\le r}\big\|\widetilde a_{j\ell} - a_{j\ell}\big\|_2^2 \le  \widetilde\psi_\ell^2/(1-1/(4r))$ in the proof of Proposition \ref{prop-ICO}, we have the simpler $\max_{j\le r}\big\|\widetilde a_{j\ell} - a_{j\ell}\big\|_2
\le  \widetilde\psi_\ell$ here. Modifying the proof there accordingly, we have
$\max_{i\le r}\big\|\widetilde a_{i\ell} - a_{i\ell}\big\|_2\|\widetilde b_{j\ell}\|_2 
\le \widetilde\psi_\ell
\big/\big(\sqrt{1-\delta_\ell}-r^{1/2}\widetilde \psi_\ell\big)_+ = \widetilde \phi_\ell$. Again let 
$w_{ij,\ell} = a_{i\ell}^\top \widetilde b_{j\ell}/a_{j\ell}^\top \widetilde b_{j\ell}$, 
$v_{i,jk} = (\lam_i/\lam_j)\prod_{\ell\in[N]\setminus \{k\}} w_{ij,\ell}$, 
$v_{jk}\in\R^r$ be the vector with elements $v_{i,jk}$, and 
$\widetilde \lam_j = \lam_j\prod_{\ell \in [N]\setminus \{k\}}(a_{j\ell}^\top\widetilde b_{j\ell})$. 
As $\widetilde a_{i\ell}^\top \widetilde b_{j\ell} = I_{\{i=j\}}$, 
for $i\neq j$, 
\bes
|w_{ij,\ell}| = \frac{|( a_{i\ell} - \widetilde a_{i\ell})^\top \widetilde b_{j\ell}| }
{|1 + (a_{j\ell} - \widetilde a_{j\ell})^\top \widetilde b_{j\ell}|}
\le \frac{\widetilde \phi_\ell}{1-\widetilde\phi_\ell},\quad  
\big|v_{i,jk}\big| \le (\lam_1/\lam_j)\prod_{\ell\in [N]\setminus \{k\}}\bigg(\frac{\widetilde \phi_\ell}{1-\widetilde\phi_\ell}\bigg). 
\ees
As $\widetilde  T_{jk}^*/\widetilde\lam_j = \sum_{i=1}^r a_{ik} v_{i,jk}$ and $v_{j,jk}=1$,  
\bes
\big\|\widetilde  T_{jk}^*/\widetilde\lam_j - a_{jk}\big\|_2^2  
&=& \textsum_{i_1\in [r]\setminus \{j\}}\textsum_{i_2\in [r]\setminus \{j\}}\sigma_{i_1i_2,k}v_{i_1,jk}v_{i_2,jk} 
\cr &\le & (r-1) \big(1 + \delta_k\big)(\lam_1/\lam_j)^2 \prod_{\ell\in [N]\setminus \{k\}}(\widetilde\phi_\ell/(1-\widetilde\phi_\ell))^2. 
\ees
Let $2\theta$ be the angle between $a_{jk}$ and $\widetilde a^*_{jk} =\widetilde  T_{jk}^*/\|\widetilde  T_{jk}^*\|_2$. We have 
$2\big(1-a_{jk}^\top \widetilde a^*_{jk}\big) = \|a_{jk} - \widetilde a^*_{jk}\|_2^2 
= (2\sin \theta)^2 = 2(1-\cos(2\theta))
\le 2(1-\cos^2(2\theta)) = 2\sin^2(2\theta) 
\le 2 \big\|\widetilde  T_{jk}^*/\widetilde\lam_j - a_{jk}\big\|_2^2$. 

Similarly, as $\widetilde\lambda_j^*-\lambda_j = \lambda_j (\textprod_{\ell\in [N]} a_{j\ell}^\top \widetilde b_{j\ell} -1) + \textsum_{i\in [r]\setminus \{j\}}\lambda_i\prod_{\ell\in [N]} a_{i\ell}^\top \widetilde b_{j\ell} $, we have
\begin{align*}
\big|\widetilde\lambda_j^*/\lambda_j -1| \le \sum_{\ell \in [N]} \widetilde\phi_{\ell} +(r-1)(\lambda_1/\lambda_j) \prod_{\ell\in [N]} \widetilde \phi_{\ell} .    
\end{align*}
\end{proof}

\TheoremNoiselessGen*

\begin{proof}[\bf Proof of Theorem \ref{thm:noiseless-gen}] 
By definition $\mat_S(T^*)=A_S\Lambda A_{S^c}^\top= \textsum_{j=1}^r \cpcalam_j \widehat u_j \widehat v_j^\top$, so that for the $U=(u_1,\ldots,u_r)$ and $V=(v_1,\ldots,v_r)$ in  Proposition~\ref{lemma-transform-ext} we have 
$\|a_{jS}a_{jS}^\top - u_ju_j^\top\|_{\rm S}\le\delta_S$,  $\|a_{jS^c}a_{jS^c}^\top - v_jv_j^\top\|_{\rm S}\le\delta_{S^c}$, $1\le j\le r$, and $\|\mat_S(T^*)-U\Lambda V^\top\|_{\rm S}\le \sqrt{2}\lam_1\delta$. 
These and Proposition \ref{prop-rank-1-approx} yield \eqref{thm:noiseless-gen-0} as in the proof of Theorem \ref{thm:noiseless}.  
Moreover, under  \eqref{thm:noiseless-gen-2} we have
$\psi_0^2<1$, so that
$2(1- |a_{j\ell}^\top \cpca_{j\ell}|) \le 2\big\|\cpca_{j\ell}\cpcatop_{j\ell} - a_{j\ell}a_{j\ell}^\top \big\|_{\rm S}^2
\le \psi_0^2$. 
Define $\psi_{m,k}'=\max_{j\le r} \big(2- 2\big|a_{jk}^\top\widehat a_{jk}^{(m)}\big|\big)^{1/2}$ 
with $\widehat a_{jk}^{(0)}=\cpca_{jk}$. 
By Proposition \ref{prop-ICO-gen} and \eqref{thm:noiseless-gen-2}, $\psi_{1,1}'\le \rho \psi_0$ and 
similar to the proof of Theorem \ref{thm:noiseless}, we have $\psi_{m,k}' \le \rho^{n_{(m-1)N+k}}\psi_0$ with 
$n_1 =1$, $n_2=2, \ldots, n_{N}=2^{N-1}$, and $n_{k+1} = 1+\sum_{\ell=1}^{N-1}n_{k+1-\ell}$ for $k>N$. 
By induction, for $k=N, N+1,\ldots$. 
\bes
n_{k+1} \ge \gamma_N^{k-1}+\cdots+\gamma_N^{k-N+1} = \gamma_N^k \frac{1-\gamma_N^{-N+1} }{\gamma_N-1} 
= \gamma_N^k. 
\ees
The function $f(\gamma) = \gamma^N - 2\gamma^{N-1}+1$ is decreasing in $(1,2-2/N)$ and increasing $(2-2/N,\infty)$. 
Because $f(1)=0$ and $f(2)=1>0$, we have $2-2/N < \gamma_N <2$. 
By Proposition \ref{prop-ICO-gen}, \eqref{thm:noiseless-gen-2} and the upper bound for $\psi_{m,k}'$, we have the desired upper bound for $\max_{j\le r}\big| \widehat\lambda_j^{(m)}/\lambda_j -1 \big|$. 
\end{proof}

\PropositionRotate*

\begin{proof}[\bf Proof of Proposition \ref{prop-rotation}] 
Suppose $M$ is a rank-1 tensor in SP$=\text{span}\{a_1,...,a_r\}$, where $a_j=\vec(\otimes_{k=1}^K a_{jk})$. Thus, there exist coefficients $\beta_j,j\le r$, such that
\begin{align*}
M=\beta_1 \vec(\otimes_{k=1}^K a_{1k}) + \cdots +\beta_r \vec(\otimes_{k=1}^K a_{rk}).    
\end{align*}
In matrix form, it follows that
\begin{align*}
\mat_1(M)=\beta_1 a_{11}\vec(\otimes_{k=2}^K a_{1k})^\top + \cdots +\beta_r a_{r1} \vec(\otimes_{k=2}^K a_{rk})^\top,   
\end{align*}
where $\{a_{j1},j\le r\}$ is a set of linearly independent vectors, and $\{\vec(\otimes_{k=2}^K a_{jk}),j\le r\}$ is also a set of linearly independent vectors. Note that the matrix on the left hand side has rank 1 while the matrix on the right hand side has rank $|j\in[r]:\beta_j\neq 0|$. Since the rank of a matrix is unambiguously determined, we must have $|j\in[r]:\beta_j\neq 0|=1$. Hence, $M=\beta_{j_*} a_{j_*}$ holds for some $j_*\in[r]$.
\end{proof}

\section{Analysis of CPCA and ICO for noisy tensors} 

This section provides the analysis of the CPCA and ICO algorithms in the noisy case of models \eqref{model} and \eqref{model2}. In addition to the propositions provided before, we use concentration inequalities to derive the statistical error bounds. 

\subsection{Proofs of main theorems}

\theoremfactorcpca*

\begin{proof}[\bf Proof of Theorem \ref{thm:initial2}]
Recall that $\lambda_j=w_j^2 $ with $\lam_1\ge\cdots\ge\lam_r>0$, $A=(A_1,\ldots,A_r)$ with $a_j=\vec( a_{j1}\otimes a_{j2}\otimes\cdots\otimes a_{jK})$, $T=n^{-1}\sum_{i=1}^n \cX_i\otimes \cX_i$ 
and $d=d_1d_2...d_K$. 
Write
\begin{align}\label{eq:initial2:decomp}
\mat_{[K]}(T) = \sum_{j=1}^r \lam_j \big(\text{vec}(\otimes_{k=1}^K a_{jk})\big)^{\otimes 2} + \sigma^2 I_{d} + \Psi^* 
= A \Lambda A^{\top} + \sigma^2 I_{d} + \Psi^*,
\end{align}
where $\Lambda=\text{diag}(\lambda_1,...,\lambda_r)$ and $\Psi^*=\mat_{[K]}(T-\E[T]) = \mat_{[K]}(\Psi)-\sigma^2 I_d$. 
Let $U=(u_1,\ldots,u_r)$ be the orthonormal matrix corresponding to $A$ as in Proposition~\ref{lemma-transform}. We have $\|AA^\top - UU^{\top}\|_{\rm S}\le \delta$ and $\|A \Lambda A^\top - U \Lambda U^{\top}\|_{\rm S}\le \lam_1\delta$ by two applications of the error bound in Proposition~\ref{lemma-transform} with $\Lambda=I_r$ the first time.
Let the top $r$ eigenvectors of $\mat_{[K]}(T)$ be $ \widehat U=(\widehat u_1, ..., \widehat u_r)\in\R^{d\times r}$. 
By Wedin's perturbation theorem \citep{wedin1972} 
for any $1\le j\le r$,
\begin{align} 
\|\widehat  u_j \widehat  u_j^\top - u_j  u_j^{\top} \|_{\rm S} \le  
2 \|
A \Lambda A^{\top} - U \Lambda U^{\top}
+ \Psi^* \|_{\rm S}/\lam_{j,\pm} 
\le 
\big(2\lambda_1 \delta + 2 \|\Psi^*\|_{\rm S}\big)\big/\lam_{j,\pm}.
\label{eq2:thm:initial2}
\end{align}
Combining \eqref{eq2:thm:initial2} and the inequality 
$\|AA^\top - UU^{\top}\|_{\rm S}\le \delta$,  
we have
\begin{align}\label{eq3:thm:initial2}
\|\widehat u_j \widehat u_j^\top - a_j a_j^\top \|_{\rm S}&\le \delta+
\big(2\lambda_1 \delta + 2 \|\Psi^*\|_{\rm S}\big)/\lam_{j,\pm}
\end{align}

We formulate each $\widehat u_j\in\R^d$ to be a $K$-way tensor $\widehat U_j\in \R^{d_1\times\cdots\times d_K}$. Let $\widehat U_{jk}=\mat_k(\widehat U_j)$, which is viewed as an estimate of  $a_{jk}\vec(\otimes_{l\neq k}^K a_{jl})^\top\in\R^{d_k\times (d/d_k)}$. Then $\cpca_{jk}$ is the top left singular vector of $\widehat U_{jk}$. 
By Proposition \ref{prop-rank-1-approx}, 
\begin{align}
\|\cpca_{jk} \cpcatop_{jk} - a_{jk} a_{jk}^\top \|_{\rm S}^2 \wedge (1/2) \le 
\|\widehat u_j \widehat u_j^\top - a_j a_j^{\top} \|_{\rm S}^2 .
\end{align}
Substituting \eqref{eq3:thm:initial2} and Lemma \ref{lem:psi} into the above equation, 
based on the definition of SNR and $R^{(0)}$, we have the desired results. 
We note that \eqref{thm:initial2:eq1} holds automatically when the right-hand side is greater than 1, e.g. $\delta\ge 1$.  
\end{proof}

\begin{lemma}\label{lem:psi}
Suppose the assumptions in Theorem \ref{thm:initial2} hold 
and $\delta<1$. Let $\Psi^*=\mat_{[K]}(T-\E[T])$ and $\lambda_j=w_j^2$ in \eqref{model}.  
In an event with probability at least $1-e^{-t}$, 
we have
\bes  
1\wedge \big(\|\Psi^*\|_{\rm S}/\lam_1\big) 
\le C\max\Big(\sqrt{(r_{\rm \tiny eff}/n)(1+1/\hbox{\rm SNR})(1+(r_{\rm \tiny eff}/d)/\hbox{\rm SNR})},\sqrt{t/n}\Big)
\ees
for all $0\le t\le d$, 
where $C$ is a numerical constant. 
\end{lemma}

\begin{proof}
Let $T^* = \E[T]$. 
As $\Psi^*=\mat_{[K]}(T-T^*)$, it follows from Theorem 2 of \cite{koltchinskii2016asymptotics} that 
\bes 
\P\Big\{\|\Psi^*\|_{\rm S}
\ge C\|\mat_{[K]}(T^*)\|_{\rm S}\Big(\sqrt{r^*/n}\vee(r^*/n)\vee\sqrt{t/n}\vee(t/n)\Big)\Big\}\le e^{-t}
\ees 
where $r^*=\trace(\mat_{[K]}(T^*))/\|\mat_{[K]}(T^*)\|_{\rm S}$ is the effective rank of $\mat_{[K]}(T^*)$ and $C$ is a numeric constant. 
Because 
$\mat_{[K]}(T^*) = \E[\mat_{[K]}(T)]
= \sum_{j=1}^r \lam_ja_ja_j^\top + \sigma^2 I_d$, 
$\|\mat_{[K]}(T^*)\|_{\rm S} \le 2\lam_1+\sigma^2$ and 
$\trace(\mat_{[K]}(T^*)) = \lam_1r_{\rm \tiny eff}+\sigma^2d =\lam_1r_{\rm \tiny eff}(1+1/\hbox{\rm SNR})$ 
with $r_{\rm \tiny eff}\le r\le d$, so that  
\bes 
&& \min\Big\{1,(3\lam_1)^{-1}\|\mat_{[K]}(T^*)\|_{\rm S}\Big(\sqrt{r^*/n}\vee(r^*/n)\vee\sqrt{t/n}\vee(t/n)\Big)\Big\}
\cr &\le& \max\Big(
\sqrt{(r_{\rm \tiny eff}+\sigma^2d/\lam_1)(2/3+\sigma^2/(3\lam_1))/n},(2/3+\sigma^2/(3\lam_1))\sqrt{t/n}\Big)
\cr &\le& \max\Big(
\sqrt{(r_{\rm \tiny eff}/n)(1+1/\hbox{\rm SNR})(1+(r_{\rm \tiny eff}/d)/\hbox{\rm SNR})},\sqrt{t/n}\Big). 
\ees 
Note the component of the maximum with $\sqrt{t/n}$ is smaller when $\lam_1\le\sigma^2$ and $0\le t\le d$. 
\end{proof}

\theoremfactorico*

\begin{proof}[\bf Proof of Theorem \ref{thm:projection2}]
We divide the proof into three steps. 

\noindent\underline{\bf Step 1 (Error bound for a single update).} 
Consider given $(j,k)$ in this step. 
Recall that $(b_{1k},\ldots,b_{rk})=A_k(A_k^\top A_k)^{-1}$ with $A_k=(a_{1k},\ldots,a_{rk})$. Let 
$z_n\sim N(0,I_n)$. 
For $g = \{g_1,\ldots,g_{2K}\}$ with $g_k = g_{k+K} \in \R^{d_k}$, define $T_k(g)$ as 
\bes
T\times_{\ell\in [2K]\setminus\{k,k+K\}}g_\ell^\top 
= X_{k}(g)^\top X_{k}(g)/n\in\R^{d_k\times d_k}
\ees
with $X_k(g)=(\cX_i\times_{\ell\in [K]\setminus\{k\}} g_\ell, i\in [n])^\top\in\R^{n\times d_k}$. 
Write 
\bes
X_{k}(g) = M_{jk}(g) +M_{jk}^c(g) + E_{k}(g)
\ees
where $E_{k}(g) = \big(\cE_i\times_{\ell\in [K]\setminus\{k\}} g_\ell^\top, i\in [n]\big)^\top\in\R^{n\times d_k}$, 
\bes
M_{jk}(g) = \big(f_{ij}, i\in [n]\big)^\top \big(w_j\textprod_{\ell\in [K]\setminus \{k\}} a_{j\ell}^\top g_\ell\big)a_{jk}^\top
\ees
as a rank-one $n\times d_k$ random matrix with signal, and 
\bes
M_{jk}^c(g) = \textsum_{h\in [r]\setminus\{j\}} M_{hk}(g)\in\R^{n\times d_k}. 
\ees
As $T_k(g) = T\times_{\ell\in [2K]\setminus\{k,k+K\}}g_\ell^\top$, it follows that 
\bel{T-decomp}
T_k(g) = \widebar\lambda_j(g)a_{jk}a_{jk}^\top + \sigma^2 I_{d_k}+\Delta_{jk}(g), 
\eel
where $\Delta_{jk}(g) = \sum_{i=1}^5\Delta_{jk}^{(i)}(g)$,  
\bes
 \widebar\lambda_j(g) &=& \lam_j \big\{\textprod_{\ell\in [K]\setminus \{k\}}(a_{j\ell}^\top g_\ell)^2\big\}
 \textsum_{i=1}^n f_{ij}^2/n, 
\cr \Delta_{jk}^{(1)}(g) &=& M_{jk}^{c\top}(g)M_{jk}^{c}(g)/n, 
\cr \Delta_{jk}^{(2)}(g) &=& E_{jk}^{\top}(g)E_{jk}(g)/n  - \sigma^2 I_{d_k}, 
\cr \Delta_{jk}^{(3)}(g) &=& E_{jk}^{\top}(g)M_{jk}^{c}(g)/n + M_{jk}^{c\top}(g)E_{jk}(g)/n, 
\cr \Delta_{jk}^{(4)}(g) &=& E_{jk}^{\top}(g)M_{jk}(g)/n + M_{jk}^{\top}(g)E_{jk}(g)/n, 
\cr \Delta_{jk}^{(5)}(g) &=& M_{jk}^{\top}(g)M_{jk}^{c}(g)/n + M_{jk}^{c\top}(g)M_{jk}(g)/n. 
\ees
We bound $\widebar\lambda_j(g)$ and $\|\Delta_{jk}(g)\|_{\rm S}$ over $g_\ell\in G_{j\ell}$ with 
\begin{align}\label{G_jl}
G_{j\ell}=\{g_\ell\in \mathbb{S}^{d_\ell-1}: \|g_\ell - b_{j\ell}/\|b_{j\ell}\|_2\|_2\le \phi, 
|a_{j\ell}^\top g_\ell|\ge \alpha, 
\textmax_{h\neq j} |a_{h\ell}^\top g_\ell|\le 
\psi'_\ell=\psi_\ell/\sqrt{1-1/(4r)}\}
\end{align}
for $\ell\neq k$. In addition, we set $G_\ell=\mathbb{S}^{d_\ell-1}$ and $G_{jk}=G_k$. 

By the Gaussian concentration of $(\sum_{i=1}^n f^2_{ij})^{1/2}$, 
\bes
&& \inf_{g_\ell\in G_{j\ell}, \ell\in [K]\setminus \{k\}}\widebar\lam_j(g)
\ge \frac{\lam_j\alpha^{2K-2}}{(1-1/\sqrt{n}-\sqrt{2t/n})^{-2}}
\qquad
\ees
with at least probability $1-e^{-t}$. 

Similarly, in an event 
with at least probability $1-e^{-t}$, 
\bes
\big\|\textsum_{i=1}^n F_iF_i^\top/n\big\|_{\rm S}\le (1+\sqrt{r/n}+\sqrt{2t/n})^2
\ees
with $F_i=(f_{i1},\ldots,f_{ir})^\top$, and in the same event
\bes
 \sup_{g_\ell\in G_{j\ell}, \ell\in [K]\setminus \{k\}}\big\|\Delta_{jk}^{(1)}(g)\big\|_{\rm S}
 &\le& \frac{\sup_{g_\ell\in G_{j\ell}, \ell\in [K]}\sum_{h\neq j}\lam_h\prod_{\ell\in [K]}(a_{h\ell}^\top g_\ell)^2}
{(1+\sqrt{r/n}+\sqrt{2t/n})^{-2}}
\cr &=& \frac{\big\|\sum_{h\in[r]\setminus\{j\}}\lam_h a_{hk}a_{hk}^\top\big\|_{\rm S}
\big(\prod_{\ell\in [K]\setminus \{k\}}\psi'_\ell\big)^2}{(1+\sqrt{r/n}+\sqrt{2t/n})^{-2}}
\cr &\le & \frac{\lam_1(1+\delta_k)\big(\prod_{\ell\in [K]\setminus \{k\}}\psi'_\ell\big)^2}
{(1+\sqrt{r/n}+\sqrt{2t/n})^{-2}}.
\ees

Let $\phi'=\phi\wedge 1$. 
For the noise component, the Sudakov-Fernique and Gaussian concentration inequalities provide  
\bes
\sup_{g_\ell\in G_{j\ell}, \ell\in [K]\setminus \{k\}} \|E_{k}(g)\|_{\rm S}
&\le& \sigma \E[\|z_n\|_2] +\sigma \sqrt{2t} 
+ \E\bigg[\sup_{g_\ell\in G_{j\ell}\forall \ell\in [K]}\cE_1\times_{\ell=1}^Kg_\ell\bigg] 
\cr&=& \sigma\bigg(\E[\|z_n\|_2] +\sqrt{2t} 
+ \E[\|z_{d_k}\|_2] +\phi'\sum_{\ell\neq k}  \E[\|z_{d_\ell}\|_2]\bigg)
\ees
with at least probability $1-e^{-t}$. 
Similarly, the smallest singular value $\sigma_1(E_k(g))$ is bounded from below by 
\bes
\inf_{g_\ell\in G_{j\ell}, \ell\in [K]\setminus \{k\}} \sigma_1\big(E_{k}(g)\big)
&\ge& \sigma\bigg(\E[\|z_n\|_2] - \sqrt{2t} 
- \E[\|z_{d_k}\|_2] - \phi'\sum_{\ell\neq k}  \E[\|z_{d_\ell}\|_2]\bigg)
\ees
with at least probability $1-e^{-t}$. Thus, 
\bes
\sup_{g_\ell\in G_{j\ell}, \ell\in [K]\setminus \{k\}}\big\|\Delta_{jk}^{(2)}(g)\big\|_{\rm S}
&\le& \sigma^2\bigg\{\bigg(1+\frac{\sqrt{2t}+ \sqrt{d_k}}{\sqrt{n}}
+\phi'\sum_{\ell\neq k}\frac{\sqrt{d_\ell}}{\sqrt{n}}\bigg)^2-1\bigg\}
\ees
with at least probability $1-2e^{-t}$. 

For each of the three cross-product terms, the two matrix factors are independent. 
Thus, an application of the above calculation in the proof of Lemma G.2 of \cite{han2020iterative} yields
\begin{align*}
&\sup_{g_\ell\in G_{j\ell}, \ell\in [K]\setminus \{k\}}\big\|\Delta_{jk}^{(3)}(g)\big\|_{\rm S} \\
\le& 2\sup_{g_\ell\in G_{j\ell}, \ell\in [K]\setminus \{k\}} \|E_{k}^\top(g)M_{jk}^c(g)\|_{\rm S}/n \\
\le& 2\sigma\sqrt{\lam_1(1+\delta_k)}\bigg(\prod_{\ell\in [K]\setminus \{k\}}\psi'_\ell\bigg)
\bigg\{\bigg(1+\frac{\sqrt{r}+\sqrt{2t}}{\sqrt{n}}\bigg)\bigg(\frac{\sqrt{d_k}}{\sqrt{n}}
+\phi'\sum_{\ell\neq k}\frac{\sqrt{d_\ell}}{\sqrt{n}}\bigg)
 + \frac{\sqrt{r}}{\sqrt{n}}\bigg(1+\frac{\sqrt{2t}}{\sqrt{n}}\bigg)
+\frac{2t+2\sqrt{2t}}{n}\bigg\} 
\end{align*}
with at least probability $1-2e^{-t}$, 
\bes
\sup_{g_\ell\in G_{j\ell}, \ell\in [K]\setminus \{k\}}\big\|\Delta_{jk}^{(4)}(g)\big\|_{\rm S}
&\le& 2\sup_{g_\ell\in G_{j\ell}, \ell\in [K]\setminus \{k\}} \|E_{k}^\top(g)M_{jk}(g)\|_{\rm S}/n
\cr&\le& 2 \sigma \lam_j^{1/2} 
\bigg\{\bigg(1+\frac{1+\sqrt{2t}}{\sqrt{n}}\bigg)\bigg(\frac{\sqrt{d_k}}{\sqrt{n}}
+ \phi'\sum_{\ell\neq k}\frac{\sqrt{d_\ell}}{\sqrt{n}}\bigg)  + \frac{1}{\sqrt{n}}
+\frac{2t+3\sqrt{2t}}{n}\bigg\} 
\ees
with at least probability $1-2e^{-t}$, and 
\bes
\sup_{g_\ell\in G_{j\ell}, \ell\in [K]\setminus \{k\}}\big\|\Delta_{jk}^{(5)}(g)\big\|_{\rm S}
 &\le & 2\sup_{g_\ell\in G_{j\ell}, \ell\in [K]\setminus \{k\}} \|M_{jk}^\top(g)M_{jk}^c(g)\|_{\rm S}/n
\cr&\le&2 \lam_j^{1/2}\sqrt{\lam_1(1+\delta_k)}\bigg(\prod_{\ell\in [K]\setminus \{k\}}\psi'_\ell\bigg)
 \bigg\{\bigg(1+\frac{1+\sqrt{2t}}{\sqrt{n}}\bigg)\frac{\sqrt{r}}{\sqrt{n}} + \frac{1}{\sqrt{n}}
+\frac{2t+3\sqrt{2t}}{n}\bigg\} 
\ees
with at least probability $1-2e^{-t}$. 

Putting the above inequalities together, we find that 
for $r\le n$ and with at least probability $1-e^{-2(d_1\wedge \sqrt{n})}$, 
\bel{lam-bd}
&& \inf_{g_\ell\in G_{j\ell}, \ell\in [K]\setminus \{k\}}\widebar\lam_j(g)
\ge \lam_j\alpha^{2K-2}/C_0'
\eel
and with $\psi_{-k}=\prod_{\ell\in [K]\setminus\{k\}}\psi_\ell$ 
\begin{align}\label{Delta-bd}
 & \sup_{g_\ell\in G_{j\ell}, \ell\in [K]\setminus \{k\}}\big\|\Delta_{jk}(g)\big\|_{\rm S} \notag\\
 \le & C_0'\lam_1\psi_{-k}^2 + C_0'\sigma^2 \big(d_{k,\phi}^{1/2}/n^{1/2}+ d_{k,\phi}/n\big)
 + C_0'\lam_j^{1/2}\sigma d_{k,\phi}^{1/2}/n^{1/2}+ C_0'(\lam_1\lam_j r/n)^{1/2}\psi_{-k}, 
\end{align}
where $d_{k,\phi}=\big(d_k^{1/2}+(\phi \wedge 1)\sum_{\ell\in [K]\setminus \{k\}} d_\ell^{1/2}\big)^2$ 
and $C_0'$ is a numeric constant.  
Here the upper bound for $\Delta_{jk}^{(3)}$ is absorbed into those for $\Delta_{jk}^{(1)}$ and $\Delta_{jk}^{(2)}$ 
by Cauchy-Schwarz.  

Let $\widehat a_{jk}(g)$ be the top eigenvector of 
$T_k(g)$ in \eqref{T-decomp}. 
As $\|a_{jk}\|_2=\|\widehat a_{jk}(g)\|_2=1$, 
\eqref{T-decomp}, \eqref{lam-bd} and \eqref{Delta-bd} imply 
\begin{align}\label{a(g)-bd}
\sup_{g_\ell\in G_{j\ell}, \ell\in [K]\setminus \{k\}}
\|\widehat a_{jk}(g)\widehat a_{jk}^\top (g) - a_{jk}a_{jk}^\top\|_{\rm S} 
\le& 
C_{0,\alpha}\max\Big\{(\lam_1/\lam_j)\psi_{-k}^2, R^{\ideal}_{jk,\phi}, 
\sqrt{(\lam_1/\lam_j)(r/n)}\psi_{-k}\Big\} 
\end{align}
with at least probability $1-e^{-2(d_1\wedge \sqrt{n})}$, 
where $G_{j\ell}$ are as in \eqref{G_jl}, 
$R^{\ideal}_{jk,\phi}$ as in \eqref{R_jk-ideal} 
and $C_{0,\alpha}=C_0\alpha^{2-2K}$ with a numeric constant $C_0$. 
Here we assume $C_0$ can be taken as the constant in \eqref{alpha} and \eqref{C_0}.

\noindent\underline{\bf Step 2 (Error bound sequences).} 
Recall that $\widehat A_\ell^{(m)}
=(\widehat a_{1\ell}^{(m)}, \ldots, \widehat a_{r\ell}^{(m)})\in \R^{d_\ell\times r}$, 
$\widehat\Sigma_\ell^{(m)}=\widehat A_\ell^{(m)\top}\widehat A_\ell^{(m)}$, 
and $\widehat B_\ell^{(m)} = \widehat A_\ell^{(m)} (\widehat\Sigma_\ell^{(m)})^{-1} 
= (\widehat b_{1\ell}^{(m)},...,\widehat b_{r\ell}^{(m)}) \in\R^{d_\ell\times r}$. 
Let 
\bel{Omega_mell}
\Omega_{m,\ell}=\Big\{\max_{h\le r}\|\ahatm_{h\ell}\ahatmtop_{h\ell} - a_{h\ell}a_{h\ell}^\top\|_{\rm S}
\le \psi_{m,\ell}\Big\} 
\eel
with constants $\psi_{m,\ell}\le\psi_0$ to be specified later sequentially. 
As the PCA of $T(g)$ in \eqref{T-decomp} does not depend on the signs of $g_{h\ell}$, 
we may assume without loss of generality $a_{h\ell}^\top\ahatm_{h\ell}\ge 0$ 
for all $(h,\ell)$. Thus, in $\Omega_{m,\ell}$ the proof of Proposition \ref{prop-ICO} provides  
\begin{align}\label{a-bd}
\max_{h\le r}\|\ahatm_{h\ell}-a_{h\ell}\|_2 \le \psi_{m,\ell}/\sqrt{1-1/(4r)}, \ \ 
\displaystyle \big\|\bhatm_{h\ell}\big\|_2 \le \|\Bhatm_\ell\|_{\rm S}^{1/2}
\le \bigg(\sqrt{1-\delta_\ell}-\frac{r^{1/2}\psi_0}{\sqrt{1-1/(4r)}}\bigg)^{-1}. 
\end{align}
Let $P_\ell =A_\ell(A_\ell^\top A_\ell)^{-1}A_\ell^\top$ and 
$P_\ell^\perp =I_{d_\ell} - P_\ell^\top$. 
As $\Bhatm_{\ell} - B_{\ell} = P_\ell^\perp \big(\Ahatm_\ell-A_\ell\big)(\widehat\Sigma_\ell^{(m)})^{-1} 
-  B_\ell\big(\Ahatm_\ell - A_\ell\big)^\top \Bhatm_\ell$,  
\bes
\big\|\bhatm_{h\ell} - b_{h\ell}\big\|_2^2 
 &\le& \big\|\Ahatm_\ell-A_\ell\big\|_{\rm S}^2
\big(\|\Bhatm_\ell\|_{\rm S}^2+\|B_\ell\|_{\rm S}^2\big)\big(\|\bhatm_{h\ell}\|_2^2\wedge\|b_{h\ell}\|_2^2\big)
\cr &\le& \{r\psi_{m,\ell}^2/(1-1/(4r))\} 
(2/\alpha^2)\big(\|\bhatm_{h\ell}\|_2^2\wedge\|b_{h\ell}\|_2^2\big)
\ees 
by the algebraic symmetry between the estimator and estimand, 
where $\alpha$ is as in \eqref{alpha}. 
Let $\ghatm_{h\ell} = \bhatm_{h\ell}/\|\bhatm_{h\ell}\|_2$. 
As $\big\|\ghatm_{h\ell} - b_{h\ell}\big\|_2\le \big\|\bhatm_{h\ell} - b_{h\ell}\big\|_2$ 
for $\|\bhatm_{h\ell}\|_2\ge \|b_{h\ell}\|_2=1$, 
\bel{b-bd}
\big\|\ghatm_{h\ell} - b_{h\ell}/\|b_{h\ell}\|_2\big\|_2
\le (\psi_{m,\ell}/\alpha)\sqrt{2r/(1-1/(4r))} 
\eel
by scale invariance. 
Moreover, 
\eqref{a-bd} provides 
\bel{g-bd}
\max_{h\neq j}\big|a_{h\ell}^\top\ghatm_{j\ell}\big| \le \psi_{m,\ell}/\sqrt{1-1/(4r)},\ 
\big|a_{j\ell}^\top\ghatm_{j\ell}\big| \ge \alpha, 
\eel
as $\ahatmtop_{h\ell}\ghatm_{j\ell}=I\{h=j\}/\|\bhatm_{j\ell}\|_2$. 
Thus, in the event $\Omega_{m,\ell}$, $\ghatm_{h\ell}=\bhatm_{h\ell}/\|\bhatm_{h\ell}\|_2\in G_{j,\ell}$ 
for $\ell\neq k$ in \eqref{G_jl} with $\psi_\ell=\psi_{m,\ell}$, the $\alpha$ in \eqref{alpha} and 
any upper bound $\phi$ for \eqref{b-bd}.  

Let 
\bes
\Omega_{m,j,k}=\big\{\|\ahatm_{jk}\ahatmtop_{jk} - a_{jk}a_{jk}^\top\|_{\rm S}
\le \psi_{m,j,k}\big\}. 
\ees
Let $\psi_{0,j,k} = \psi_{0,k}=\psi_0$ and sequentially update them by 
\bel{psi_mk}
\psi_{m,j,k} &=& C_{0,\alpha}\Big\{\Big((\lam_1/\lam_j)\textprod_{\ell=1}^{K-1}\psi_{m,k-\ell}^2\Big)
\vee R^{\ideal}_{jk,\phi_{m,k}} \vee \Big(\sqrt{r/n}\sqrt{\lam_1/\lam_j}
\textprod_{\ell=1}^{K-1}\psi_{m,k-\ell}\Big)\Big\},  
\cr \psi_{m,k} &=& \psi_{m,r,k}, 
\eel
$k=1,\ldots,K$, $m=1,2,\ldots,$  
with the $C_{0,\alpha}$ in \eqref{a(g)-bd} and 
\bes
\phi_{m,k}=1\wedge\big(\textmax_{1\le \ell<K}(\psi_{m,k-\ell}/\alpha)\sqrt{2r/(1-1/(4r))}\big). 
\ees
Here and in the sequel, we take the convention $(m,\ell)=(m-1,K+\ell)$ with the subscript $(m,\ell)$. 
We note that $\psi_{m,k}$ depends on $\psi_{m,k-1},\ldots,\psi_{m,k-K+1}$ only as 
an increasing function of their product and maximum.  
Thus, as $\psi_{1,k}\le \psi_{0,k}=\psi_0$ by \eqref{C_0}, 
$\psi_{m,k} \le \psi_{m-1,k}$ for all $k\in [K]$ and $m\ge 1$ by induction. 

By \eqref{a(g)-bd}, \eqref{G_jl}, \eqref{b-bd} and \eqref{g-bd}, 
the events $\Omega_{m,\ell}$ in \eqref{Omega_mell} satisfy
\bel{prob-bd}
\P\big\{\big(\cap_{\ell=1}^{K-1}\Omega_{m,k-\ell}\big) \cap \Omega_{m,j,k}^c\big\} 
\le e^{-2(d_1\wedge \sqrt{n})} 
\eel
with $\cap_{j=1}^r \Omega_{m,j,k}\subseteq \Omega_{m,k}$.

Let $\phi_0= (\psi^*/\alpha)\sqrt{2r/(1-1/(4r))}$ be as in \eqref{alpha} 
with $\psi^*=C_{0,\alpha}R^{\ideal}_{rK,1}$. 
A simple way of dealing with the dynamics of \eqref{psi_mk} 
is to compare $\psi_{m,j,k}$ with 
\bel{psi*}
\psi^*_{m,j,k} &=& C_{0,\alpha}\Big\{
\Big((\lam_1/\lam_j)\textprod_{\ell\neq k}\psi^{*2}_{m-1,\ell}\Big)
\vee R^{\ideal}_{jk,\phi_0} \vee \Big(\sqrt{r/n}\sqrt{\lam_1/\lam_j}
\textprod_{\ell\neq k}\psi^*_{m-1,\ell}\Big)\Big\}, 
\cr \psi^*_{m,k} &=& \psi^*_{m,r,k}, 
\eel
with initialization $\psi^*_{0,j,k}=\psi_0$. 
Compared with \eqref{psi_mk}, \eqref{psi*} is easier to analyze due to 
the use of static $\phi_0$ and the monotonicity of $\psi^*_{m,k}$ in $k$. 
While \eqref{psi_mk} uses inputs with indices $(m,k-[K-1])$, 
\eqref{psi*} uses inputs with indices $(m-1,[K]\setminus\{k\})$. 
Thus, as $\max_{j,k,\phi}C_{0,\alpha}R^{\ideal}_{jk,\phi}\le \psi^*$, 
$\psi_{m,k}\le \psi^*_{m,k}$ before $\psi^*_{m,k}$ first hits $(0,\psi^*]$ 
at a certain $(m^*,k^*)$. 
As $\psi^*_{m^*,k}\le\psi^*$ for $k\in [K]$, 
$\psi_{m^*,k}\le \psi^*$ for $k\in [K]$, 
so that $\phi_{m,k}\le \phi_0$ for $m>m^*$. 
It follows that 
\bel{psi-compare}
& \psi_{m+1,j,k}\le\psi^*_{m,j,k}\le \psi^*_{m,k},\quad\forall (m,j,k). 
\eel

\noindent\underline{\bf Step 3 (Contraction of error bounds).} 
Recall that $\eps_{jk}=C_{0,\alpha}R^{\ideal}_{jk,\phi_0}$. 
By \eqref{prob-bd} and \eqref{psi-compare}, 
\eqref{th-3-bd} follows from 
\bes
\psi^*_{m,j,k} \le \eps \vee \eps_{jk} \quad \forall j,k,
\ees
for $\eps_{r2} \le \eps\le\psi_0$ and $m\ge m_\eps+2$.  
Let $\eps_{r,K+1}=\psi_0$. 
By induction, it suffices to prove 
that for $\eps_{rk_0}\le \eps < \eps_{r,k_0+1}$ 
\bel{psi-bd}
\psi^*_{m,j,k} \le \eps \vee \eps'_{jk} \quad \forall m\ge m_\eps+2, \forall j, k
\eel
where $\eps'_{rk}=\psi_0$ for $k>k_0$, 
$\eps'_{jk}=\eps_{jk}$ for $j<r$ or $k\le k_0$, 
with each fixed $k_0\ge 2$. 
This is done by comparing \eqref{psi*} with 
\bel{psi'}
\psi'_{m,j,k} &=& 
\Big(C_{0,\alpha}(\lam_1/\lam_j)\textprod_{\ell\neq k}\psi'^2_{m-1,\ell}\Big)
\vee \epsilon'_{jk} \vee \Big(C_{0,\alpha}\sqrt{r/n}\sqrt{\lam_1/\lam_j}
\textprod_{\ell\neq k}\psi'_{m-1,\ell}\Big), 
\cr \psi'_{m,k} &=& \psi'_{m,r,k}, 
\eel
with $\psi'_{0,j,k}=\psi_0$. Because $\epsilon_{jk} \le \epsilon'_{jk}$, 
\bel{compare-psi'}
\psi^*_{m,j,k}\le \psi'_{m,j,k},\ \forall\ m,j,k. 
\eel
Let $m_*=\min\{m: \psi'_{m,k_0}\le\eps\}$. 
For $m < m_*$, $\psi'_{m,k}=\psi'_{m,k_0}$ for $k\le k_0$ 
and $\psi'_{m,k}=\psi_0$ for $k_0<k\le K$, so that 
\bes
& C_{0,\alpha}(\lam_1/\lam_r)\psi'^{2(k_0-1)}_{m_*-1,k_0}\psi_0^{2(K-k_0)} \le \eps, 
\cr & C_{0,\alpha}\sqrt{r/n}\sqrt{\lam_1/\lam_r}\psi'^{k_0-1}_{m_*-1,k_0}\psi_0^{K-k_0} \le \eps, 
\ees
which implies $\psi'_{m_*,j,k} \le \eps \vee \eps'_{jk}\, \forall j,k$, 
due to $\lam_j\ge\lam_r$ and $\psi'_{m_*-1,k_0}\le\psi_0$. 
Consequently, \eqref{psi-bd} holds by \eqref{compare-psi'}. 
We note that when $\psi^*_{m,2}=\eps_{r2}$, $\psi^*_{m+1,j,k}\le \eps_*\vee\sqrt{\eps_*\eps_0}\vee\eps_{jk}$.

It remains to prove $m_*\le m_\eps+2$. 
For $\eps \ge C_{0,\alpha}r/n$, 
\bes
\eps<\psi'_{m,k_0} = \psi_0\rho^{1+(2k_0-2)+\cdots+(2k_0-2)^{m-1}}  
\ees
with the $\rho<1$ in \eqref{C_0} for $m < m_*$, so that 
\bes
m_* - 2 \le \lceil  \log(\log(\eps/\psi_0)/\log\rho)/\log(2k_0-2)\rceil \le m_\eps. 
\ees
Let $m_1=\min\{m: \psi'_{m,k_0}\le C_{0,\alpha}r/n\}$ and $n_1=\{1+\ldots+(2k_0-2)^{m_1-2}\}I\{m_1\ge 2\}$. 
For $\eps < C_{0,\alpha}r/n$, we have 
\begin{align*}
\eps <& \psi'_{m,k_0}  = \psi_0\rho_1^{1+\ldots+(k_0-1)^{m-m_1}}\rho^{n_1(k_0-1)^{m-m_1+1}} 
\le \psi_0\rho^{1+\ldots+(k_0-1)^{m-1}}
\end{align*}
for $m_1\le m<m_*$, so that 
\begin{align*}
m_* - 2 
\le&\begin{cases} 
\lceil  \log(\log(\eps/\psi_0)/\log\rho)/\log(k_0-1)\rceil, & k_0>2, \cr 
\lceil  \log(\eps/\psi_0)/\log\rho)\rceil, & k_0 = 2.
\end{cases}
\end{align*}
Again $m_*\le m_\eps+2$. 
\end{proof}

\theoremdenoisecpca*

\begin{proof}[\bf Proof of Theorem \ref{thm:initial}]
Let $a_{j,S}=\vec ( \otimes_{k \in S}a_{jk})$ and $a_{j,S^c} =\vec (\otimes_{k \in [N]\backslash S}~a_{j,k})$. 
Let $U = (u_1,\ldots,u_r)$ and $V = (v_1,\ldots,v_r)$ be the orthonormal matrices in 
Proposition \ref{lemma-transform-ext} with $A$ and $B$ there replaced respectively 
by $A_S$ and $A_{S^c}$. By Proposition \ref{lemma-transform-ext},  
\bel{eq1:thm:initial}
\|a_{j,S} a_{j,S}^\top -  u_{j}  u_{j}^{\top} \|_{\rm S}\vee 
\| a_{j,S^c} a_{j,S^c}^\top -  v_{j}  v_{j}^{\top} \|_{\rm S} \le \delta,\quad \big\|\mat_{S}(T^*) - U\Lambda V^\top\big\|_{\rm S}\le \sqrt{2}\delta\lam_1, 
\eel
where $T^*=\sum_{j=1}^r \lambda_j \otimes_{k=1}^N a_{jk}$. 
Let $\Psi^* = \mat_{S}(\Psi)=\mat_{S}(T-T^*)$. We have 
\bes
\big\|\mat_{S}(T) - U\Lambda V^\top\big\|_{\rm S}\le \sqrt{2}\delta\lam_1 + \|\Psi^*\|_{\rm S}. 
\ees
As $\lambda_1>\lambda_2>...>\lambda_r> \lambda_{r+1}=0$, Wedin's perturbation theorem \citep{wedin1972} provides 
\begin{align} 
\max\left\{ \|\widehat  a_{j,S}\widehat  a_{j,S}^\top - u_{j}  u_{j}^{\top} \|_{\rm S}, \|\widehat  a_{j,S^c} \widehat  a_{j,S^c}^\top - v_{j}  v_{j}^{\top} \|_{\rm S} \right\} 
\le \frac{2 \sqrt{2} \lambda_1 \delta + 2 \|\Psi^*\|_{\rm S} }{\min\{ \lambda_{j-1}-\lambda_j, \lambda_j-\lambda_{j+1}\}}. \label{eq3:thm:initial}
\end{align}
Combining \eqref{eq1:thm:initial} 
and \eqref{eq3:thm:initial}, we have
\begin{align}\label{eq4:thm:initial}
\max\left\{ \|\widehat  a_{j,S} \widehat  a_{j,S}^\top - a_{j,S}  a_{j,S}^{\top} \|_{\rm S}, \|\widehat  a_{j,S^c} \widehat  a_{j,S^c}^\top - a_{j,S^c}  a_{j,S^c}^{\top} \|_{\rm S} \right\} &\le \delta+\frac{2\sqrt{2}\lambda_1 \delta + 2 \|\Psi^*\|_{\rm S} }{\min\{ \lambda_{j-1}-\lambda_j, \lambda_j-\lambda_{j+1}\}}.
\end{align}
By Theorem II.13 in \cite{davidson2001local}, for any $x>0$,
\begin{align*}
\P \left( \|\Psi^*\|_{\rm S}/\sigma > \sqrt{\prod_{k \in S} d_k}+ \sqrt{\prod_{k\in [N]\backslash S} d_k}+x \right)  \le e^{-x^2/2}.
\end{align*}
It implies that, choosing $x=2\sqrt{d_{S}}+ 2\sqrt{d_{S^c}}$, in an event with probability at least $1-e^{-2 d_{S} -2 d_{S^c} }$,
\begin{align}\label{eq5:thm:initial}
\|\Psi^*\|_{\rm S} \le 3\sigma\sqrt{d_{S}}+ 3\sigma\sqrt{d_{S^c}} .
\end{align}
We formulate each $\widehat u_j\in\R^d$ to be a $K$-way tensor $\widehat U_j\in \R^{d_1\times\cdots\times d_K}$. Let $\widehat U_{jk}=\mat_k(\widehat U_j)$, which is viewed as an estimate of $a_{jk}\vec(\otimes_{l \in S\backslash \{k\}}~a_{jl})^\top\in\R^{d_k\times (d_S/d_k)}$. Then $\cpca_{jk}$ is the top left singular vector of $\widehat U_{jk}$. 
By Proposition \ref{prop-rank-1-approx}, for any $k\in S$
\bes
\|\cpca_{jk} \cpcatop_{jk} - a_{jk} a_{jk}^\top \|_{\rm S}^2 \wedge (1/2) \le 
\|\widehat a_{j,S} \widehat a_{j,S}^\top - a_{j,S} a_{j,S}^{\top} \|_{\rm S}^2 .
\ees
Similar bound can be obtained for $\|\cpca_{jk} \cpcatop_{jk} - a_{jk} a_{jk}^\top \|_{\rm S}$ 
for $k\in S^c$. Substituting \eqref{eq4:thm:initial} and \eqref{eq5:thm:initial} into the above equation, we have the desired results. 
\end{proof}

\theoremdenoiseico*

\begin{proof}[\bf Proof of Theorem \ref{thm:projection}]
Let $\psi_{0,\ell}=\psi_0$ and define sequentially 
\bel{psi_mk2}
\phi^*_{m,k-1} &=& (N-1)\alpha_*^{-1}\sqrt{2r}\max_{1\le \ell<N}\psi_{m,k-\ell},
\cr \psi_{m,k} &=& \Big(6\alpha_*^{1-N}\sqrt{r-1}(\lambda_1/\lambda_j) \textprod_{\ell=1}^{N-1}\psi_{m,k-\ell}\Big)
\vee \Big(6\alpha_*^{1-N}R^{*\ideal}_{jk,\phi^*_{m,k-1}}\Big),
\eel 
$k=1,\ldots,N$, $m=1,2,\ldots$ 
By induction, \eqref{thm-projection:eq1} gives $\psi_{m,k}\le\psi_{m-1,k} \le\psi_0$. 
Here and in the sequel, we 
take the convention that $(m,\ell)=(m-1,N+\ell)$ with the subscript $(m,\ell)$, 
and that $\times_\ell \widehat\theta_{j,\ell}^{(m)} = \times_{N+\ell} \widehat\theta_{j,N+\ell}^{(m-1)}$ 
for any estimator $\widehat\theta_{jk}^{(m)}$. Let 
\bel{thm4:psi}
\Omega^*_{m,k-1} =\cap_{\ell=1}^{N-1}\Omega_{m,k-\ell}
\eel 
with $\Omega_{m,\ell}=\big\{\textmax_{h\le r}\big(2- 2|a_{h\ell}^{\top}\widehat a_{h\ell}^{(m)}| \big)^{1/2} \le \psi_{m,\ell}\big\}$.
Let $\widehat g_{j\ell}^{(m)} =\widehat b_{j\ell}^{(m)}/\|\widehat b_{j\ell}^{(m)}\|_2$, 
and $g_{j\ell}=b_{j\ell}/\|b_{j\ell}\|_2$. 
By \eqref{a-bd}, \eqref{b-bd} and \eqref{g-bd} in the proof of Theorem \ref{thm:projection2}, 
\bel{old-bd}
\big\|\ghatm_{j\ell} - g_{j\ell}\big\|_2
\le (\psi_{m,\ell}/\alpha_*)\sqrt{2r},\ \ 
\big|a_{j\ell}^\top\ghatm_{j\ell}\big| \ge \alpha_*, \ \ 
\eel
in $\Omega_{m,\ell}$ with the $\alpha_*$ in \eqref{alpha2}.  

Given $\{\widehat a_{j,k-\ell}^{(m)}, j\in[r], \ell\in [N-1]\}$, the $m$-th iteration for tensor mode 
$k$ produces estimates $\widehat a_{jk}^{(m)}$ as the normalized version of 
$T \times_{\ell=k-1}^{k-N+1} \widehat b_{j,\ell}^{(m)\top}$. Because 
$T =\sum_{j=1}^r\lambda_j\otimes_{k=1}^{N} a_{jk} + \Psi$, 
the ``noiseless" version of this update is given by
\begin{equation}
T \times_{\ell\in [ N] \backslash\{k\} }  b_{j\ell}^{\top}
=\lambda_{j} a_{jk} + \Psi \times_{\ell\in [ N] \backslash\{k\} }  b_{j\ell}^{\top} \in\R^{d_k}.
\end{equation}
Similarly, for any $1\le j\le r$, 
\begin{align*}
T \times_{\ell=k-1}^{k-N+1} \widehat b_{j,\ell}^{(m)\top} 
= \sum_{h=1}^r \widetilde\lambda_{h,j}  a_{hk}  + \Psi \times_{\ell=k-1}^{k-N+1} \widehat b_{j,\ell}^{(m)\top} \in \R^{d_k},
\end{align*}
where 
$\widetilde\lambda_{h,j}=\lambda_h\prod_{\ell=1 }^{N-1}  a_{h,k-\ell}^\top\widehat b_{j,k-\ell}^{(m)}$.   
Let 
\begin{align*}
\widetilde \phi_{m,\ell} = \psi_{m,\ell}/\big(\sqrt{1-\delta_{\max}}-\sqrt{r}\psi_{m,\ell} \big)_+.
\end{align*} 
By the definition of $\alpha_*$ in \eqref{alpha2} and the condition  
$\psi_{m,\ell}\le\psi_0$, $\widetilde\phi_{m,\ell}/(1-\widetilde\phi_{m,\ell})\le \psi_{m,\ell}/\alpha_*$.  
Thus, by the arguments in the proof of Proposition \ref{prop-ICO-gen}, 
\begin{align}\label{eqthm2:wedin72}
\big(2- 2| a_{jk}^{\top}\widehat a_{jk}^{(m)}| \big)^{1/2} \le& \frac{\sqrt{2}\big\| \Psi \times_{\ell=k-1}^{k-N+1} \widehat g_{j,\ell}^{(m)\top}\big\|_{2} }{\lambda_j\prod_{\ell=1 }^{N-1}  a_{j,k-\ell}^\top\widehat g_{j,k-\ell}^{(m)}} 
 + \frac{\lam_1\sqrt{2(1+\delta_k)}}{\lam_j/\sqrt{r-1}}\prod_{\ell=1}^{N-1} 
\frac{\psi_{m,k-\ell} }{\alpha_*}
\end{align}
in $\Omega^*_{m,k-1}$. As $\Psi \times_{\ell=k-1}^{k-N+1} \widehat g_{j,\ell}^{(m)\top}$ is linear in each $\widehat  g_{j\ell}^{(m)}$, 
\begin{align*}
\big\|  \Psi \times_{\ell=k-1}^{k-N+1} \widehat g_{j,\ell}^{(m)\top} \big\|_{2}
\le& (N-1)\max_{\ell<N} \| \widehat g_{j,k-\ell}^{(m)}-  g_{j,k-\ell} \|_2 \| \Delta \|
+ \big\| \Psi \times_{\ell\in [ N] \backslash\{k\} } g_{j\ell}^{\top} \big\|_{2}, 
\end{align*}
where $\|\Delta \|= \max_{v_\ell\in \mathbb S^{d_\ell-1}\forall\ell}\big( \Psi  \times_{\ell=1}^N v_\ell^\top\big)$.  
As we also have $\big\|  \Psi \times_{\ell=k-1}^{k-N+1} \widehat g_{j,\ell}^{(m)\top} \big\|_{2}\le \|\Delta\|$, 
\eqref{old-bd} yields 
\begin{align}\label{eqthm2:norm-bd}
\big\|\Psi \times_{\ell=k-1}^{k-N+1} \widehat g_{j,\ell}^{(m)\top} \big\|_{2} 
\le& \min\big\{\|\Delta\|, \phi^*_{m,k-1} \| \Delta \| 
+ \big\|\Psi \times_{\ell\in [ N] \backslash\{k\} } g_{j\ell}^{\top} \big\|_{2}\big\} 
\end{align}
in $\Omega^*_{m,k-1}$, in view of the definition of $\phi^*_{m,k-1}$ in \eqref{psi_mk2}. 
By the Sudakov-Fernique and Gaussian concentration inequalities, 
\begin{align*} \P\left( \| \Delta \|/\sigma > \sum_{\ell=1}^N \sqrt{d_\ell}+x \right) \le e^{-x^2/2} 
\end{align*}
and $\P\{\|\Psi \times_{\ell\in [N] \backslash\{k\}} g_{j\ell}^\top \|_{2}>\sqrt{d_k}+x\}\le e^{-x^2/2}$. Thus, 
\bes
&& \|\Delta \| \le \sigma \textsum_{\ell=1}^N\sqrt{d_\ell} + \sigma\sqrt{2d_N},
\cr && \big\|\Psi \times_{\ell\in [ N] \backslash\{k\} } g_{j\ell}^{\top} \big\|_{2}  \le 
(1+\sqrt{2})\sigma \sqrt{d_k},
\ees
in an event $\Omega_1$ with at least probability $1-\sum_{k=1}^N e^{-d_k} - e^{-d_N}$. 
Consequently, by \eqref{eqthm2:norm-bd}, in $\Omega_1\cap\Omega^*_{m,k-1}$, 
\bel{eqthm2:bdd-psi}
&& \big\| \Psi \times_{\ell=k-1}^{k-N+1} \widehat g_{j,\ell}^{(m)\top}\big\|_{2}
\big/\big(\lam_j\textprod_{\ell=1 }^{N-1}  a_{j,k-\ell}^\top\widehat g_{j,k-\ell}^{(m)}\big)
\notag\\
&\le& (1+\sqrt{2})\sigma \big(d_k^{1/2} + (\phi^*_{m,k-1}\wedge 1)\textsum_{\ell=1}^N d_\ell^{1/2}\big)\alpha_*^{1-N}/\lam_j  
\notag\\
& \le & \sqrt{8}\alpha_*^{1-N}R^{*\ideal}_{jk,\phi^*_{m,k-1}}
\eel
Substituting
\eqref{eqthm2:bdd-psi}  into \eqref{eqthm2:wedin72}, we have, 
in the event $\Omega_1\cap\Omega_{m,k-1}$, 
\bel{bdd1:thm-projection0}
\big(2- 2| a_{jk}^{\top}\widehat a_{jk}^{(m)}| \big)^{1/2} 
\le \psi_{m,j,k} 
\eel
with 
\bes
\psi_{m,j,k} =\max\bigg\{\frac{6R^{*\ideal}_{jk,\phi^*_{m,k-1}}}{\alpha_*^{N-1}} , 
\frac{6\lam_1\sqrt{r-1}}{\lam_j\alpha_*^{N-1}} \prod_{\ell=1}^{N-1}\psi_{m,k-\ell}\bigg\}. 
\ees
Consequently, $\Omega_{m,k}\subset \Omega_1\cap\Omega^*_{m,k-1}$ and 
%
the upper bound for required number of iterations follows from 
the same (but much simpler) argument in Steps 2 and 3 of the proof of Theorem~\ref{thm:projection2}. 

As for the estimation of $\lambda_j$, similar to \eqref{eqthm2:wedin72}, we can obtain
\begin{align}\label{eqthm2:eigen}
\big|\widehat\lambda_j^{(m)} - \lambda_j \big| 
\le& \big\| \Psi \times_{\ell\in [N]} \widehat b_{j\ell}^{(m)\top} \big\|_{2} + (r-1)\lambda_1 \prod_{\ell=1}^{N}\phi_{m,\ell}  +\sum_{\ell=1}^{N} \phi_{m,\ell}.
\end{align}
Then, employing similar procedures as above, we can prove the bound \eqref{thm-projection:eq3}. 
\end{proof}

\subsection{Technical Lemmas}

We collect all technical lemmas that has been used in the theoretical proofs throughout the paper in this section. We denote the Kronecker product $\odot$ as $A\odot B\in \RR^{m_1 m_2 \times r_1 r_2}$, for any two matrices $A\in\RR^{m_1\times r_1},B\in \RR^{m_2\times r_2}$. 
\begin{lemma}\label{lemma:epsilonnet}
Let $d, d_j, d_*, r\le d\wedge d_j$ be positive integers, $\epsilon>0$ and
$N_{d,\epsilon} = \lfloor(1+2/\epsilon)^d\rfloor$. \\
(i) For any norm $\|\cdot\|$ in $\R^d$, there exist
$M_j\in \R^d$ with $\|M_j\|\le 1$, $j=1,\ldots,N_{d,\epsilon}$,
such that $\max_{\|M\|\le 1}\min_{1\le j\le N_{d,\epsilon}}\|M - M_j\|\le \epsilon$.
Consequently, for any linear mapping $f$ and norm $\|\cdot\|_*$,
$$
\sup_{M\in \R^d,\|M\|\le 1}\|f(M)\|_* \le 2\max_{1\le j\le N_{d,1/2}}\|f(M_j)\|_*.
$$
(ii) Given $\epsilon >0$, there exist $U_j\in \R^{d\times r}$
and $V_{j'}\in \R^{d'\times r}$ with $\|U_j\|_{\rm S}\vee\|V_{j'}\|_{\rm S}\le 1$ such that
$$
\max_{M\in \R^{d\times d'},\|M\|_{\rm S}\le 1,\text{rank}(M)\le r}\
\min_{j\le N_{dr,\epsilon/2}, j'\le N_{d'r,\epsilon/2}}\|M - U_jV_{j'}^\top\|_{\rm S}\le \epsilon.
$$
Consequently, for any linear mapping $f$ and norm $\|\cdot\|_*$ in the range of $f$,
\begin{equation}\label{lm-3-2}
\sup_{M, \widetilde M\in \R^{d\times d'}, \|M-\widetilde M\|_{\rm S}\le \epsilon
\atop{\|M\|_{\rm S}\vee\|\widetilde M\|_{\rm S}\le 1\atop
\text{rank}(M)\vee\text{rank}(\widetilde M)\le r}}
\frac{\|f(M-\widetilde M)\|_*}{\epsilon 2^{I_{r<d\wedge d'}}}
\le \sup_{\|M\|_{\rm S}\le 1\atop \text{rank}(M)\le r}\|f(M)\|_*
\le 2\max_{1\le j \le N_{dr,1/8}\atop 1\le j' \le N_{d'r,1/8}}\|f(U_jV_{j'}^\top)\|_*.
\end{equation}
(iii) Given $\epsilon >0$, there exist $U_{j,k}\in \R^{d_k\times r_k}$
and $V_{j',k}\in \R^{d'_k\times r_k}$ with $\|U_{j,k}\|_{\rm S}\vee\|V_{j',k}\|_{\rm S}\le 1$ such that
$$
\max_{M_k\in \R^{d_k\times d_k'},\|M_k\|_{\rm S}\le 1\atop \text{rank}(M_k)\le r_k, \forall k\le K}\
\min_{j_k\le N_{d_kr_k,\epsilon/2} \atop j'_k\le N_{d_k'r_k,\epsilon/2}, \forall k\le K}
\Big\|\odot_{k=2}^K M_k - \odot_{k=2}^K(U_{j_k,k}V_{j_k',k}^\top)\Big\|_{\rm op}\le \epsilon (K-1).
$$
For any linear mapping $f$ and norm $\|\cdot\|_*$ in the range of $f$,
\begin{equation}\label{lm-3-3}
\sup_{M_k, \widetilde M_k\in \R^{d_k\times d_k'},\|M_k-\widetilde M_k\|_{\rm S}\le\epsilon\atop
{\text{rank}(M_k)\vee\text{rank}(\widetilde M_k)\le r_k \atop \|M_k\|_{\rm S}\vee\|\widetilde M_k\|_{\rm S}\le 1\ \forall k\le K}}
\frac{\|f(\odot_{k=2}^KM_k-\odot_{k=2}^K\widetilde M_k)\|_*}{\epsilon(2K-2)}
\le \sup_{M_k\in \R^{d_k\times d_k'}\atop {\text{rank}(M_k)\le r_k \atop \|M_k\|_{\rm S}\le 1, \forall k}}
\Big\|f\big(\odot_{k=2}^K M_k\big)\Big\|_*
\end{equation}
and
\begin{equation}\label{lm-3-4}
\sup_{M_k\in \R^{d_k\times d_k'},\|M_k\|_{\rm S}\le 1\atop \text{rank}(M_k)\le r_k\ \forall k\le K}
\Big\|f\big(\odot_{k=2}^K M_k\big)\Big\|_*
\le 2\max_{1\le j_k \le N_{d_kr_k,1/(8K-8)}\atop 1\le j_k' \le N_{d_k'r_k,1/(8K-8)}}
\Big\|f\big(\odot_{k=2}^K U_{j_k,k}V_{j_k',k}^\top\big)\Big\|_*.
\end{equation}
\end{lemma}
\begin{proof}
See technical lemmas in \cite{han2020iterative}.
\end{proof}

\begin{lemma}\label{lm-G2}
Let $G\in \R^{n\times d}$ be a Gaussian matrix with i.i.d. $N(0,1)$ entries. Then,
\begin{align*}
\left\|\frac1n G'G-I_{d} \right\|_{\rm S} \le 2 \left(\sqrt{\frac{d}{n}}+x \right) + \left(\sqrt{\frac{d}{n}}+x \right)^2,   
\end{align*}
with at least probability $1-2e^{-nx^2/2}$ for any $x>0$.
\end{lemma}
\begin{proof}
Let $X=n^{-1}G'G$ and $t=2 (\sqrt{\frac{d}{n}}+x ) + (\sqrt{\frac{d}{n}}+x )^2$. Then, we have 
\begin{align*}
&\{ \sigma_1(X) > 1+t \} \subset \{\sigma_1(G) >\sqrt{n} +\sqrt{d} +\sqrt{n}x \},\\
&\{ \sigma_{\min}(X) < 1-t \} \subset \{\sigma_{\min}(G) < \sqrt{n} - \sqrt{d} - \sqrt{n}x \}.
\end{align*}
By Theorem II.13 in \cite{davidson2001local}, for any $x>0$,
\begin{align*}
\max\left\{\P \left( \sigma_1(G) > \sqrt{n}+\sqrt{d}+\sqrt{n}x \right), \P \left( \sigma_{\min}(G) < \sqrt{n}-\sqrt{d}-\sqrt{n}x \right) \right\} \le e^{-nx^2/2}.  
\end{align*}
As $\|X-I_{d}\|_{\rm S}=\max\{\sigma_1(X)-1, 1-\sigma_{\min}(X)\}$, it follows that
\begin{align*}
\P\left( \|X-I_{d}\|_{\rm S} > t  \right) &\le \P\left( \sigma_1(X) > 1+t  \right) +\P\left( \sigma_{\min}(X) < 1-t\right)\\
&\le  \P \left( \sigma_1(G) > \sqrt{n}+\sqrt{d}+\sqrt{n}x \right) + \P \left( \sigma_{\min}(G) < \sqrt{n}-\sqrt{d}-\sqrt{n}x \right) \\
&\le 2e^{-nx^2/2}.
\end{align*}
This complete the proof.
\end{proof}

\bibliographystyle{apalike}
\bibliography{cp}

\end{document}